\documentclass[10pt,journal,compsoc]{IEEEtran}
\ifCLASSOPTIONcompsoc
  \usepackage[nocompress]{cite}
\else
  \usepackage{cite}
\fi
\ifCLASSINFOpdf
\else
\fi
\usepackage{booktabs}
\usepackage{amsfonts}
\usepackage{amsmath}
\usepackage{graphicx}
\usepackage{bm}
\usepackage{multirow}
\graphicspath{{./}{./figures/}}
\usepackage{color} 
\usepackage{colortbl}
\usepackage{cases}

\usepackage{tikz}
\newcommand{\tikzsquare}[2][red,fill=red]{\tikz[baseline=-0.5ex]\draw[#1] (-0.5*#2,-0.5*#2) rectangle(#2,#2) ;}%

\definecolor{ATTColor}{HTML}{FFFEEE}
\definecolor{MHAColor}{HTML}{E3F4FE}
\definecolor{FFColor}{HTML}{FDF4F3}
\newcommand{\attlayer}{\tikzsquare[fill=ATTColor]{3.5pt}}
\newcommand{\mhalayer}{\tikzsquare[fill=MHAColor]{3.5pt}}
\newcommand{\fflayer}{\tikzsquare[fill=FFColor]{3.5pt}}

\newcommand{\hidden}{\mathbf{h}}

\newcommand{\point}{\mathbf{x}}

\newcommand{\weight}{\mathbf{W}}
\newcommand{\query}{\mathbf{q}}
\newcommand{\key}{\mathbf{k}}
\newcommand{\valvec}{\mathbf{v}}
\DeclareMathOperator{\val}{val}
\DeclareMathOperator{\att}{att}
\DeclareMathOperator{\mlp}{MLP}
\newcommand{\setof}[1]{\{ #1 \}}

\usepackage{xspace}
\newcommand{\vonm}{\text{VON-m}\xspace}

\newcommand{\vona}{\text{VON-a}\xspace}
\newcommand{\vonc}{\text{VON-c}\xspace}
\newcommand{\abvonmue}{$\text{VON-m}_{\bar{e}}$\xspace}
\newcommand{\abvonmur}{$\text{VON-m}_{\bar{r}}$\xspace}
\newcommand{\abvonmub}{$\text{VON-m}_{\bar{b}}$\xspace}

\newcommand{\vonafg}{\text{VON-a(50g)}\xspace}
\newcommand{\vonafl}{\text{VON-a(50l)}\xspace}
\newcommand{\vonmfg}{\text{VON-m(50g)}\xspace}
\newcommand{\vonmfl}{\text{VON-m(50l)}\xspace}
\newcommand{\voncfg}{\text{VON-c(50g)}\xspace}
\newcommand{\voncfl}{\text{VON-c(50l)}\xspace}
\newcommand{\vonmag}{\text{VON-m(100g)}\xspace}
\newcommand{\vonmal}{\text{VON-m(100l)}\xspace}
\newcommand{\vonmafg}{\text{VON-m(150g)}\xspace}
\newcommand{\vonmafl}{\text{VON-m(150l)}\xspace}
\newcommand{\amfg}{\text{AM(50g)}\xspace}
\newcommand{\amfl}{\text{AM(50l)}\xspace}

\newcommand{\vonmfmix}{\text{VON-m50mix}\xspace}
\newcommand{\vonmfgl}{\text{VON-m(50gl)}\xspace}
\newcommand{\vonmflg}{\text{VON-m(50lg)}\xspace}
\newcommand{\amfmix}{\text{AM(50mix)}\xspace}
\newcommand{\amfgl}{\text{AM(50gl)}\xspace}
\newcommand{\amflg}{\text{AM(50lg)}\xspace}

\newcommand{\fmdatafg}{\text{50g}\xspace}
\newcommand{\fmdatafl}{\text{50l}\xspace}
\newcommand{\fmdataag}{\text{100g}\xspace}
\newcommand{\fmdataal}{\text{100l}\xspace}
\newcommand{\fmdataafg}{\text{150g}\xspace}
\newcommand{\fmdataafl}{\text{150l}\xspace}
\newcommand{\fmdatafmix}{\text{50mix}\xspace}

\DeclareMathOperator{\conv}{Conv}

\newenvironment{myitemize}{
\begin{itemize}
  \setlength{\itemsep}{1pt}
  \setlength{\parskip}{0pt}
  \setlength{\parsep}{0pt}}{\end{itemize}
}

\usepackage{hyperref}
\hypersetup{
  pdfauthor = {},
  pdftitle = {},
  pdfsubject = {},
  pdfkeywords = {},
  colorlinks=true,
  linkcolor= black,
  citecolor= black,
  pageanchor=true,
  urlcolor = black,
  plainpages = false,
  linktocpage
}

\usepackage[capitalise]{cleveref}

\begin{document}
%
\title{Versatile Ordering Network: An Attention-based Neural Network for Ordering Across Scales and Quality Metrics}
%
%

\author{
  Zehua~Yu, Weihan~Zhang, Sihan~Pan,
  and Jun~Tao*,~\IEEEmembership{Member,~IEEE}
\IEEEcompsocitemizethanks{
\IEEEcompsocthanksitem * denotes the corresponding author.
\IEEEcompsocthanksitem Z. Yu, W. Zhang, and S. Pan are with the School of Computer Science and Engineering, Sun Yat-sen University, China. E-mail: \{yuzh58, zhangwh79, pansh25\}@mail2.sysu.edu.cn.
\IEEEcompsocthanksitem J. Tao is with the School of Computer Science and Engineering, Sun Yat-sen University and the National Supercomputer Center in Guangzhou, China. E-mail: taoj23@mail.sysu.edu.cn. He is the corresponding author.
\protect
}
}


%
%
%
\markboth{IEEE Transactions on Visualization and Computer Graphics}{Yu \MakeLowercase{\textit{et al.}}: Versatile Ordering Network: An Attention-based Neural Network for Ordering Across Scales and Quality Metrics}
%



\IEEEtitleabstractindextext{%
\begin{abstract}
Ordering has been extensively studied in many visualization applications, such as axis and matrix reordering, for the simple reason that the order will greatly impact the perceived pattern of data. Many quality metrics concerning data pattern, perception, and aesthetics are proposed, and respective optimization algorithms are developed. However, the optimization problems related to ordering are often difficult to solve (e.g., TSP is NP-complete), and developing specialized optimization algorithms is costly. In this paper, we propose Versatile Ordering Network (VON), which automatically learns the strategy to order given a quality metric. VON uses the quality metric to evaluate its solutions, and leverages reinforcement learning with a greedy rollout baseline to improve itself. This keeps the metric transparent and allows VON to optimize over different metrics. Additionally, VON uses the attention mechanism to collect information across scales and reposition the data points with respect to the current context. This allows VONs to deal with data points following different distributions. We examine the effectiveness of VON under different usage scenarios and metrics. The results demonstrate that VON can produce comparable results to specialized solvers. The code is available at https://github.com/sysuvis/VON.
\end{abstract}


}


\maketitle

\IEEEpubidadjcol

\IEEEdisplaynontitleabstractindextext

%

\IEEEraisesectionheading{\section{Introduction}\label{sec:introduction}}

\IEEEPARstart{O}{rdering} is a key ingredient in visualization approaches, as the order of entities in visualization may greatly impact the perceived patterns, leading to different understandings of data. Therefore, ordering has been studied in many visualization scenarios. For example, axis reordering has gained considerable attention in visualization with multiple axes, such as parallel coordinates~\cite{artero2006enhanced, tyagi2023pcexpo}, star coordinates~\cite{hu2021shape}, and Radviz~\cite{di2010radviz}; branch ordering is studied in river-like visualization techniques~\cite{di2016there, bu2021sinestream, di2021sequence}; row and column reordering is studied in matrix visualization~\cite{behrisch2020guiro, vanbeusekom2022simul, kwon2022deep}; and, images ordering is studied in large image collection visualization. 

The ordering techniques seek to address two central questions: What is an ideal order (quality metric), and how to achieve the ideal order (solver)? Many quality metrics are proposed to evaluate the order of visual elements in various applications, and respective solvers are developed to optimize the order for best quality. For example, GUIRO~\cite{behrisch2020guiro} incorporates 16 quality metrics for matrix representation and 70 reordering algorithms. Although the reordering algorithm often shares similar components (e.g., swapping elements), designing algorithms for individual metric still consumes a significant amount of research effort. This leads to our question: \textbf{Given an arbitrary quality metric, can machines learn to solve the ordering problem by themselves?} By answering this question, we may free the visualization researchers from designing solvers, and allow them to focus on perception and analytic tasks, which may not be tackled solely by machines.

\begin{figure}[!htbp]
    \centerline{\includegraphics[width=.98\linewidth]{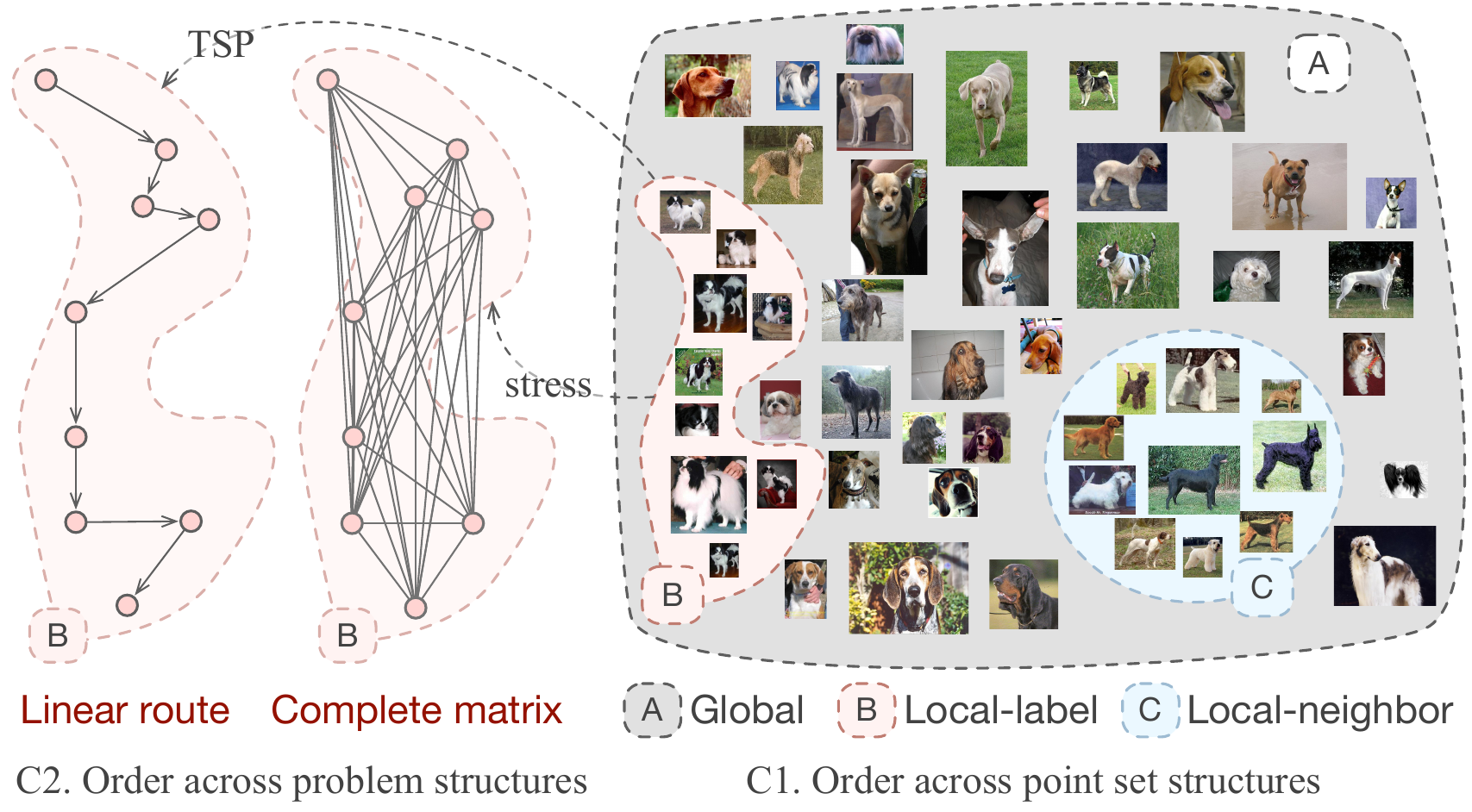}}
    \caption{Challenges in learning to order across scales and quality metrics: \textbf{C1}. The images in (A), (B), and (C) may span different regions in space and exhibit \emph{different point set structures.} \textbf{C2}. The quality metrics may enforce \emph{different problem structures}, as indicated by the graphs on the right. TSP enforces a linear route structure, while stress considers a complete graph between points. The different scales of points and quality metrics may require different ordering strategies to be learned.
    }
    \label{fig:challenges}
\end{figure}

Reinforcement learning can leverage the quality metric to evaluate solutions and improve the solver without understanding the problem structure, which is suitable in our scenario. In theoretical computer science, reinforcement learning has been applied to similar problems~\cite{nazari2018reinforcement, kool2019attention}, but the existing approaches still leave the following challenges open in visualization:

\textbf{C1. Generalizability across quality metrics}. The problem structures and factors to consider may vary across metrics. As shown in \cref{fig:challenges}, the linear route structure is enforced by the traveling salesman problem (TSP), which only evaluates the distance between neighboring data points. In contrast, stress considers all pairwise relationships between points, leading to a significantly different problem structure of a complete graph. We expect that the problem structure can be more different when it comes to perceptual or aesthetic metrics. This requires the network to be able to incorporate different kinds of information. 

\textbf{C2. Transferability across data distributions and scales}. In visualization, the data items being ordered may be dynamically generated during interaction. For example, when visualizing a collection of dog images, as shown in \cref{fig:challenges}, users may view images from the entire space (A), they may view the images of a certain breed of dogs (B), or they may explore images similar to a specified one (C). 
The images in (A), (B), and (C) correspond to different distributions, exhibiting different point cloud structures and requiring different ordering strategies.

\textbf{C3. Efficiency}. The solver should produce the order in real time, especially if it is used in an interactive system. It should enhance development efficiency by minimizing the effort required to develop specialized algorithms for different metrics or data distributions.

In this paper, we propose Versatile Ordering Network (VON) to tackle the above challenges. VON is a set-to-sequence network that consumes a set of data points and produces a sequence (order). It is trained via reinforcement learning with a given quality metric as the loss function. Note that VON only uses the quality metric as a black box to evaluate the ordering results without explicitly understanding the problem structure. This allows VON to apply on arbitrary quality metric (\textbf{C1}). 
In addition, VON leverages the attention mechanism to collect information about global structures from all points and the local neighborhood of the ordering front. It also enables the message passing between all nodes, including the ordered and unordered nodes. Therefore, it can adapt to various kinds of underlying data patterns and quality metrics (\textbf{C1}). To order data points with different scales and spatial locations, VON is also equipped with a specially designed repositioning module. This module can be considered a learnable normalization operator that removes the location, orientation, and scale information so that the network can focus on the structure of data, especially the local neighborhood around the ordering front. The repositioning module allows the learned ordering strategy to be applied to points at unseen locations or scales (\textbf{C2}). This enables VON to order data items during interaction without further updating in most cases (\textbf{C3}).

We examine the performance of VON under two scenarios: First, ordering a subset of a large collection of data items with a given quality metric. This may apply to interactive visualization, such as filtering a large image collection. Second, ordering a given set of points according to a given quality metric. Axes reordering and matrix reordering are examples of this scenario. We comprehensively evaluate the performance of VON across metrics, data scales, and distributions with extensive experiments involving twelve datasets, twelve sampling strategies, thirteen baselines, and eleven metrics. 

Our contributions are as follows. First, we propose a novel ordering network that can be used across tasks, objectives, and data distributions. By leveraging the power of reinforcement learning and a greedy rollout strategy, the network can automatically learn ordering strategies to improve itself. Second, we propose a conceptual framework for ordering and provide several realizations to implement the conceptual model. In particular, we design a repositioning module that automatically learns to place the data points in the current ordering context. This can be seen as a learnable nonlinear transformation that removes the information irrelevant to the structure, improving transferability. In addition, we conduct extensive experiments to examine the effectiveness and robustness of our method.

\section{Related work}
The ordering problem has been studied in the visualization from two aspects: how to design visual quality metrics for ordering, and how to achieve the optimal ordering given a metric? Our approach is more closely related to the approaches in the second category. It aims to develop a generic solver that automatically learns how to optimize the order based on a quality metric definition. However, it does not study whether the quality metric will enhance perception or improve task performance in analytics. In this section, we discuss the techniques across various ordering applications in visualization: such as axis reordering, branch ordering, and matrix reordering. We also review the ordering techniques in theoretical computer science, with an emphasis on the learning-to-rank approaches. For the visual quality metric design, We refer the readers to the surveys \cite{liiv2010seriation,behrisch2016matrix,behrisch2018quality}, as they are less relevant.

{\bf Image ordering.} Image ordering is often modeled as a 1D-grid arrangement problem. For example, SOM~\cite{kohonen1990som} learns the low-dimensional representations, SSM~\cite{strong2013organizing} arranges items based on position swapping, and LDG~\cite{frey2022optimizing} uses local search. Kernelized sorting~\cite{quadrianto2008kernelized} models the arrangement as a quadratic assignment problem. IsoMatch~\cite{fried2015isomatch} uses the Hungarian algorithm for optimal matching between Isomap embedding and target grid points. DS++~\cite{Nadav2017DS} introduces a convex quadratic programming relaxation for the matching process. LAS/FLAS\cite{barthel2023improved} designs a distance preservation metric based on neighborhood preservation and employs linear assignment ordering.

{\bf Axes reordering in visualization.} Axes ordering aims to enhance patterns revealed by visualization with multiple axes, such as parallel coordinates~\cite{johansson2015evaluation} and star coordies~\cite{miller2019evaluating}, etc. These approaches often design a metric to guide the ordering optimization. For example, Peng et al. \cite{peng2004clutter} reduced glyph clutter by considering monotonicity and symmetricity, and Artero et al.~\cite{artero2006enhanced} arranged the axes based on their similarity. Tyagi et al. \cite{tyagi2023pcexpo} introduced twelve local properties with customizable weights to order axes for parallel coordinates. For star plots, shape similarities of contours are often considered~\cite{fuchs2014influence,ankerst1998similarity}. For interaction, SMARTexplore~\cite{blumenschein2018smartexplore} simplified the analysis of high-dimensional data, and RankAxis~\cite{Liu2023RankAxis} is designed to order multiple attributes.

{\bf Branch ordering in river-like visualization.} 2D River-like visualizations (e.g., streamgraphs, ThemeRivers, storylines) often use one dimension to represent time evolution and arrange branches in the other dimension to encode their relationships.
Byron and Watternberg~\cite{Lee2008stacked} formalized a series of aesthetic criteria for streamgraphs.
Di Bartolomeo et al.~\cite{di2016there} swapped the layers in a streamgraph to reduce the wiggle value, and Bu et al.~\cite{bu2021sinestream} introduced Sinttream to enhance readability.
Zhang et al.~\cite{zhang2021surfriver} adapted 3D stream surfaces to 2D, preserving inter-branch distances.
For storytelling, sequence ordering is guided by logical structures~\cite{shi2021calliope}, user intentions~\cite{tang2021plotthread}, and minimizing edge crossings~\cite{di2021sequence}.

{\bf Matrix reordering.} 
The data topology revealed by matrix representation depends on the order of nodes. However, reordering nodes often involves various factors~\cite{vanbeusekom2022simul}.
Behrisch et al.~\cite{behrisch2020guiro} proposed GUIRO to enhance the transparency of matrix reordering algorithms. Users can explore the permutation space with 70 reordering algorithms and 16 quality metrics, viewing matrix rows or columns as 2D paths on the screen.
Van Beusekom et al.~\cite{vanbeusekom2022simul} simultaneously ordered a collection of matrices by maximizing their Moran's I. The desired order is derived from the solution of TSP, which can be approximated by nearest neighbor heuristics.
Kwon et al. \cite{kwon2022deep} use an autoencoder to capture order patterns, enabling the exploration of matrix reordering results of 27 algorithms.
Al-Naami et al.~\cite{al2024improved} investigated how orderable node-link layouts enhanced the visual saliency of graph clusters, by assessing the impact of factors such as graph layout and node ordering methods on the accuracy and completion time of cluster count judgments.

{\bf Other applications of ordering in visualization.} Ordering is studied in other visualization scenarios as well. For example, in volumetric visualization, Frey et al.~\cite{frey2021temporally} ordered 2D curves based on their shape similarities using self-organization map.
Weiss et al.~\cite{weiss2019dynamic} used the 1D Hilbert space-filling curve to linearize a volume and Zhou et al.~\cite{zhou2021data} proposed a data-driven space-filling curve based on Hamiltonian paths. 
Dobler et al. \cite{dobler2022computing} proposed a TSP-based algorithm to compute linear diagrams. 
Fuchs et al.~\cite{fuchs2024quality} explored how to reveal topological patterns in BioFabric visualizations by proposing combinations of various quality metrics and ordering strategiessy. Nate et al.~\cite{nate2024attribute} proposed a Hilbert reordering scheme to reduce the memory consumption of trees in smooth particle visualization.
Instead of the metric-based optimization, ranking approaches often emphasize the user involvement to weight multiple attributes for decision-making, such as ValueCharts~\cite{Giuseppe2004ValueCharts}, LineUp~\cite{Samuel2013LineUp}, WeightLifter~\cite{Stephan2017WeightLifter}, Podium~\cite{Emily2018Podium}, FairSight~\cite{Ahn2019FairSight}, SRVis~\cite{Weng2019SRVis}, FairFuse~\cite{Hilson2022FairFuse}, and RankAxis~\cite{Liu2023RankAxis}. We refer the readers to the original papers for details as this direction is less relevant.

{\bf Ordering in theoretical computer science.} In theoretical computer science, ordering is often studied in the form of traveling sales problem or Hamiltonian circle. 
Both problems are proved to be NP-complete without efficient deterministic solvers. 
Conventional approaches are often natural-inspired.
For example, Dorigo et al.~\cite{dorigo1997ant} proposed the Ant Colony System, an algorithm with multiple agents (i.e., ants). Ants cooperate by depositing information on edges to build solutions.
Chen et al. \cite{chen2009ranking} revealed the relationship between ranking measures and loss functions in learning-to-rank methods.

{\bf Set-to-sequence in machine learning.} 
Set-to-sequence approaches~\cite{jurewicz2021set} generate sequences from unordered data points, often driven by the need for permutation invariance. 
These approaches are often equipped with a set-input encoding module for sequence predictions, including RNN~\cite{sutskever2014sequence}, reinforcement learning~\cite{bahdanau2016actor}, and transformer~\cite{Ashish2017Attention, devlin2018bert}. Transformers, especially those like Set Transformer~\cite{lee2019set} and Pointer Network~\cite{vinyals2015pointer}, are increasingly popular for ranking and selection tasks due to their attention mechanisms. For solving the TSP, transformers often frame route construction as a set-to-sequence learning task~\cite{chen2023set, goh2024hierarchical}.

Recent approaches use neural networks to learn ordering strategies~\cite{prates2019learning,kool2019attention}, which are similar to our work.
Kool et al.~\cite{kool2019attention} combined graph attention network with reinforcement learning for solving routing problems. Their approach maintains a hidden status of currently ordered nodes and outputs the next one. Other approaches may explore rendering in different scenarios, such as text ranking using CNNs~\cite{severyn2015learning}, and document and image tretrieval~\cite{rahangdale2019deep,guo2023rankdnn}, item ranking~\cite{li2022learning}, and heterogeneous data ordering~\cite{li2023s2phere}.

{\bf Connections and comparisons.} 
Our approach differs from the ordering techniques in visualization, as it aims to provide a quality metric-agnostic solver to optimize the order of data items. Therefore, instead of identifying an ideal metric that enhances human perception, our approach focuses on how to optimize the order given any quality metric. This allows visualization experts to focus on the criteria that lead to an ideal order without spending additional effort to design a solver for achieving those criteria.

In comparison to the learning-based techniques for ordering, such as~\cite{prates2019learning, kool2019attention}, our approach emphasizes the generalizability across metrics, data distributions, and data scales.
This is particularly important for visualization systems: the ordering criteria may vary across applications, and the data points involved in the interaction are likely to exhibit different distributions, as shown in \cref{fig:challenges}.
The existing learning-to-rank approaches may be less effective in visualization scenarios, as they often assume the same distribution for point sets being ranked.
In contrast, our framework is equipped with a novel repositioning module that allows the network to focus on structural information that can be transferred between distributions.

\section{Versatile Ordering Network}
We design versatile ordering network (VON) for learning to order data items with an arbitrary quality metric. Instead of designing a specialized algorithm to optimize for the quality metric, VON uses the quality metric to evaluate different orders of sample points, so that it can automatically learn effective ordering strategy for that specific metric.

\begin{figure}[thbp]
\begin{center}
    \includegraphics[width=0.9\linewidth]{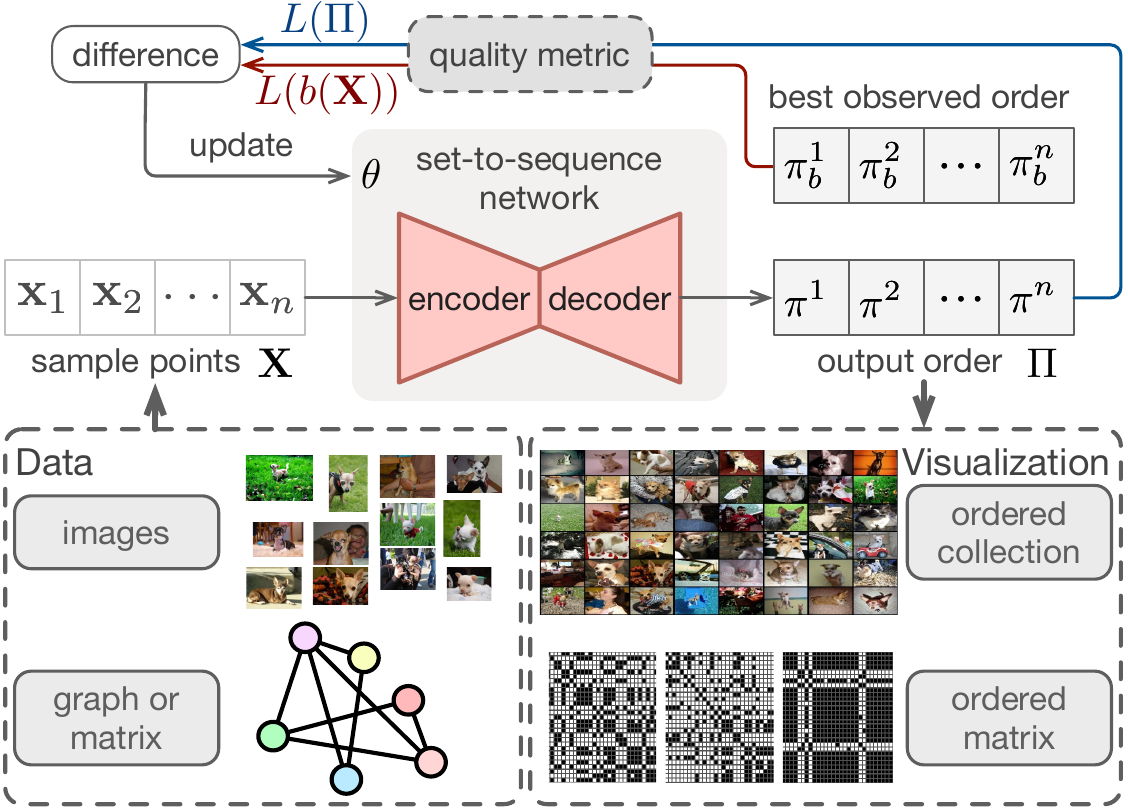}
\end{center}
    \caption{
    An overview of VON workflow. The upper diagram illustrates the overarching framework and training process of VON, and the lower shows example data and ordering applications in visualization. 
    }
    \label{fig:overview}
\end{figure}

\textbf{Overview.} The overarching framework is shown in \cref{fig:overview}. The data are fed into a set-to-sequence network, that converts input sample points into an output order. The output order is evaluated by the quality metric and compared against the best-observed order. The difference in order quality is computed and used to update the network. This reinforcement learning procedure forces the network to improve itself and produce better order quality. Please refer to \cref{sec:background} for details.

The network uses an encoder-decoder architecture. The encoder converts the data points from their original space to a latent space for ordering. The decoder takes the points in the latent space, and outputs the points one by one. At each step, the decoder computes the context of the current ordering status and selects one unordered point for output. Please refer to \cref{sec:conceptual} for details. In this section, we will start with the notation and technical background. Then, we will introduce an abstract framework and explain how the conceptual modules are realized.

\subsection{Notation and Background}
\label{sec:background}
This section introduces the notations and briefly explains the greedy rollout strategy, which is the core of our optimization framework. For other key techniques, such as the attention, multi-head attention (MHA), and the detailed optimization procedure, please refer to Appendix A.

\textbf{Notation.} As shown in \cref{tab:notation}, we denote the input set of data points as $\setof{\point_i}$ and an individual point as $\point_i$. Similarly, we denote the set of hidden encodings at a layer as $\setof{\hidden_i}$ and the encoding of an individual point as $\hidden_i$. We denote the order of a sequence as $\Pi=\setof{\pi^1, \pi^2, \cdots, \pi^n}$, where each element $\pi^t$ indicates the $t$-th data point in the ordered sequence. 
In general, we use the superscript to indicate the ordering step, and the subscripts to indicate specific instances. For subscripts, we use $i$ and $j$ to denote indices of points, $o$ to denote all ordered points, and $c$ to denote the ordering context (considering the ordering front and all ordered points).
We use square brackets to denote the concatenation of multiple vectors. For example, $[\hidden_i, \hidden_j]$ indicates the concatenation of $\hidden_i$ and $\hidden_j$. 

\renewcommand{\arraystretch}{1.7}
\begin{table}[!htbp]
\caption{Notation used in this paper. The superscripts correspond to the ordering step, and the subscripts correspond to specific instances.}
\label{tab:notation}
\scriptsize
\centering
\begin{tabular}{r|l}
\toprule
notation & meaning \\
\midrule
$i$, $j$ & indices of data points \\
$\point_i$, $\point_j$ & input data points\\ 
$\hidden_i$, $\hidden_j$ & hidden encodings of data points\\ 
$\bar{\hidden}$ & the hidden encoding for all data points being ordered\\
$t$ & the ordering step where $t$ points are already ordered\\
$\hidden_o^t$ & the hidden encoding of the already ordered points at step $t$\\
$\hidden_c^t$ & the hidden encoding of the current ordering context at step $t$\\
$\pi^t$ & the output data point at step $t$\\
$[\hidden_i,\hidden_j]$ & the concatenation of two hidden encodings $\hidden_i$ and $\hidden_j$\\
$\att(\hidden_i,\hidden_j)$ & the attention between hidden encodings $\hidden_i$ and $\hidden_j$\\
\midrule               
\bottomrule
\end{tabular}
\end{table}
\renewcommand{\arraystretch}{1}

\textbf{Greedy rollout strategy.} At the core of our learning procedure is the greedy rollout strategy in reinforcement learning~\cite{williams1992simple}. This strategy simulates the greedy selection of most likely data points through a sequence of steps, so that the model can efficiently learn the optimal solution from multiple greedy actions. 
Our optimization minimizes the expected loss by comparing the current model with the best-performing model, which is regularly updated to ensure consistent performance. To stabilize training, we freeze the baseline model's parameters for a certain number of epochs, similar to DQN~\cite{mnih2015human}. Please refer to Appendix A for details.

\begin{figure*}[!htbp]
    \centerline{\includegraphics[width=0.85\linewidth]{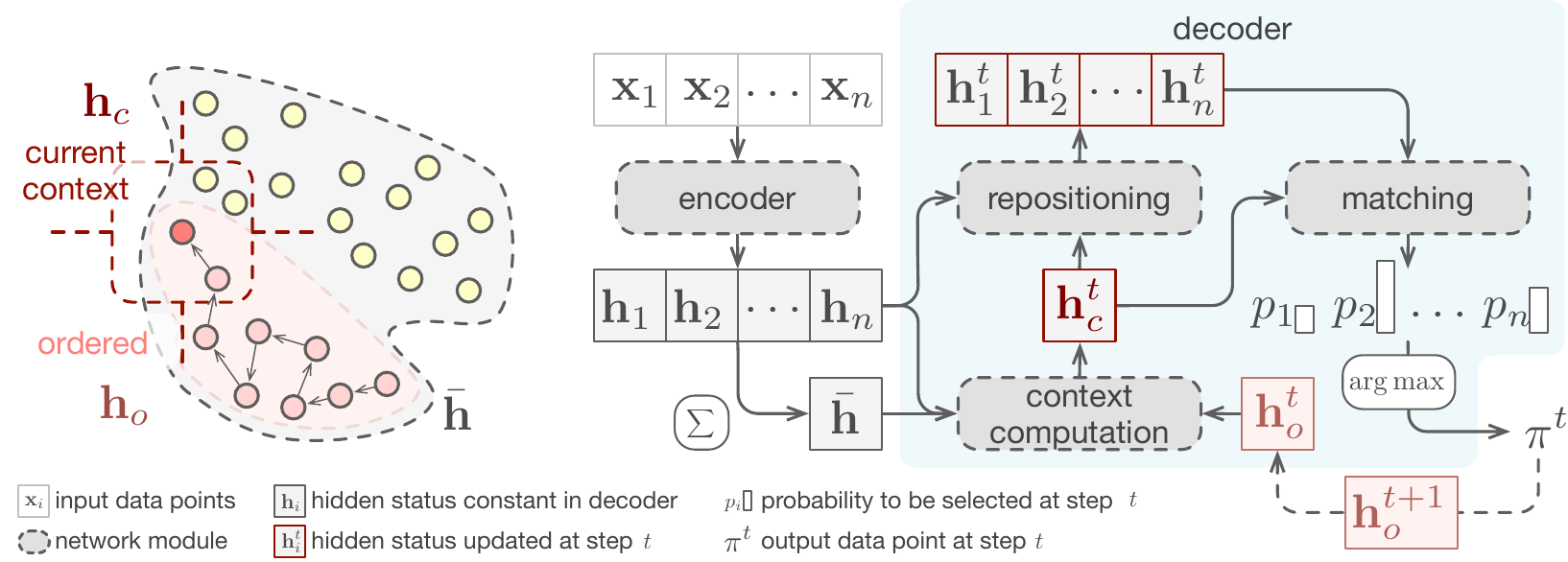}}
    \caption{The conceptual framework of VON. The left part shows the information considered in determining the next data point. The right part shows the data flow and the conceptual network modules. Note that the conceptual modules can be realized with various implementations.}
    \label{fig:framework}
\end{figure*}

\subsection{Conceptual framework}
\label{sec:conceptual}
VON produces the order by sequentially outputting the data points, one at a step. As shown on the left of \cref{fig:framework}, VON considers both the overall structure of all input data points (gray area) and the structure of the previously ordered points (light red area). In particular, VON emphasizes the most recent order point (the red circle) in the ordered nodes. By stitching these three kinds of information, VON forms the context of the current ordering status (the dashed rectangle centering at the red circle). Finally, given the current context, VON determines the next data point to output. 

The conceptual framework is illustrated on the right of \cref{fig:framework}. Note that the detailed network structure is omitted in this illustration, as we would like to focus on the data flow and the goal of each module. 
VON consists of an encoder and a decoder. The \textbf{encoder} transforms the input data points $\setof{\point_i}$ into high-dimensional latent vectors $\setof{\hidden_i}$ for ordering. The overall structure of points $\bar{\hidden}$ is obtained through a common practice by averaging the latent vectors of individual points~\cite{Mikolov2013Distributed}.

The \textbf{decoder} takes in the latent vectors of individual points $\setof{\hidden_i}$ and the overall description $\bar{\hidden}$ and output one point at a step. At each step $t$, the decoder executes the following modules to identify the next point to output:

\begin{myitemize}
\item The \emph{context computation} module updates the ordering context based on the information of the already ordered points $\hidden_o^t$, together with the individual points $\setof{\hidden_i}$ and the overall structure $\bar{\hidden}$. The context computation merges these pieces of information into a single vector $\hidden_c^t$, describing the entire context to pick the next point.

\item The \emph{repositioning} module repositions the latent vector of each point $\hidden_i$ with respect to the current context $\hidden_c^t$. This can be considered as a learnable operator that normalizes the point coordinates based on the current context, as indicated by the red dashed rectangle in \cref{fig:framework}. It allows similar ordering strategies to be applied to points at different locations and scales.

\item The \emph{matching} module evaluates the reposition points $\hidden_i^t$ given the context $\hidden_c^t$. Each point is assigned a probability $p_i$ of being the next point in the sequence. The point with the largest probability will be the output $\pi^t$ at step $t$. 

\end{myitemize}

\noindent At the end of step $t$, the hidden encoding of ordered points $\hidden_o^{t+1}$ will be updated accordingly. By repeating this procedure, the decoder will produce the entire sequence. 
After the entire sequence is generated, the quality metric will be applied to evaluate the sequence and guide the optimization of VON. This forces the VON to look forward and avoid greedy solutions, as the reward is evaluated on the complete sequences instead of at each step. In the following sections, we will discuss the potential realizations of each module. 

\begin{figure}[!htbp]
    \centerline{\includegraphics[width=0.9\linewidth]{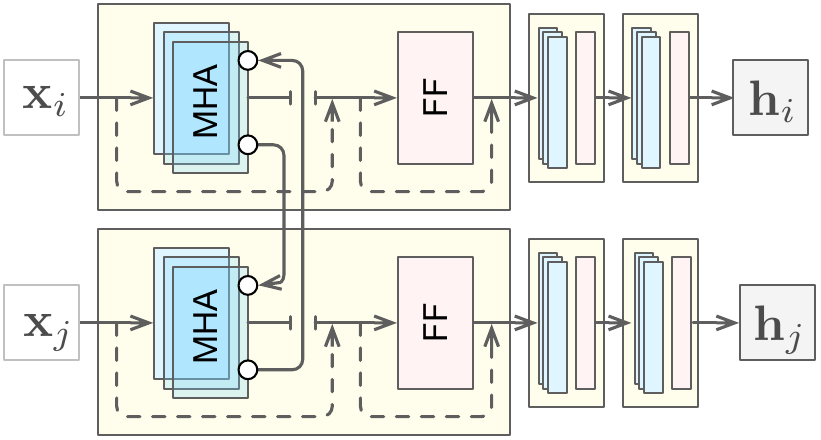}}
    \caption{The encoder network structure. Each attention layer (\attlayer) consists of a multi-head attention layer (MHA \mhalayer) and a feed-forward layer (FF \fflayer). Data points interact with each other through the MHA layer. The dashed arrows indicate skip connections.}
    \label{fig:encoder}
\end{figure}

\subsection{Encoder}
The encoder should transform the input space into a latent space that is appropriate for ordering. As ordering multiple data points should fully consider their relationships, we use the attention mechanism from the transformer~\cite{Ashish2017Attention} to capture the interactions among points. As shown in \cref{fig:encoder}, the encoder is a stack of multiple attention layers (yellow boxes), where each attention layer consists of a multi-head attention (MHA, blue box) layer and a feed-forward (FF, red box) layer. 

Note that, unlike the original transformer~\cite{Ashish2017Attention}, we do not incorporate the positional encoding, as the input point set is expected to be unordered. We should also note that the conceptual encoder can be realized by any other approaches that consider the joint structure formed by the data points as well. The alternative choices include the graph convolutional network and various kinds of message-passing networks. In addition, while VON expects the coordinates of points as the input, it may still handle the data points without intrinsic coordinates. Please refer to \cref{sec:prepare_data} for details.

\subsection{Decoder}
\label{sec:decoder}
At each step, the decoder computes the context of the current ordering status, repositions all points to this context, and matches up the points with the context to identify the next point in the ordered sequence. In this section, we will explain the realizations of the conceptual modules, i.e., context computation, repositioning, and matching modules.

\textbf{Context computation.} 
The context computation aggregates information of the overall structure $\bar{\hidden}$, individual points $\setof{\hidden_i}$, and the already order points $\hidden_o^t$ at step $t$. Similar to the encoder, we use the attention mechanism for incorporating the interactions among multiple entities:

\begin{align}
    \hidden_c^t = \sum_{i} \att_c ([\bar{\hidden},\hidden_o^t], \hidden_i) \cdot \val_c(\hidden_i),
\end{align}

\noindent where $\att_c$ and $\val_c$ form the attention module in context computation. 

This module updates the context using both the ordered and unordered data points, for fully considering the structure formed by all points and coping with various quality metrics (\textbf{C1}). 
This is necessary for quality metrics involving pairwise computation (e.g., the stress in 1D stress majorization). However, for the quality metrics where only consecutive points in the resulting sequence are evaluated (e.g., path length in TSP), the ordered points may be ignored when updating the context. 
In addition, by using all points to update the context, we find that it is safe to concatenate the first point and the most recent point for representing the ordered nodes, i.e., $\hidden_o^t=[\hidden_{\pi_1},\hidden_{\pi_{t-1}}]$.

\textbf{Repositioning.}
The repositioning module aligns the individual points to the current ordering context $\hidden_o^t$. Conceptually, this can be expressed as a function that given the latent vector of a point $\hidden_i$ and the context $\hidden_o^t$, produces the corresponding repositioned vector $\hidden_i^\prime$. While we rely on the attention mechanism to aggregate information in most of the modules, we examine different realizations for this module, as the performance of different network structures is not fully known for repositioning. Specifically, we experiment with the following structures:

First, an \emph{attention-based} realization. This realization attaches the latent representation to the ordered points using two attention layers:

\begin{align}
    \hidden_i'=\att_1(\hidden_o^t, \hidden_c^t) \cdot \att_2(\hidden_o^t, \hidden_i) \cdot \hidden_o^t + \hidden_i.
\label{eq:vona}
\end{align}

\noindent In this process, we consider two major factors in weighting the ordered points $\hidden_o^t$: what are the connections between the current ordering context and the already points (i.e., $\att_1(\hidden_o^t, \hidden_c^t)$), and what are the relationships between each data point and the ordered ones (i.e., $\att_2(\hidden_o^t, \hidden_i)$). 

Second, an \emph{multi-layer perceptron (MLP)-based} realization. This realization can be expressed formally as:

\begin{align}
    \begin{aligned}
        \hidden_i' &= \mlp ([\hidden_c^t, \hidden_i]). 
    \end{aligned}
\label{eq:vonm}
\end{align}

\noindent This means that we will transform the latent $\hidden_i$ by jointly considering the context $\hidden_c^t$. Note that MLP may not be used solely to realize other modules, as it does not scale well with varying numbers of input data points. For repositioning, we mainly consider the interactions between individual points and the context. Therefore, MLP can be sufficient as it blends the different dimensions in the two vectors and allows them to fully interact with each other for capturing both local and global information. In our implementation, we use a two-layer MLP with ReLU activation. 

Third, a \emph{convolution-based} realization. The convolutional module transforms the latent representations with the following formula:

\begin{align}
    \begin{aligned}
        \hidden_i' &= \conv([\hidden_c^t, \hidden_i]) = \sum_{k} \weight * [\hidden_c^t, \hidden_i]_k + b,
    \end{aligned}
\label{eq:vonc}
\end{align}

\noindent where $k$ represents a channel in convolution and ``$*$'' is the valid cross-correlation operator. The convolution-based realization is similar to the MLP-based one, as they both concatenate the two vectors so that different dimensions can fully interact. Instead of stacking multiple layers of networks to approximate complex transformations, the convolutional module aggregates information collected from multiple channels. The effectiveness of these three variations will be examined in our experiments. Please refer to \cref{sec:eval} for details.

\textbf{Matching.} The matching module produces a probability for each point, indicating how likely will the point be selected to be the next output. To incorporate the relationships among data points, we use two MHA layers for this module. We explicitly removed the previously ordered points from being selected by setting their probability to negative infinite. Note that the matching module does not evaluate the points using the quality metric explicitly. Instead, it learns the matching criteria guided by the metric as the loss function.

\subsection{Data preparation}
\label{sec:prepare_data}
The input of VON is a set of data points of the same number of dimensions. For points in Euclidean space (i.e., 2D or 3D point clouds), it is straightforward to use their coordinates as input. For points without intrinsic coordinates, the input may be derived with the following strategies: 

\textbf{Distance-based initialization}. If a distance metric is available, we may compute the distance matrix to form the input. For example, for graph data, the graph distance between nodes may be used. In this way, each row/column corresponding to a data point can be used as its coordinates. An alternative choice is to use embedding methods (e.g., t-SNE) to embed the points into an Euclidean space based on the distance metric. This strategy may be applied to graph data and axis reordering, where similarities between coordinates are often defined. Note that  VON can tolerate less ideal input as its encoder will reshape the input space. Therefore, the input does not need to be perfect, as long as the data points are distinguishable.

\textbf{Deep representation-based initialization}. We may use deep representation approaches to embed the data points into latent space. For example, we may use an autoencoder for images, a graph representation network for graphs, and multi-layer perceptrons (MLPs) for data points with attributes. As pretraining becomes a common technique, latent embeddings are often available for various kinds of data. In our experiment, we use the deep representations of three image datasets (i.e., CIFAR-10, FashionMINST, and ImageNet) and one scholarly dataset (i.e., CORA).

\section{Evaluation}
\label{sec:eval}
In this section, we evaluate the performance of VON under different experimental settings. 
We first use an ablation study to determine the best instantiation of VON and examine the effectiveness of each module. The best configuration of VON will be used for the remaining experiments. For these experiments, we consider two scenarios:
ordering randomly selected points and ordering fixed data points. The first scenario is suitable for ordering data points that may be repeatedly sampled from the same distribution multiple times, e.g., ordering images interactively selected from a collection. The second scenario mimics the applications such as matrix reordering and axis ordering. 
For the first scenario, we consider eight baselines, six metrics, six datasets, four sampling strategies, and three sampling sizes, leading to \textbf{1059 sets} of experiments for a comprehensive evaluation.
For the second scenario, we compare VON with baselines in axes reordering and matrix reordering. For axes reordering, we consider four baselines and four datasets; and, for matrix reordering, we consider eight baselines, two dynamic datasets, and five embedding approaches. Please refer to \cref{sec:setting} and the Appendix for detailed settings.

\subsection{Experimental settings}
\label{sec:setting}
This section lists the datasets, methods being compared, and quality metrics used in our experiment. Note that for datasets without coordinates, embedding approaches are used. The embedding approaches being studied are also listed. More details can be found in the Appendix. In this section, we explain our sampling strategies, as they are closely related to our experiment.

\textbf{Datasets.} We experiment with twelve datasets, including six datasets with data points in respective latent embedding spaces (i.e., Fashion-MNIST, MNIST, CIFAR-10, ImageNet, CORA, and DBLP), four high dimensional datasets (i.e., Coal Disaster, Cars, AAUP, and Census Income) and two dynamic graph datasets (i.e., FLT and SCH).

\textbf{Methods compared.}
We compare VON against three generic approaches (i.e., attention model (AM)~\cite{kool2019attention}, simulated annealing (SA)~\cite{Granville1994Simulated} and nearest neighbor (NN)~\cite{NN}), four specialized methods for images reordering (i.e., IsoMatch~\cite{fried2015isomatch}, Kernelized Sorting (KS)~\cite{quadrianto2008kernelized}, LAS and FLAS~\cite{barthel2023improved}), one specialized method for axes reordering (i.e., random swapping), and four specialized algorithms for matrix reordering (i.e., C-LO-$\delta_I$, U-LO-$\delta_I$, C-BC and U-BC~\cite{vanbeusekom2022simul}).

We choose the baseline algorithms considering their targeted problem structures, usage scenarios, and popularity. For the generic solvers, SA~\cite{dennis2006Simulated} is widely applicable, as it does not assume any specific problem structure; NN~\cite{NN} simulates the most common strategy in ordering by selecting neighboring/similar nodes; and AM~\cite{kool2019attention} allows us to examine performance against the SOTA learning-based approach. For the specialized solvers, SM~\cite{Gansner2005Graph} optimizes the stress, a general function modeling the pairwise relationships between points; and other baselines are selected from typical visualization applications, including image ordering, axis reordering, and matrix reordering. Please see Appendix B for detailed explanations of the baselines and their setup. We believe these approaches cover representative problem structures and usage scenarios.

\textbf{Quality Metrics.}
Various kinds of quality metrics have been proposed to evaluate the quality of ordering results in terms of preserving structures, perception, and aesthetics. In this experiment, we study two quality metrics for the 1D layout problem (i.e., the traveling salesman problem (TSP) distance and the stress in stress majorization), three metrics (i.e., distance preservation quality (DPQ), energy in IsoMatch, and kernelized sorting objective) for images reordering, two metrics (i.e., symmetry and correlation) for axes reordering, and four metrics (i.e., Moran's I, Linear Arrangement (LA), profile (PR), and bandwidth (BW)) for matrix reordering.

\textbf{Embedding methods.} When the coordinates of data points are not available in a dataset (e.g., the SCH dataset), an embedding approach is needed to produce the coordinates. We examine the impact of five embedding approaches with distinct characteristics, namely, principal component analysis (PCA), truncated singular value decomposition (T-SVD), autoencoder (AE), Isomap, and locally linear embedding (LLE).

\textbf{Sampling strategies.} Four sample strategies are used to resemble dynamic behaviors during interaction and examine the performance across scales. The four strategies are: 1) the \textbf{global sampling (`g')} that randomly samples from all points; 2) the \textbf{nearest neighbor sampling (local-`n')} that samples the nearest points of the specified number; 3) the \textbf{radius-based sampling (local-`r')} that randomly samples the points within a given radius of a specific point; and, 4) the \textbf{label-based sampling (local-`l')} that randomly samples points with a specific label. The strategy `g' mimics the behavior of browsing the entire collection, `n' and `r' emulates the behavior of scanning items similar to a selected one, and `l' corresponds to reviewing items of the same category. Based on the findings and experimental settings in~\cite{nguyen2008interactive, han2016tree, al2024improved, johansson2015evaluation, miller2019evaluating, miller1956magical}, three sampling sizes: 50, 100, and 150 are considered for each sampling strategy. Please refer to Appendix B for details.

We further denote a model that is first trained with global samples and then with the local samples as ``gl'', and a model trained with local samples and then global samples as ``lg''. For example, ``50gl'' indicates that the model is trained with 50 global samples and then 50 local samples. We denote a model trained with global and local samples in random orders as ``mix''.  These settings will be useful in discovering the detailed behavior of models in transferring knowledge across scales and distributions. Please note that for each setting, we have the same number of global and local samples. 

\subsection{Ablation study}
In the Ablation study, We first examine the performance of the three variants of VON and the impact of sampling strategies that produce the training data. Then, we study the individual impact of each key module in the best variant. The three variants are VONs based on attention mechanism (\vona), MLP (\vonm), and convolution (\vonc), corresponding to \cref{eq:vona} to \cref{eq:vonc}, respectively.

\begin{table}[!htbp]
  \caption{Ordering performance using Fashion-MNIST dataset and TSP quality metric. The content in the parentheses following a model name denotes the sampling strategy used to train that model. \textbf{I}. Comparing models trained with point sets of 50 points. \textbf{II}. Comparing \vonm models trained with different point sets of different sizes and scales. \textbf{III}. Comparing models with mixed-scale sampling strategies.
  }
  
  \scriptsize%
  \centering%
  \setlength{\tabcolsep}{1.6mm}{
  \begin{tabular}{%
            c|c|%
  	  	*{6}{r}%
  	}
  	\toprule
  	& & \fmdatafg & \fmdatafl & \fmdataag & \fmdataal & \fmdataafg & \fmdataafl   \\
  	\midrule
  	&\vonafg & 608.1 & 349.7 & 920.8 & 639.4 & 742.7 & 826.3 \\
        &\vonafl & 1024.3 & 246.0 & 1312.3 & 448.1 & 609.3 & 693.6 \\
        &\vonmfg &  \textbf{588.3} & 183.3 & \textbf{879.8} & 357.1 & 504.5 & 540.5 \\
        &\vonmfl & 635.9 & \textbf{177.6} & 940.1 & \textbf{343.8} & \textbf{488.6} & \textbf{526.1} \\
        \textbf{I} &\voncfg & 674.4 & 229.3 & 965.7 & 433.9 & 616.5 & 658.9 \\
        &\voncfl & 663.8 & 212.4 & 1030.4 & 408.5 & 566.7 & 605.3 \\
        &\amfg &  616.6 & 427.3 & 1018.7 & 1087.9 & 1759.5 & 2007.4 \\
        &\amfl & 677.8 & 229.1 & 1083.8 & 445.2 & 605.4 & 680.3 \\              
  	\bottomrule
        \midrule
        &\vonmag & \textbf{599.5} & 185.7 & \textbf{954.9} & 357.6 & 515.3 & 563.5  \\
        \textbf{II} &\vonmal & 635.5 & \textbf{182.3} & 971.5 & \textbf{320.9} & 531.0 & 563.9  \\
        &\vonmafg & 751.0 & 208.3 & 1113.9 & 400.7 & \textbf{487.9} & 541.6  \\
        &\vonmafl & 698.8 & 193.5 & 1220.7 & 386.2 & 581.0 & \textbf{538.2}  \\
                   
  	\bottomrule
        \midrule 
        &\vonmfmix & 614.0 & 217.8 & \textbf{919.8} & 402.6 & 545.9 & \textbf{590.9} \\
        &\vonmfgl & \textbf{609.5} & 199.6 & 941.6 &  379.2 & \textbf{540.4} & \textbf{590.9} \\
        \textbf{III} &\vonmflg & 694.0 & \textbf{180.7} & 1077.3 & \textbf{357.2} & 558.4 & 592.9 \\
        &\amfmix & 693.4 & 462.0 & {1003.1} & 1003.6 & 1010.2 & 1474.4 \\
        &\amfgl & 689.2 & 402.3 & 1128.3 & 1057.2 & 1403.7 & 1628.2 \\
        &\amflg & 2150.1 & 260.3 & 2373.9 & {620.1} & {951.7} & {1025.0}\\   
        \midrule
        \bottomrule
  	     
  \end{tabular}}
  \label{tab:FM-ablation}
\end{table}

\textbf{Identifying the best VON variant.} \cref{tab:FM-ablation}.I shows the results of \vona, \vonm, and \vonc, together with the results of AM for reference. Each model is trained with two sets of training data (i.e., $\fmdatafg$ and $\fmdatafl$). We do not create training data with multiple sampling strategies to reduce the undesired impact of sampling strategies. 

We can see that \vonm outperforms the other models in every case, especially in terms of its transferability across scales. We find that $\vonmfl$ (trained on local data) outperforms the other models on the global data $\fmdataag$, and $\vonmfg$ (trained on global data) delivers smaller distances than the other models on the local data $\fmdataal$. 

In contrast, $\vona$ and AM show a clear performance decline on testing data at different scales. For example, $\vonafl$ results in large distances in $\fmdatafg$ and $\fmdataag$ tests. This indicates that the attention mechanism is less effective in repositioning the points with linear combination. AM's performance drop is even more notable, as it does not incorporate any mechanism to transfer knowledge across scales. For example, $\amfg$ decreases from 616.6 in $\fmdatafg$ to 1759.5 in $\fmdataafg$. Therefore, \emph{in the remaining experiments, we will use \vonm as our model and compare it with other models}.

Additionally, we discuss the sampling strategy in detail and examine the impact of each module. Please refer to the Appendix C.I for details.

\begin{figure*}[!htb]
  \centering
  \includegraphics[width=\linewidth]{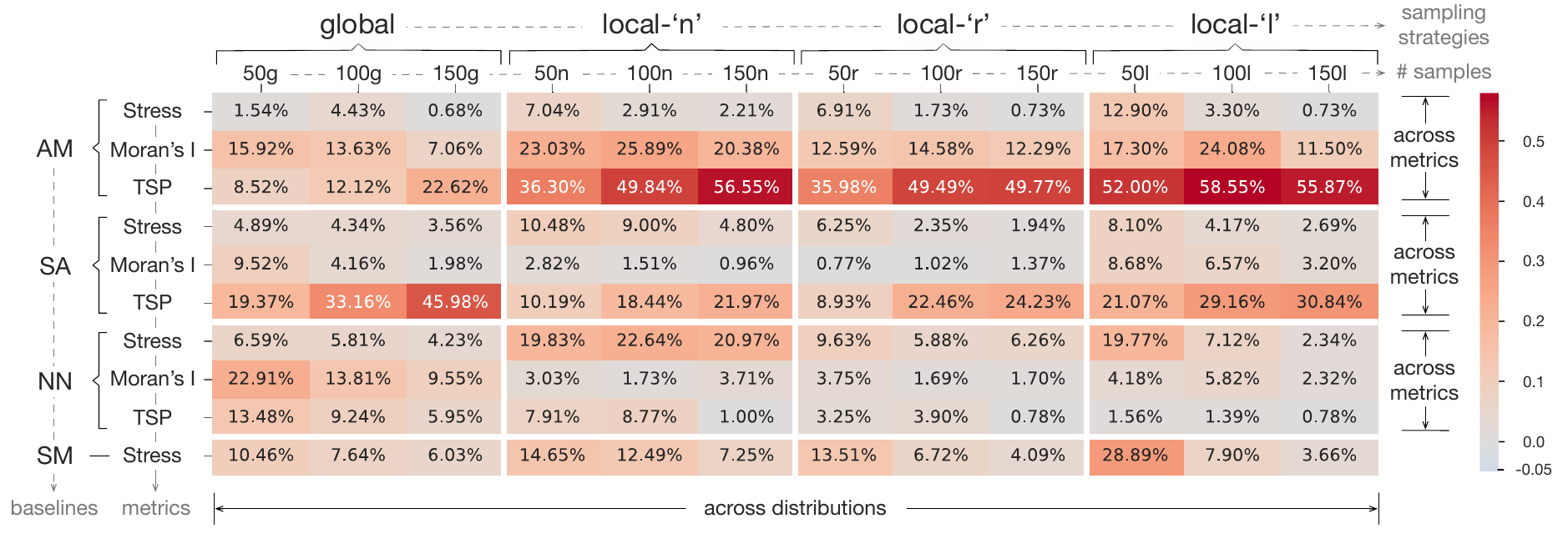}
  \caption{Average performance improvement of VON over AM, SA, NN, and SM for three metrics, using the MNIST, Fashion-MNIST, CIFAR-10, DBLP, Cora, and ImageNet datasets. All dataset is experimented with three sampling strategies: global, nearest neighbor (local-`n'), and radius-based (local-`r'). The first three datasets with labels are further examined with the label-based (local-`l') sampling. The experiment of each dataset is conducted over 100 sets of samples.}
  \label{fig:dynamic-all}
\end{figure*}

\begin{figure}[!htb]
  \centering
  \includegraphics[width=\linewidth]{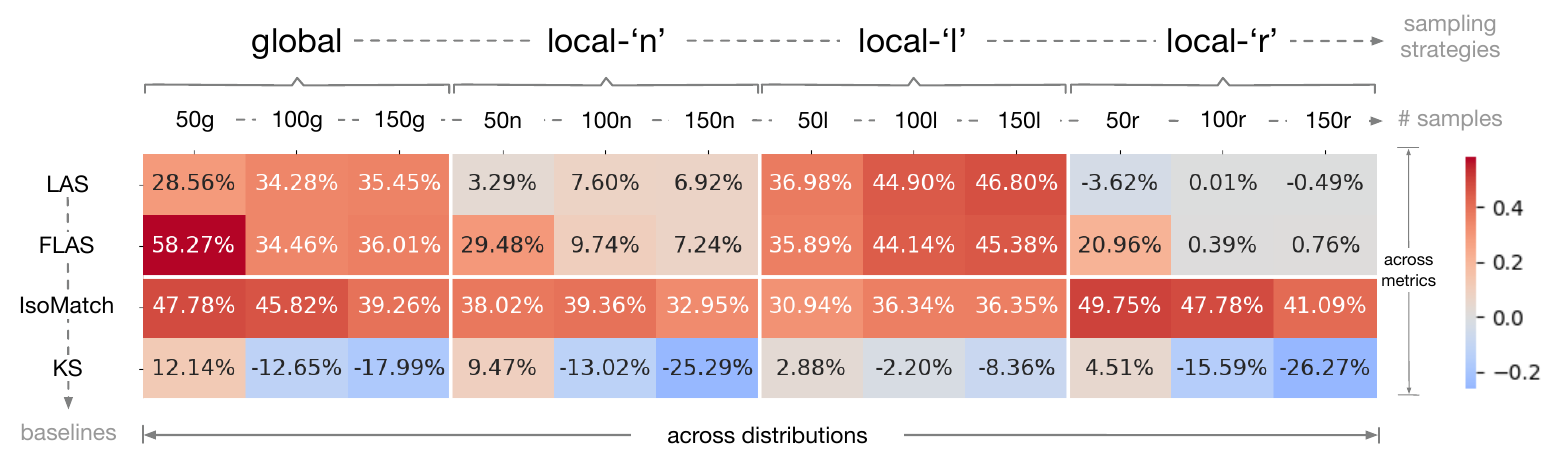}
  \caption{Average performance improvement of VON over LAS, FLAS, IsoMatch, and KS using the MNIST, Fashion-MNIST, CIFAR-10, and ImageNet datasets. Each dataset is experimented with four sampling strategies: global, nearest neighbor (local-`n'), radius-based (local-`r'), and label-based (local-`l') sampling. The experiment of each dataset is conducted on 100 sets of samples. Each specialized method uses its own applicable metric: LAS/FLAS uses DPQ, IsoMatch uses Energy Value, and KS uses its own objective function.}
  \label{fig:dynamic-spc}
\end{figure}

\subsection{Performance for Ordering in Dynamic Scenario}
We conduct extensive experiments to examine the performance of the best variant of VON (\vonm) in a dynamic scenario, where the point sets are sampled from a large collection for visualization. Specifically, for general ordering task using both the image and scholarly data, we compare VON with eight baselines with different characteristics: AM (learning-based), SA (naturally inspired), NN (heuristic), SM (specialized for the stress metric), using three quality metrics with different problem structures: TSP (linear route), stress (complete graph), and Moran's I (distance matrix). Additionally, for image ordering, we further consider four baselines (LAS, FLAS, IsoMatch, and Kernelized Sorting) with their respective objectives as optimization goals. We perform 1059 sets of experiments on six datasets. Each set of experiments is performed on 100 groups of testing points to deliver stable results.

\begin{figure*}[!htb]
  \centering
  \includegraphics[width=\linewidth]{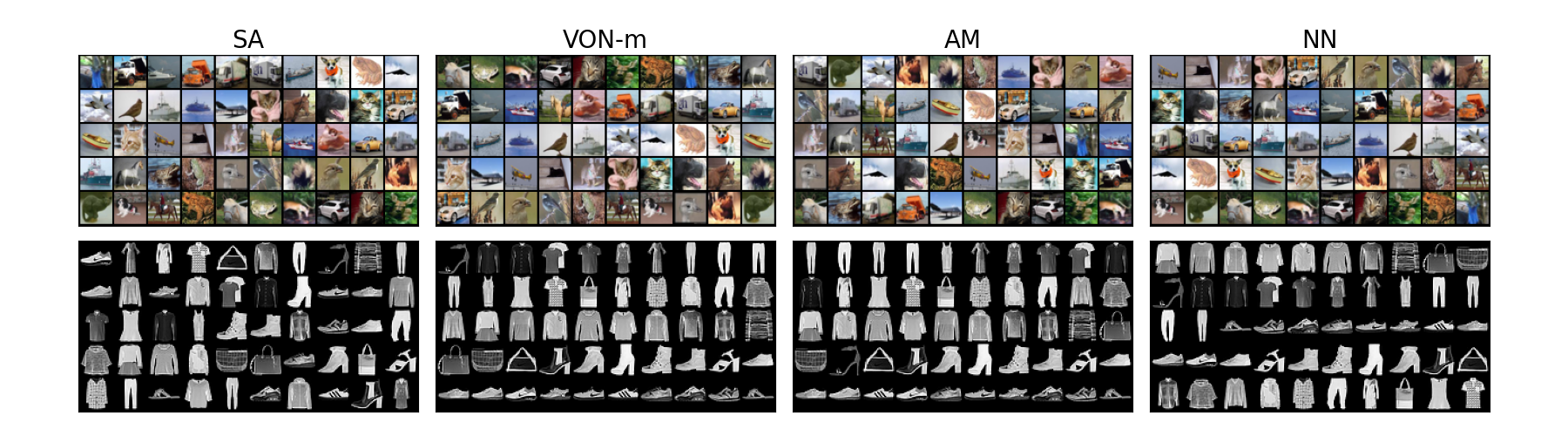}
  \caption{Comparison of image ordering results using TSP as the quality metric. The first row shows the results using CIFAR-10 dataset and the second row shows those using Fashion-MNIST. The four columns show the results of SA, \vonm, AM, and NN from left to right, respectively.}
  \label{fig:FM_CIFAR_show}
\end{figure*}

\subsubsection{Quantitative performance}
\label{sec:dynamic-quantitative}

In this section, we first discuss the general performance of \vonm using all six datasets (as shown in \cref{fig:dynamic-all}), and then the image ordering performance using the image datasets against specialized algorithms (as shown in \cref{fig:dynamic-spc}). 
Note that we use performance improvement percentages for comparison instead of the original metric values, because the original values may vary substantially in magnitude.

\textbf{Overall performance using all datasets.} For the general ordering scenario, \vonm outperforms the baselines in all the 120 pairs of comparisons, with all the improvement percentages being positive in \cref{fig:dynamic-all}. Even for individual datasets, \vonm improves the baselines in 629/630 sets of experiments.
The only exception is found in the Fashion-MNIST dataset, where the TSP distance of NN in 150r sampling is 0.29\% smaller than \vonm, as shown in \cref{fig:dynamic-individual} in Appendix C.
This shows that \vonm delivers stable performance across datasets, metrics, and sampling strategies in general ordering tasks.

\textbf{Improvement for different metrics.} We find that the pattern of performance improvement for different metrics varies across baselines. 
AM and SA share a similar pattern, where the improvement in TSP is large and the improvement in stress is small. Since stress considers all pairwise distances, we expect it to be less sensitive to the order of points, leading to smaller improvement. 
However, we observe an opposite trend for NN. This indicates that the nearest neighbor heuristic excels in planning linear routes, but falls short in providing a global picture needed to match the complete graph structure of stress.
Note that stress majorization produces the largest stress as it does not explicitly take the order into consideration. 

\textbf{Improvement for different sampling strategies.} For the learning-based approach (AM), the performance gaps in global samples are larger than the gaps in local samples for all three metrics. This indicates the lack of generalizability across scales in AM. For the conventional approaches (SA and NN) without the learning procedure, their performance across scales relies on the metric. For Moran's I and TSP, the improvement over these two approaches in global samples is larger than that in local samples; for stress, the trend is the opposite. Since SA and NN do not depend on data distribution, this may indicate that it is easier to solve Moran's I and TSP locally, while it is easier to solve stress globally.

\textbf{Improvement for different sample sizes.} Because \vonm is trained on $\fmdatafmix$, we expect a decrease in its performance when the number of samples increases from 50 to 100 and 150. This may further lead to a decrease in performance improvement. For the learning-based approaches (AM and \vonm), their performance becomes similar when less knowledge can be leveraged from the training data. For the conventional approaches (SA and NN), as their performance does not rely on the training data, the performance gap between these approaches and \vonm may shrink as well. In \cref{fig:dynamic-all}, while observing this decrease in most cases, two exceptions (AM for TSP and SA for TSP) are found. While the behavior of AM may still be attributed to the transferability, we suspect that the pairwise swapping strategy of SA may be less effective in solving a linear route problem when the number of points increases. 

\textbf{Performance in image ordering.}
For image ordering, \vonm outperforms 38/48 pairs of comparison against specialized algorithms on their respective metrics, as shown in \cref{fig:dynamic-spc}. It performs better than FLAS and IsoMatch in every setting, and LAS in most settings except for 50r and 150r. Using the KS objective function, \vonm outperforms KS in the four cases with a sampling size of 50, but loses in the other cases. The performance gap becomes larger when the size increases from 100 to 150. As \vonm is trained on the sample size of 50, the performance loss may be due to the difference between the testing data and the training data. We suspect that the ordering strategy should be adjusted when dealing with different distributions for certain metrics. In this case, we recommend collecting training data from users to better align the training and testing data. 

\textbf{Findings.} The extensive experiments demonstrate the strong impact of the complicated connections between the algorithm design, data distributions, and problem structures. This highlights the importance of transferability across scales and metrics, which is unavailable in existing learning-based approaches. For the general ordering tasks, we should note that the performance gap between AM and other approaches mainly originates from the diverse data distributions. For the main distribution (50g) in the training data, AM outperforms the other three baselines in 6 out of 7 cases, except SA for Moran's I. However, when the testing data deviates from the main pattern in training data, AM's performance quickly decreases. In contrast, \emph{VON achieves better generalizability, as evidenced by the performance improvement over generic approaches in every case, and over specialized algorithms in a considerable portion of cases (38/48) for image ordering.}

More importantly, the performance variation across scenarios infers the difficulty in selecting an appropriate solver for a given scenario. This either involves the prohibitive trial-and-error effort to examine multiple ordering techniques or requires the chosen technique to generate better results in all cases. In light of this observation, we emphasize the importance of worst case performance. In our experiments, every baseline may deliver the worst performance among all approaches in certain scenarios, e.g., AM for TSP with local samples, SA for TSP with global samples, and NN in stress with local-`nn' samples. Additionally, when considering the worst cases for individual datasets, every baseline has at least a 50\% performance gap compared to VON, as shown in \cref{fig:dynamic-individual}. In contrast, VON is at least comparable to the baselines in the worst cases for general ordering tasks, and it outperforms the image ordering algorithms in a large portion of cases (38/48). \emph{This renders it to be a reasonable choice in most scenarios without exhaustive examination.} However, the performance gap between \vonm and KS, when the sampling size deviates from training data, also emphasizes that we should keep the training data as close to the real scenario as possible.

\subsubsection{Qualitative case study}

In \cref{fig:FM_CIFAR_show}, we qualitatively compare the ordering results of \vonm and three other approaches using 50 randomly selected images from the CIFAR-10 and Fashion-MNIST datasets, respectively. The quality of the order can be visually assessed from the ordered images, where effective approaches should place similar items close to each other. 

For Fashion-MNIST, the difference in ordering quality is easier to perceive. We find that \vonm, AM, and NN produce better results, clearly dividing clothes, bags, and shoes into three groups; SA is less pronounced, as the items of different categories often appear in a mixed order.

For CIFAR-10, the quality is somewhat difficult to visually evaluate in general methods, as the similarity may be defined in multiple dimensions (e.g., color and content). We can still observe a clear pattern from the results of \vonm and NN. For example, \vonm seems to place images with green background in the first row, images with sky background in the second and third row, and images with animals in the two bottom rows; similarly, NN tends to place images with sky in the second and third row, vehicles in the third row, and animals in the last row. The visual assessment is in line with the quantitative results, where \vonm and NN excel in solving TSP. 

However, we should note that the visual assessment incorporates both the similarity measure and ordering quality. For this paper focusing on generic solver, we still mainly rely on the quantitative results. For more results regarding other metrics, please refer to Appendix C.II.

\begin{table}[!thbp]
  \caption{%
  	Axes reordering performance of \vonm, AM, SA, RS using Coal Disaster, Cars, AAUP, and Census Income datasets. The star plot is assessed with the symmetry quality metric~\cite{peng2004clutter}, and the parallel coordinate is assessed with correlation~\cite{tyagi2023pcexpo}.
  }
  \label{tab:star-pcp}
  \scriptsize%
  \centering%
  \setlength{\tabcolsep}{1.5mm}{
  \begin{tabular}{
  	  c|c|%
  	  	*{4}{c}%
  	}
  	\toprule
    \multicolumn{2}{c|}{} & Coal & \multirow{2}{*}[.15ex]{Cars (7)} & \multirow{2}{*}[.15ex]{AAUP (14)} & Census \\
    \multicolumn{2}{c|}{} & Disaster (5) & & & Income (42)   \\
  	\midrule
        & \vonm  & \textbf{0.0743} & \textbf{0.2166} & \textbf{0.1321} & \textbf{0.2484} \\
        \cmidrule{2-6}
   star & AM  & \textbf{0.0743} & 0.2285 & 0.1492 & 0.2602 \\
   plot & SA  & \textbf{0.0743} & \textbf{0.2166} & 0.1539 & 0.2641\\
        & RS  & \textbf{0.0743} & \textbf{0.2166} & 0.1482 & 0.2603\\
  	\bottomrule
        \midrule
        & \vonm & \textbf{0.0244} & \textbf{0.3593} & 0.4516 & \textbf{0.0789} \\
        \cmidrule{2-6}
parallel & AM & 0.0264 & 0.3714 & 0.4664 & 0.0843 \\
coordinate & SA & \textbf{0.0244} & 0.3749 & \textbf{0.4500} & 0.0808 \\
        & RS & \textbf{0.0244} & 0.3754 & 0.4569 & 0.0839 \\
        \midrule
  	\bottomrule
  \end{tabular}}
\end{table}

\subsection{Performance for Ordering in Static Scenario}
For the static scenario, we study two typical applications: matrix reordering and axes reordering. In these applications, the data points (i.e., axes or matrix rows/columns) do not change dynamically.

\subsubsection{Performance on axes reordering}
\label{sec:axes-reorder}
We compare \vonm against AM, SA, and random swapping~\cite{peng2004clutter} in reordering the axes of star plots and parallel coordinates. We experiment with four datasets that are commonly used in axes reordering ~\cite{peng2004clutter}: the Coal Disaster dataset with 5 attributes, Cars with 7 attributes, AAUP with 14 attributes, and Census Income with 42 attributes.

\textbf{Quantitative result.}
We used symmetry~\cite{peng2004clutter} as the quality metric for star plot axes reordering, and correlation~\cite{tyagi2023pcexpo} for parallel coordinate reordering. Note that we transform the metric for consistency, so that smaller values indicate better performance. 
\cref{tab:star-pcp} shows that, for the datasets with smaller numbers of axes (Coal Disaster and Cars), most approaches can identify the optimal solutions. This leads to identical metric values, i.e., 0.0743 in Coal Disaster, 0.2166 in Cars for star plots, and 0.0244 in Coal Disaster for parallel coordinates. However, when the number of axes increases, the solution space grows factorially and it becomes prohibitive to search the entire space for the optimal solution. For Census Income with 42 axes, we can observe larger gaps between \vonm and other approaches. Overall, \vonm performs the best in 7/8 cases. The only exception is the parallel coordinates reordering using AAUP, where \vonm ranks second (0.4516) right after SA (0.4500).

\begin{figure}[!thb]
\begin{center}
$\begin{array}{c@{\hspace{.1in}}c}
\includegraphics[width=.47\linewidth]{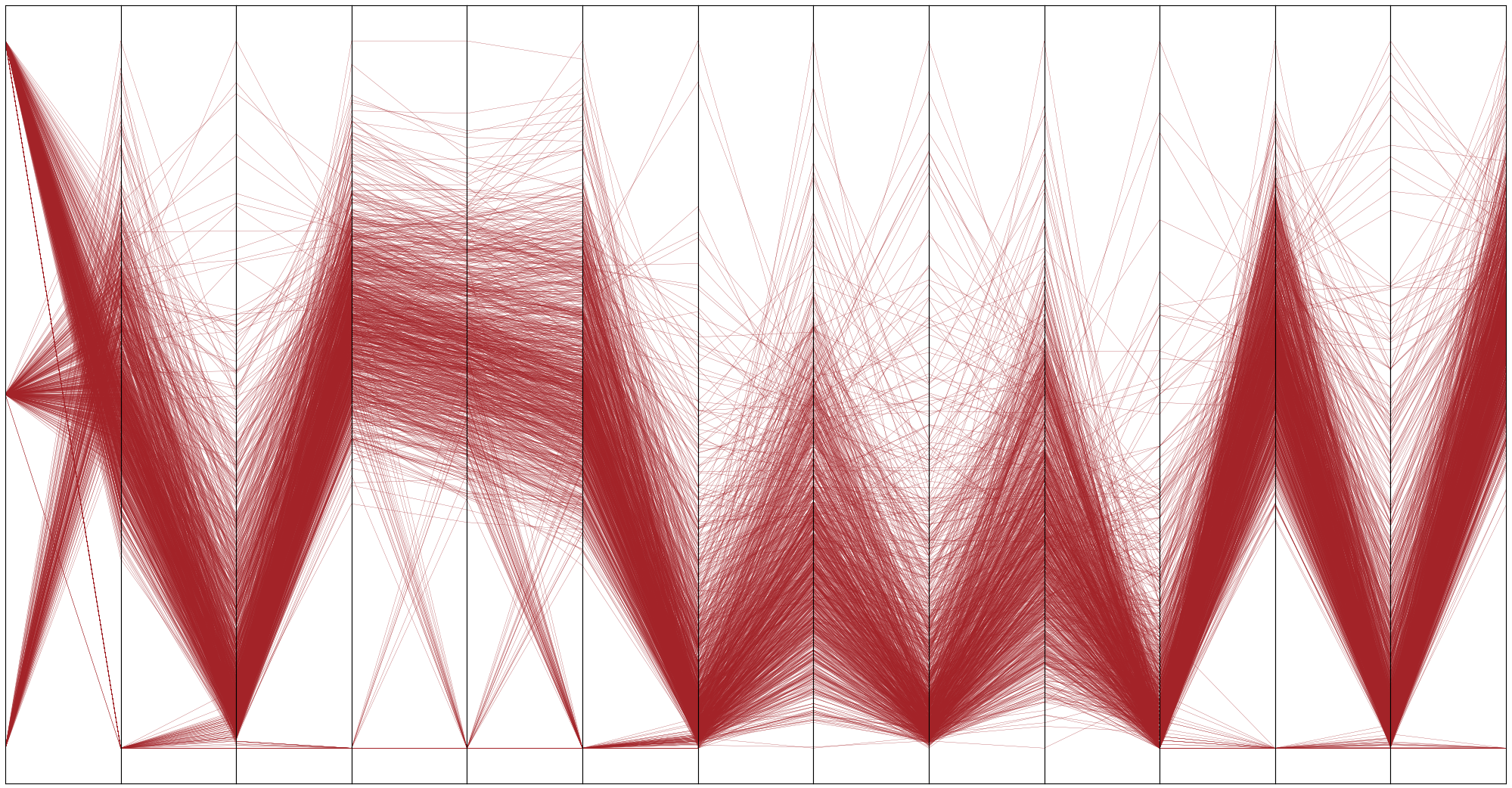}&
\includegraphics[width=.47\linewidth]{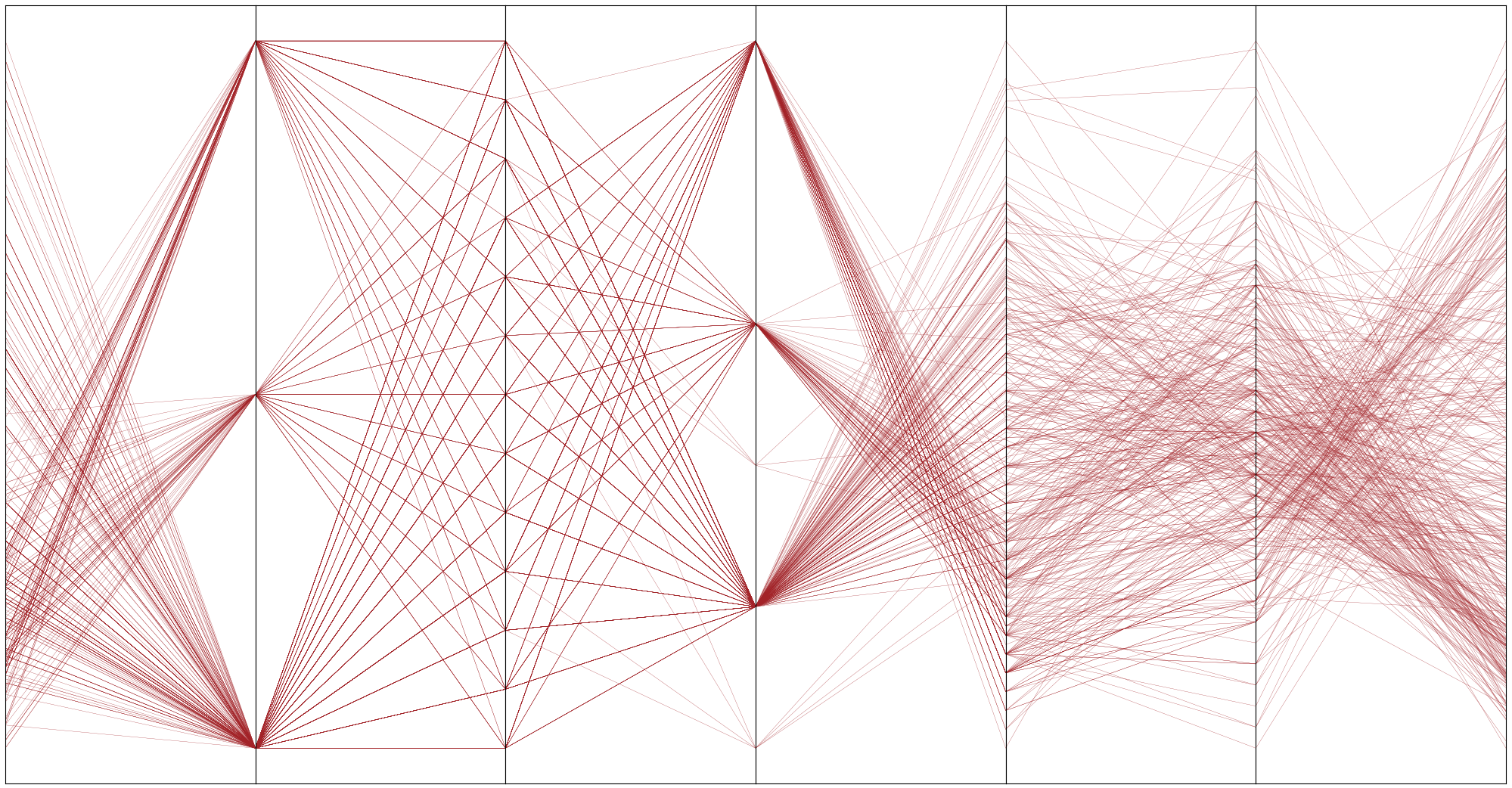}\\
\mbox{\footnotesize VON-m} & \mbox{\footnotesize VON-m}\\[3pt]
\includegraphics[width=.47\linewidth]{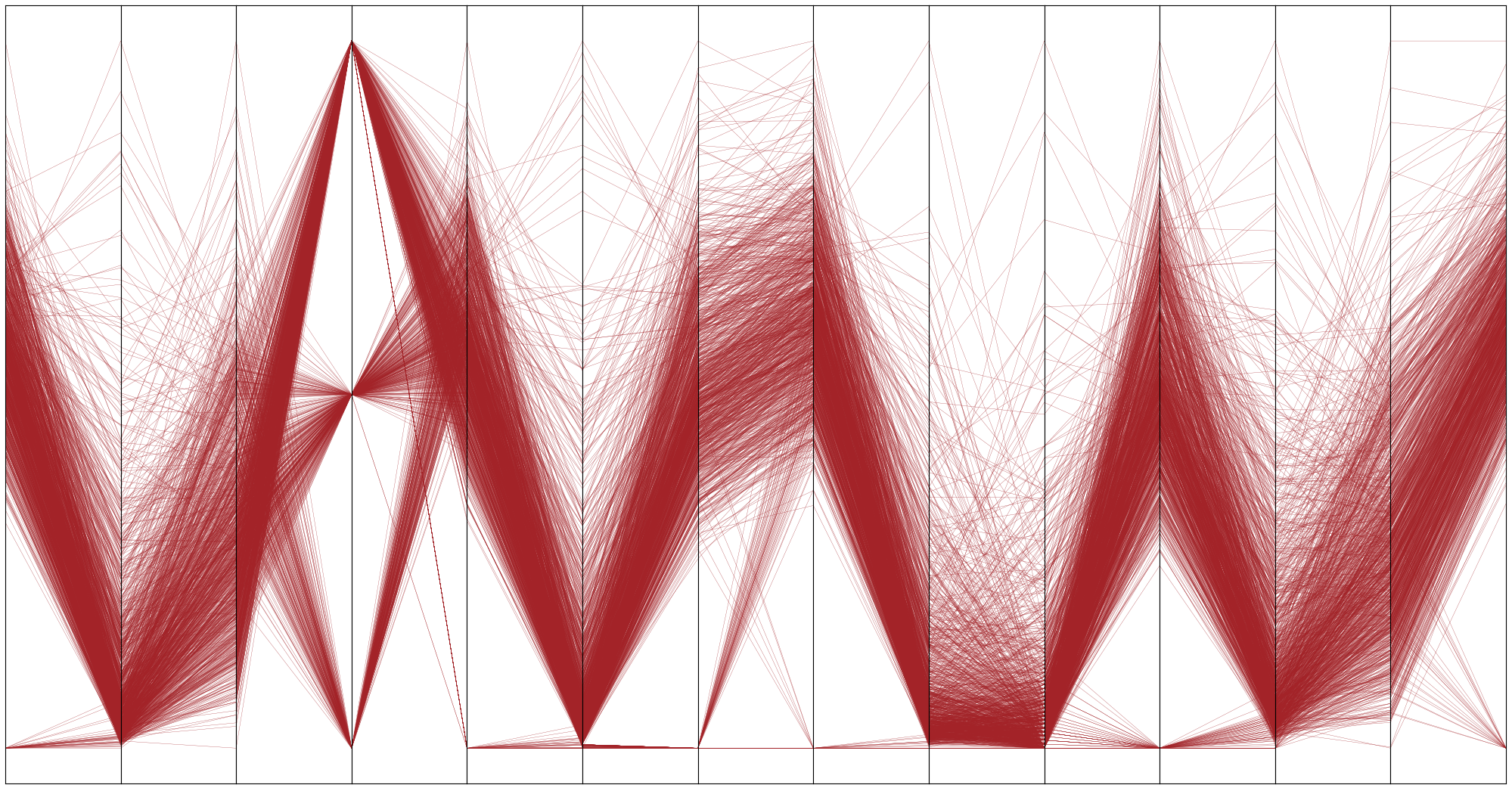}&
\includegraphics[width=.47\linewidth]{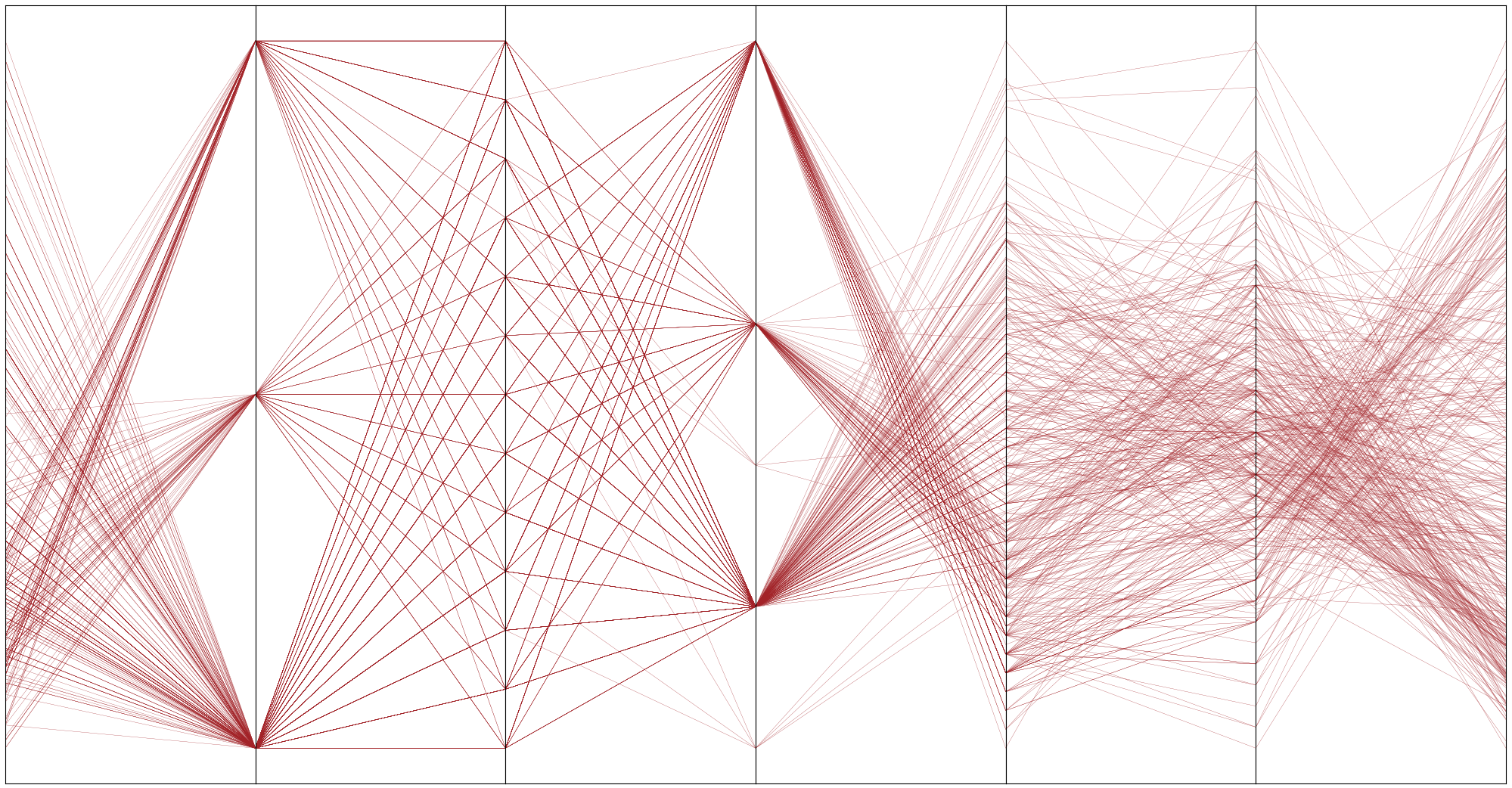}\\
\mbox{\footnotesize AM} & \mbox{\footnotesize AM}\\[3pt]
\includegraphics[width=.47\linewidth]{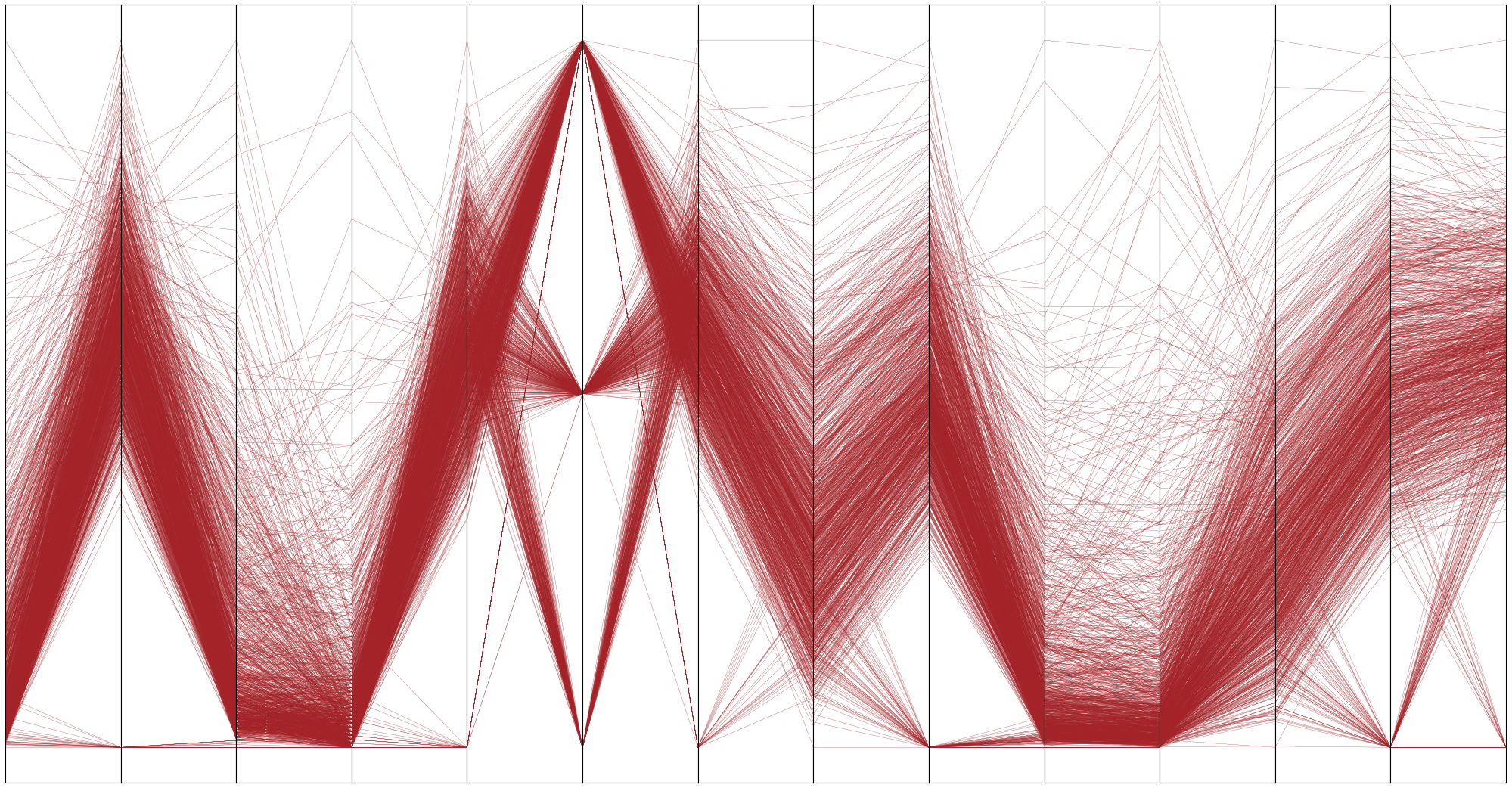}&
\includegraphics[width=.47\linewidth]{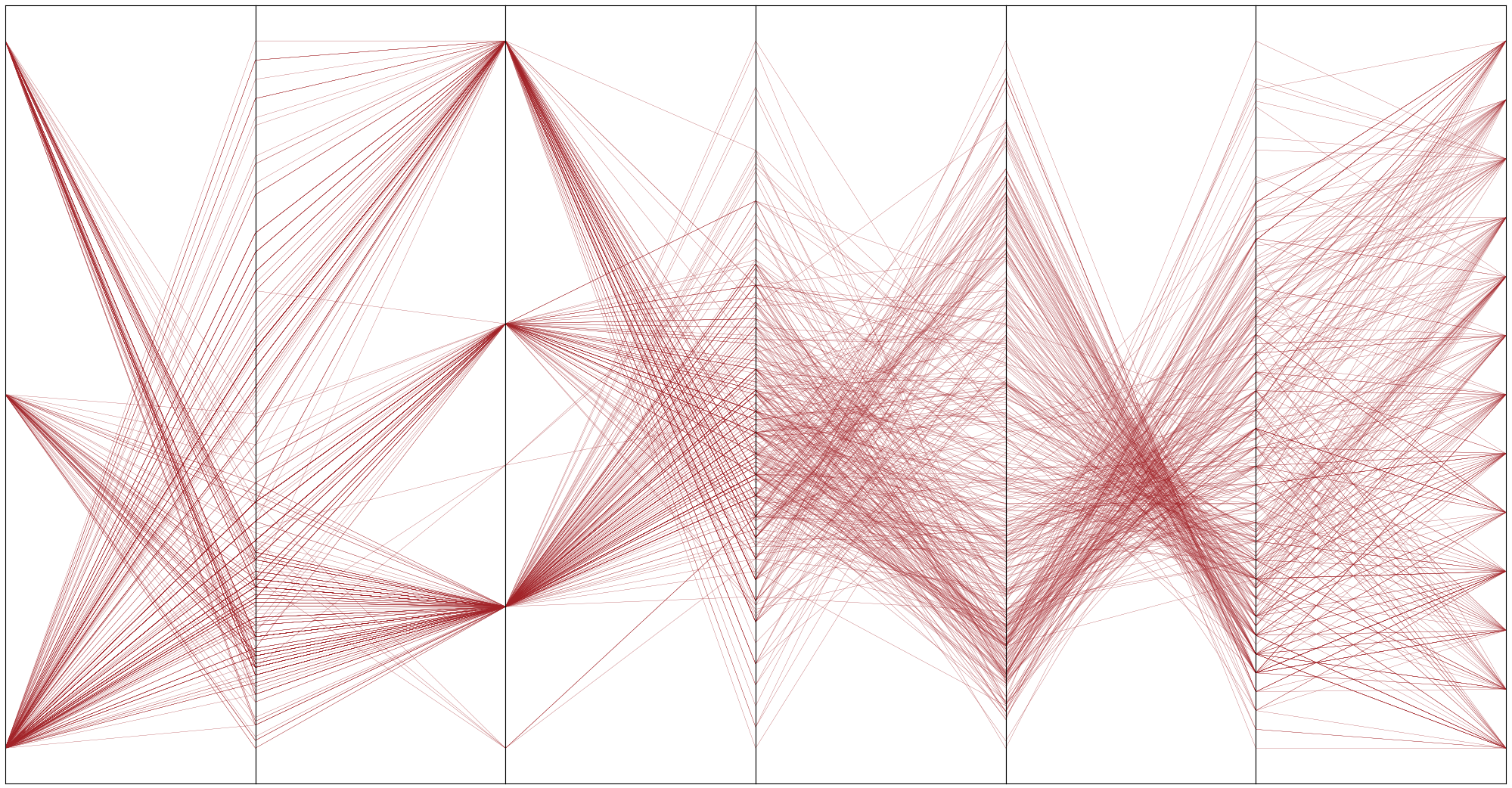}\\
\mbox{\footnotesize SA} & \mbox{\footnotesize SA}\\[3pt]
\includegraphics[width=.47\linewidth]{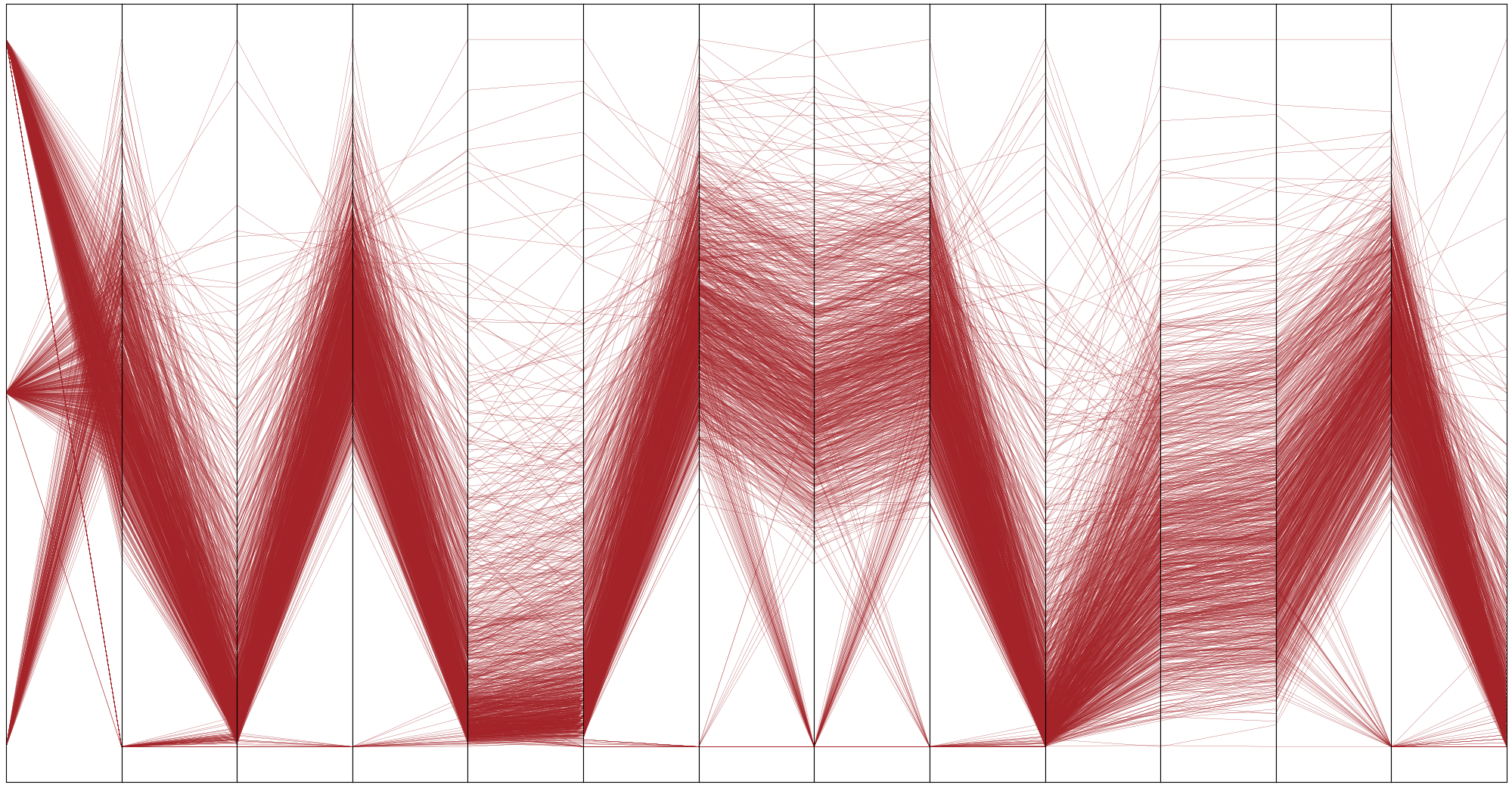}&
\includegraphics[width=.47\linewidth]{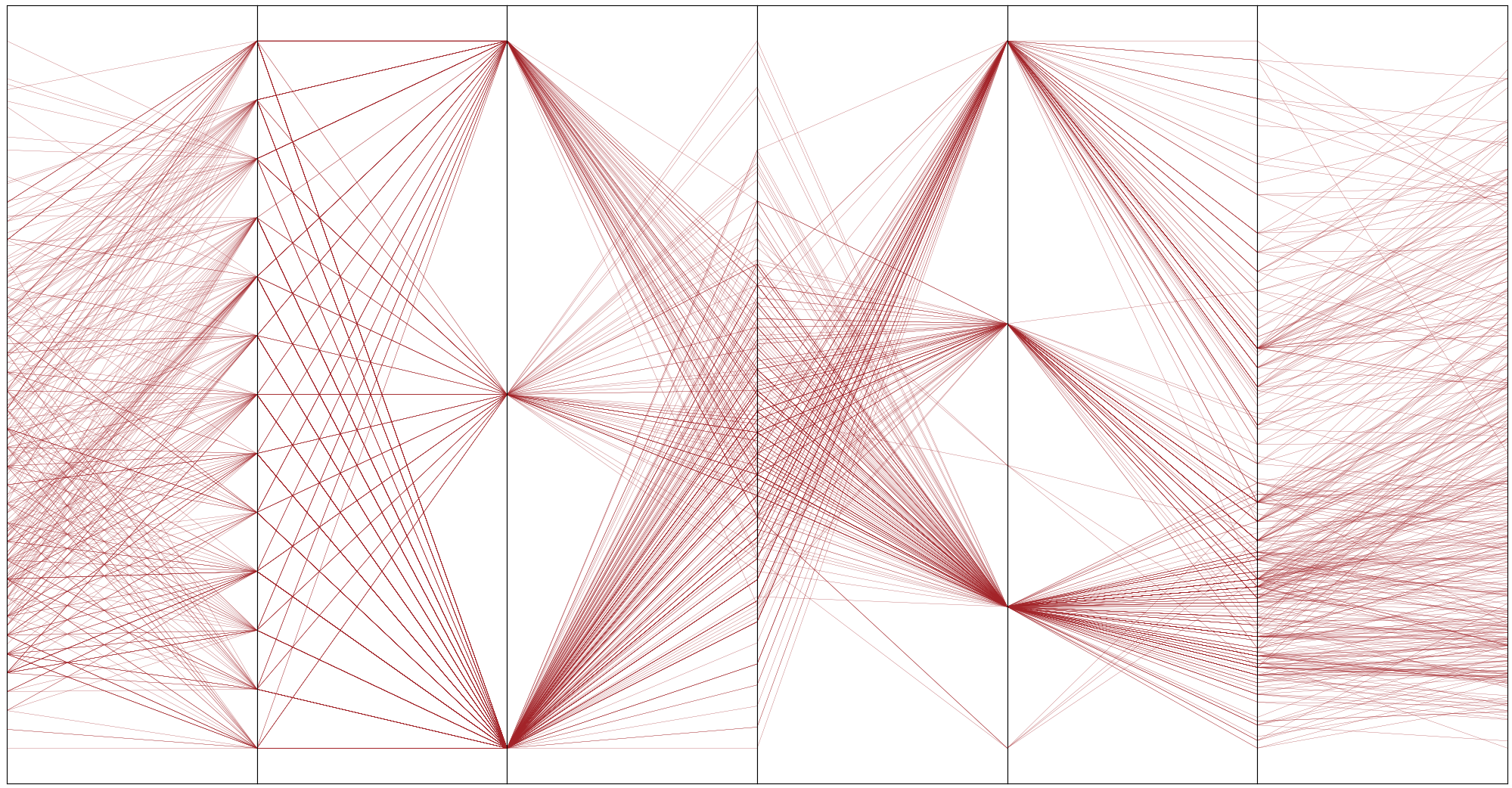}\\
\mbox{\footnotesize RS} & \mbox{\footnotesize RS}\\[5pt]
\mbox{(a) AAUP} & \mbox{(b) Cars}
\end{array}$    
\end{center}
\caption{Comparison of parallel coordinates reordering results of our \vonm, AM, and random swapping (RS), using (a) `AAUP' and (b) `Cars'  datasets and the correlation metric.}
\label{fig:parc}
\end{figure}

\textbf{Qualitative study.}
\cref{fig:parc} shows the axes reordering results for parallel coordinates, using AAUP and Cars datasets as examples. The visual quality seems to be more closely related to the nature of data rather than the approach used. We can see stronger correlations between neighboring axes in AAUP than those in Cars. Unlike SA and RS, which rely on randomness to explore the solution space, \vonm and AM seem to learn different ordering strategies. AM tends to place the axes with discrete values in the middle, leading to distracting structures. In contrast, \emph{\vonm tends to place the discrete axes at the two ends, showing a clearer pattern}. Note that SA achieves better results in AAUP, but its visualization may not demonstrate clear advantages.

\begin{figure}[!ht]
    \centering
    \includegraphics[width=.95\linewidth]{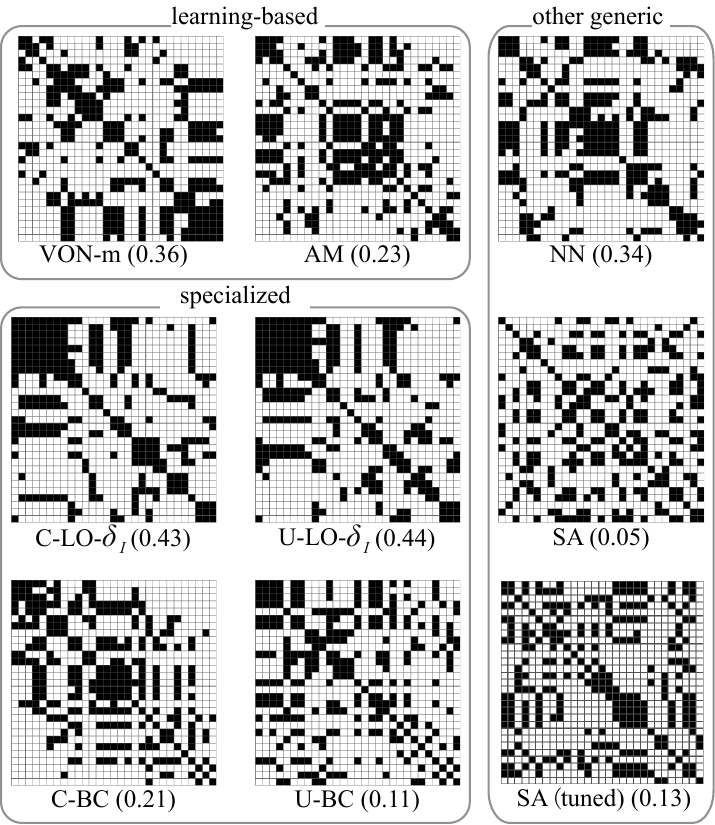}
    \caption{Matrix reordering results using FLT dataset and Moran's I. The results of different approaches are shown using a single graph $G_{78}$, which best reflects the overall performance (in terms of correlation). The number below each matrix shows the corresponding Moran's I for $G_{78}$. Note that the rows and columns are ordered simultaneously for all 96 graphs in FLT, and the performance of approaches may not be assessed in a single step. Please refer to \cref{fig:matrix-all} in the Appendix for more results sampled at an interval of ten.}
    \label{fig:Matrices-FLT}
\end{figure}

\subsubsection{Performance on matrix reordering}
\label{sec:matrix-reordering}
In this section, we examine the performance of VON in matrix reordering. We use the FLT dataset and Moran's I for comparison in the dynamic graph setting as~\cite{vanbeusekom2022simul}.
In this setting, VON orders the rows and columns in multiple matrices simultaneously, guided by the average Moran's I overall matrices. Note that in graph data, nodes may not associate with coordinates and embedding approaches may be used to generate the coordinates. We will study the impact of embedding approaches in \cref{sec:embed}. In this experiment, we average the adjacency matrices of graphs at individual steps to produce the initial coordinates.

\begin{figure}[!ht]
  \centering
  \includegraphics[width=0.95\linewidth]{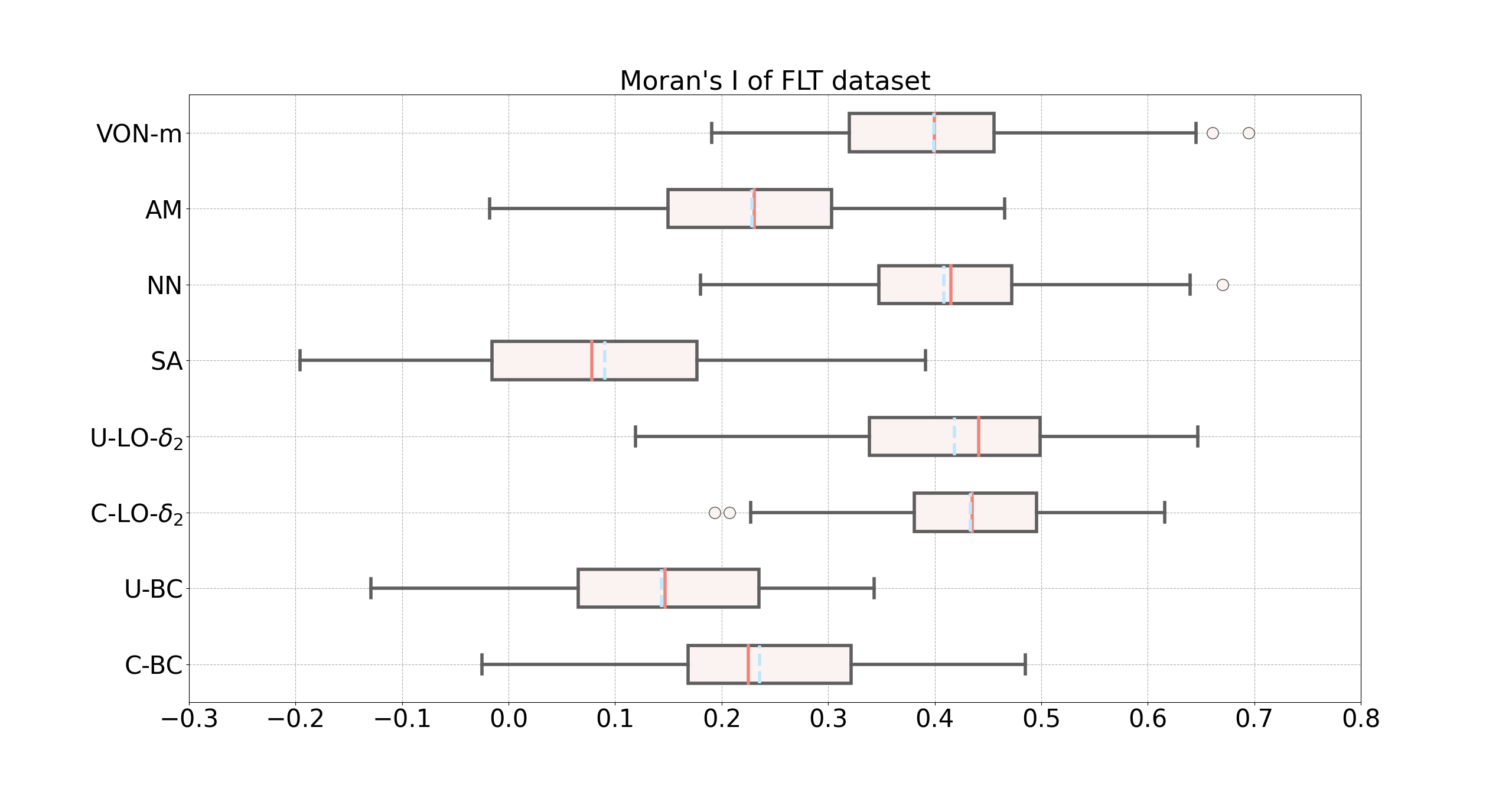}
  \caption{Matrix reordering performance using FLT dataset and Moran's I metric. The boxplot shows the distribution of Moran's I values over 96 graphs at individual time steps in FLT, the blue dashed line shows the mean value, and the red line shows the median. Larger Moran's I is better.}
  \label{fig:Moran_s_I-FLT}
\end{figure}

\textbf{Quantitative results.}
\cref{fig:Moran_s_I-FLT} shows the distribution of Moran's I at individual steps for each approach. The cyan dashed lines indicate the average Moran's I, which is used to evaluate the overall quality for dynamic graphs~\cite{vanbeusekom2022simul}. 
In terms of the average, \vonm (0.399) outperforms the other generic approaches, namely, AM (0.229), NN (0.105), and SA (0.090), by a large margin.
Compared with the specialized algorithms, \vonm clearly outperforms U-BC (0.143) and C-BC (0.236) and delivers similar performance as the SOTA approaches, i.e., C-LO-$\delta_I$ (0.433) and U-LO-$\delta_I$ (0.419). This shows the overall effectiveness of \vonm to solve the simultaneous reordering problem. 
In terms of the minimum of Moran's I (i.e., the quality of the worst matrix), \vonm (0.191), C-LO-$\delta_I$ (0.193), and U-LO-$\delta_I$ (0.119) are the only three approaches that are able to maintain positive minimums. This shows their ability to produce stable results for individual graphs.

Note that although the input of \vonm (the average graph) is known to be less ideal~\cite{vanbeusekom2022simul}, VON does not solely rely on the input to guide the ordering. Instead, it uses the quality metric to transform the input space for ordering. As Moran's I is computed at each graph, VON can still collect sufficient information to learn effective strategy.

In~\cref{fig:Matrices-FLT}, we use the graph $G_{78}$ as a representative graph, which is selected based on its correlation to the average scores. This aims to reflect the overall performance, although some methods may deliver worse performance than average at this specific graph, e.g., 0.05 at $G_{78}$ vs 0.09 on average for SA. 
We should further note that, while we fix the parameters for all methods to examine their usability across different tasks and scales, SA may benefit from parameter training on specific tasks. In \cref{fig:Matrices-FLT}, we can see that Moran's I of SA increases from 0.05 to 0.13 at $G_{78}$ with a more exhaustive searching setting. Its average performance also increases from 0.09 to 0.258. Please refer to Appendix C.IV.2 for more discussion.

\textbf{Qualitative results.}
\cref{fig:Matrices-FLT} shows the matrix reordering results with a typical graph $G_{78}$. The high-quality matrices with larger Moran's I exhibit larger black blocks, meaning that they preserve the communities in the corresponding graphs. We can see that \vonm, C-LO-$\delta_I$, U-LO-$\delta_I$, and NN produce high-quality matrix reordering results for this graph, with Moran's I larger than 0.3. However, we should also note that the order is determined simultaneously over all graphs, and the single graph does not reflect the overall quality. For example, while AM outperforms NN on average, its performance (0.23) falls behind NN (0.34).

\subsection{Discussion}

\textbf{Scope of application.} 
VON is designed to be generic and work across applications in visualization. Our experiment demonstrates that it can be applied to different scenarios, where the points are fixed (e.g., matrix reordering and axes ordering) or dynamically generated in interaction. It can optimize for various quality metrics. In terms of data, our experiment shows that VON can be applied to points with or even without coordinates. 

We expect that VON may be used for more complicated data, as pre-training on many modalities of data becomes more readily available in the era of large models. For example, well-trained encodings for text and images can be accessed easily nowadays. Additionally, a rich suite of representation techniques can be used to encode one's own data as well. 
However, we should also note that VON is still somewhat sensitive to the input embedding, as shown in \cref{sec:embed}. Therefore, improving the encoder for robustness against input should be further investigated.

VON's potential applications are diverse. It may be used by developers as a \emph{component in visual analytics interface} to order items interactively, by visualization researchers as a {reliable baseline for examining future ordering algorithms}, and by perception researchers as a {solver for customized metrics} without existing optimization algorithms. 
For example, Appendix C.II shows the gap between metric values and human perception, confirming the need for perceptual metrics. 
In this scenario, VON can be 
a reasonable choice when a specialized solver is not immediately available, allowing perception researchers to focus on metric design without worrying about how to optimize the metric. Additionally, VON's adaptability to user-specific requirements positions it as a potential solution in user-centric applications and human-in-the-loop analyses.

\textbf{Efficiency.} 
In terms of timing efficiency, VON may not outperform the most advanced specialized algorithms, but it still delivers a reasonable response time for interactive systems. More importantly, VON can be used in any system where ordering is required without additional effort to develop specialized algorithms. 

In terms of memory efficiency, the amount of network parameters and the data in one batch should be considered. By using the attention mechanism for modeling pairwise relationships between points, the size of network parameters only relies on the number of dimensions but not the number of nodes, which reduces the required space. In the training stage, the space for data relies on the number of nodes, the number of dimensions, and the batch size. In our experiment, for 2D points with a batch size of 100, ordering 20 points requires 1.5GB of graphics memory, ordering 50 points requires 4.6GB, and ordering 100 points requires 8.1GB, which is often available on modern graphics cards. After training, in the inference stage, the memory costs for ordering 20, 50, and 100 points become 1.2GB, 1.4GB, and 1.9GB, respectively. When the memory size is a concern, one may reduce the batch size to support larger point sets. One may also train VON on smaller sets and apply it to order larger ones. Additionally, we should note that the memory required in the inference stage is much smaller as only one point set is processed at a time. This means we may train VON on more powerful machines and use it on less powerful ones.

\section{Conclusions and Future Work}
We propose VON that can facilitates various kinds of ordering tasks across scales and quality metrics. We propose a conceptual framework for embedding the input points into a latent space that is optimized for ordering task. The framework uses attention to collect global information regarding all data points being ordered and local information regarding the current ordering status as well. We introduce a novel repositioning module that places the data points into the ordering context, so that it may learn common structures regardless of the location and scale of point sets. We demonstrate with experiments that our approach can outperform most baselines, especially in generalizability and transferability. We also examine the performance against specialized SOTA approaches and show that VON can achieve comparable ordering results. We evaluate VON on twelve data sets with eleven metrics, covering various kinds of applications. 

\textbf{Future work.} 
Our method still suffers from several limitations. In the future, firstly, we would like to improve the efficiency of VON in both training and inference stages. We may further leverage knowledge distillation to reduce the size of network for interactive performance in inference, and collect knowledge from similar datasets of problem structures to speedup the training procedure. This may help VON to better adapt to new tasks. Secondly, to capture global and local structures more effectively, we may further design structures that are specifically designed for repositioning purpose. Thirdly, we want to explore the encoding embedding process to enhance the information feed into the model. Finally, we can also design or discuss special methods for costs that cannot be minimized, making them applicable in VON. Finally, we would like to explore the possibility of learning ordering strategies directly from users. This may further reduce the cost to design quality metrics for different applications. 

\vspace{-0.1in}
\section*{Acknowledgements}
This research was supported in part by the National Natural Science Foundation of China through grants 62172456 and 62372484.
The authors would like to thank the anonymous reviewers for their insightful comments.

\vspace{-0.1in}
\bibliographystyle{abbrv}
\bibliography{template}

\begin{IEEEbiography}[{\includegraphics[width=1in,height=1.25in,clip,keepaspectratio]{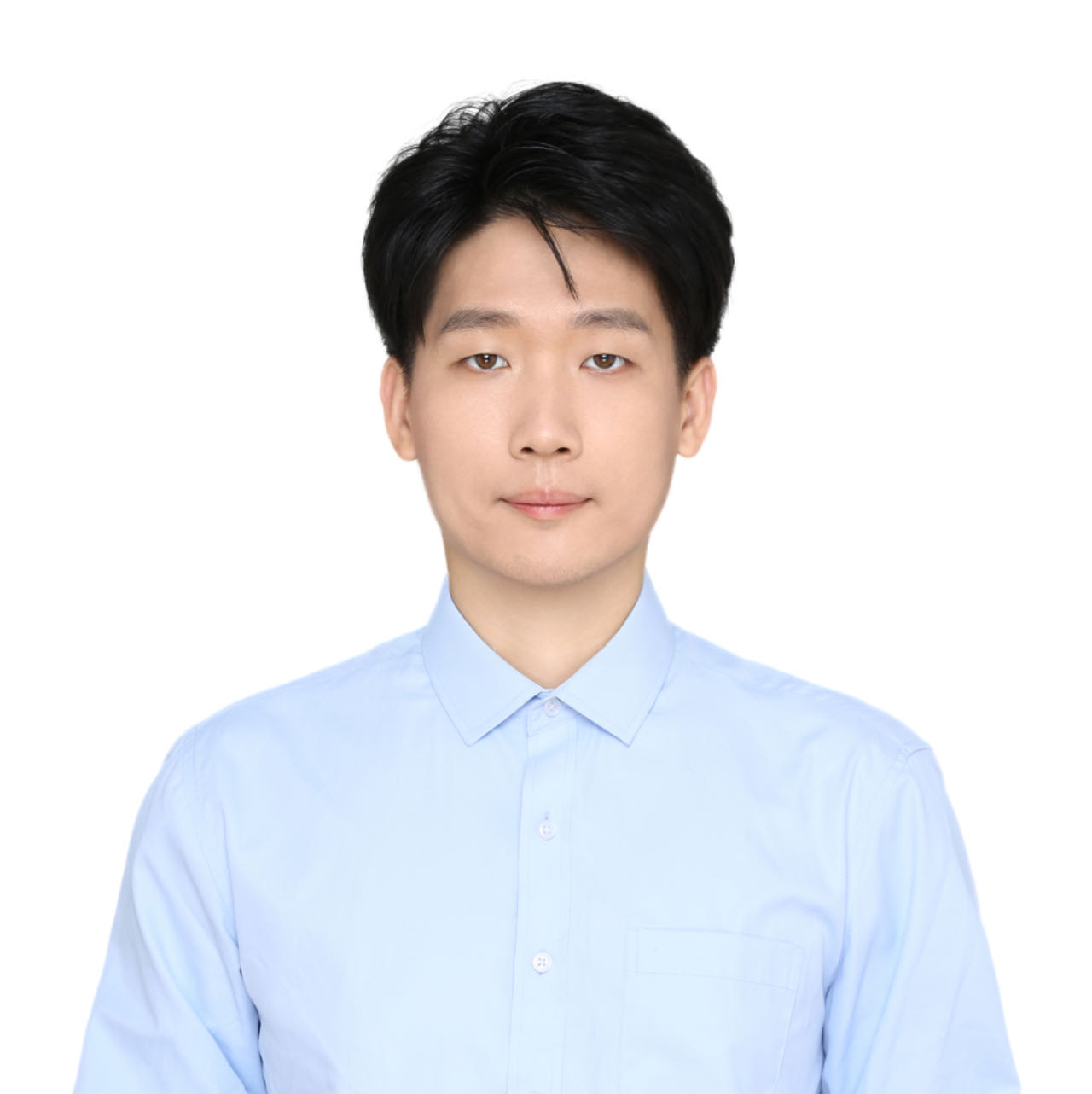}}]{Zehua Yu} is a PH.D.\ student of computer science at Sun Yat-sen University. He received a MS degree in electric and communication engineering from Shantou University in 2022. His major research interest is scientific visualization, visual analytics, deep learning and time-series analysis.
\end{IEEEbiography}

\vspace{-.5in}

\begin{IEEEbiography}[{\includegraphics[width=1in,height=1.25in,clip,keepaspectratio]{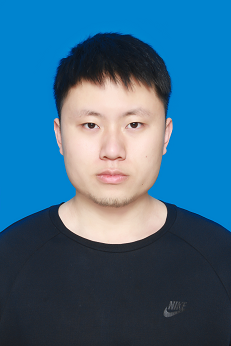}}]{Weihan Zhang} is a PH.D.\ student of computer science at Sun Yat-sen University. He received a Bachelor's degree in software engineering from Chongqing University in 2022. His research interests include scientific visualization and visual analytics. 
\end{IEEEbiography}

\vspace{-.5in}

\begin{IEEEbiography}[{\includegraphics[width=1in,height=1.25in,clip,keepaspectratio]{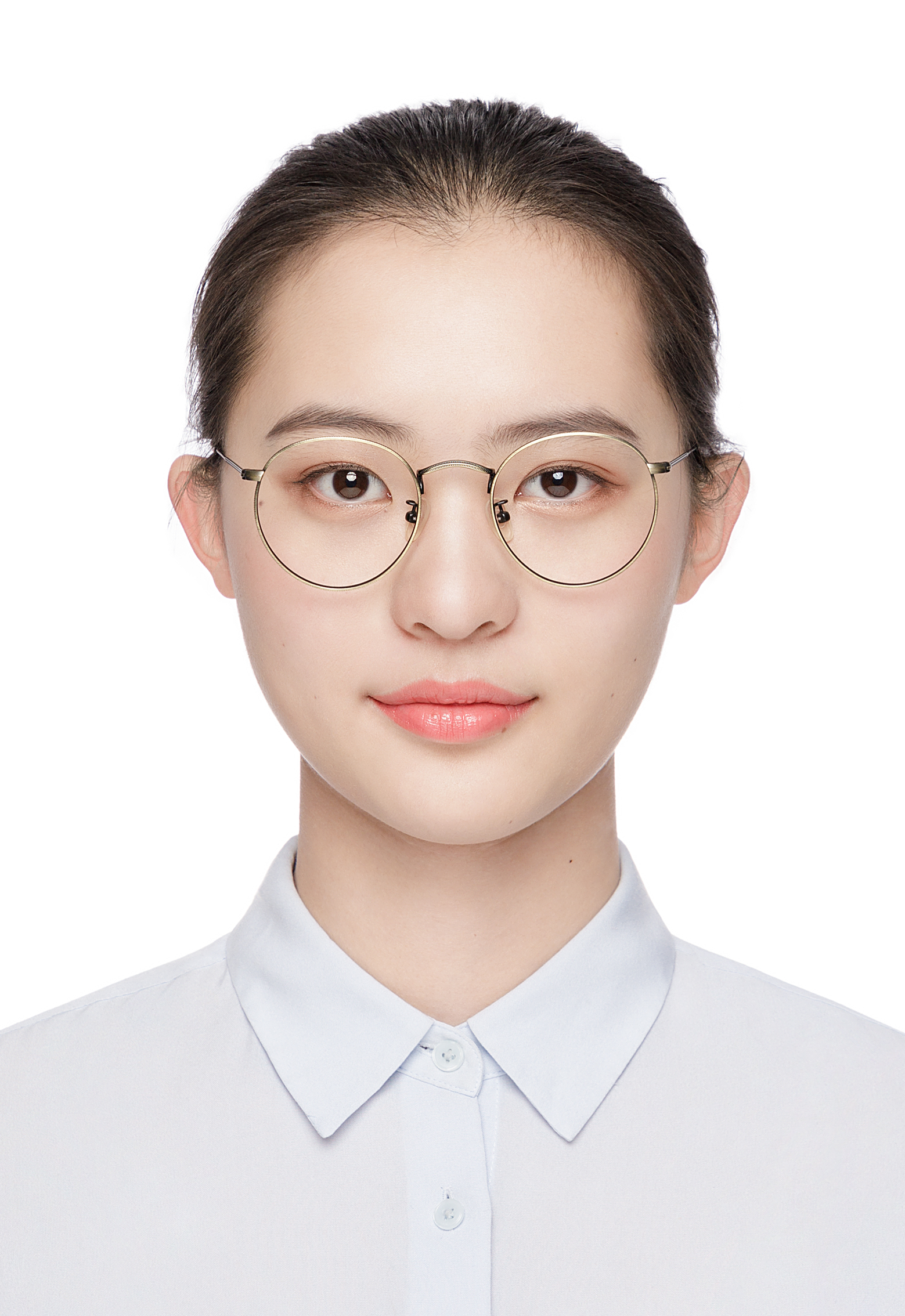}}]{Sihan Pan} is a master student at Sun Yat-sen University and National Supercomputer Center in Guangzhou. She obtained a Bachelor's degree in computer science and technology from Sun Yat-sen University in 2023. Her major research interest is scientific visualization and visual analytics.
\end{IEEEbiography}

\vspace{-.5in}

\begin{IEEEbiography}[{\includegraphics[width=1in,height=1.25in,clip,keepaspectratio]{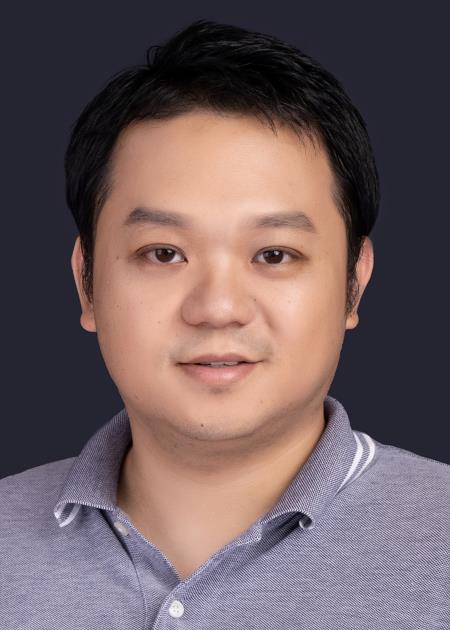}}]{Jun Tao} 
is an associate professor of computer science at Sun Yat-sen University and National Supercomputer Center in Guangzhou. He received a Ph.D.\ degree in computer science from Michigan Technological University in 2015. His major research interest is scientific visualization, especially in applying information theory, deep learning, and optimization techniques to interactive flow visualization and multivariate data exploration. 
\end{IEEEbiography}

\clearpage

\setcounter{page}{1}
\setcounter{figure}{0}
\setcounter{section}{0}
\setcounter{table}{0}

\section*{Appendix}

\section*{A. Background of Attention and Optimization}

\textbf{Attention.} As attention is heavily used in VON, we also provide a simplified notation in this paper. An attention module is parameterized by three learnable matrices: $\weight_q$, $\weight_k$, and $\weight_v$. These three matrices are used to linearly transform a vector $\hidden_i$ into three vectors: namely, the query $\query_i=\weight_q \cdot \hidden_i$, the key $\key_i=\weight_k \cdot \hidden_i$, and the value $\valvec_i=\weight_v \cdot \hidden_i$. The attention between two data points $\hidden_i$ and $\hidden_j$ is defined as:

\begin{align}
\att(\hidden_i, \hidden_j) &= \query_i \cdot \key_j \notag \\
&= (\weight_q \hidden_i)^T\cdot(\weight_k \hidden_j) \notag \\
&=\hidden_i \cdot (\weight_q^T \cdot \weight_k) \cdot \hidden_j.
\label{eq:attention}
\end{align}

\noindent This indicates that the attention between two nodes can be described by a single matrix $(\weight_q^T \cdot \weight_k)$ for transforming the product of $\hidden_i$ and $\hidden_j$. In this paper, we use $\att(\hidden_i, \hidden_j)$ to denote the attention module without introducing the matrices. Similarly, we denote the value vectors as functions of respective inputs, i.e., $\val(\hidden_i)=\weight_i \cdot \hidden_i$. To emphasize that the attention modules are trainable and they may vary across layers, we use subscripts to denote different attention modules, e.g., $\att_k$ and $\val_k$ for the $k$-th attention head.

\textbf{Multi-head attention (MHA).} MHA uses multiple attention heads to capture information from different facets. Each attention head updates a vector $\hidden_i$ by collecting information from all other data points (i.e., $\hidden_j$), and the attentions from multiple heads are aggregated through summation:

\begin{align}
    \hidden_i^{l+1} = \sum_{k} \sum_{j} \att_k (\hidden_i^l, \hidden_j^l) \cdot \val_k(\hidden_j^l),
\end{align}

\noindent where $l$ is the index of the attention layer and $k$ is the index of the attention head. In this way, the resulting vector $\hidden_i^\prime$ is able to sense the neighbor of the data point. By computing the pairwise attention for all points, this mechanism incorporates the interactions between points.

\textbf{Quality metric and loss function.} Formally, the quality metric for ordering can be any function that takes an order $\Pi$ of a point set $X$ as input, evaluates the quality of $\Pi$, and outputs a scalar value indicating the quality. VON uses the quality metric as the loss function $L(\Pi)$, so that the network learns to optimize the metric.

\textbf{Greedy rollout strategy and optimization.}
\label{sec:optimize}
Inspired by \cite{williams1992simple}, we use the greedy rollout strategy in reinforcement learning to improve the ordering neural network over itself, leveraging information in the previous ordering samples. The term ``greedy rollout'' refers to the selection of the most promising actions. In our scenario, the greedy rollout strategy means that the network outputs the most likely data point at each step to form a sequence. The use of the greedy strategy allows VON to focus on learning the difference between the optimal solution and the greedy solution, which is more effective than learning from scratch.

Formally, the optimization minimizes the expectation of loss function $\mathbf{L_E}(\theta|X)=\mathbf{E}_{p_{\theta}(\Pi|X)}[L(\Pi)]$, where $X$ is the point set to be ordered, $\Pi$ is the order, and $L(\Pi)$ is the loss evaluated on the order $\Pi$. We used the best-observed strategy $b$ as the baseline in the gradient estimator:

\begin{equation}
    \nabla \mathbf{L_E}(\theta|X)=\mathbf{E}_{p_{\theta}(\Pi|X)}[(L(\Pi)-L(b(X)))\nabla log p_{\theta}(\Pi|X)],
\end{equation}

\noindent where $L(b(X))$ is the loss evaluated on the order produced by the baseline $b$. Note that baseline $b$ is the model with the best performance, and it will be regularly updated during the training process. %
Similar to DQN \cite{mnih2015human}, we stabilize the baseline by freezing its parameters $p_{\theta}$ for a fixed number of every epoch. By using the best-observed model as the baseline, we consistently challenge the model against the top-performing model in each epoch.
We use Adam as the optimizer for parameter updates.

\section*{B. Detail of Experimental Setting}

\textbf{Data sets.} We use six data sets including four data sets with data points in respective latent embedding spaces (i.e., FashionMNIST, CIFAR-10, ImageNet, CORA, and DBLP), two dynamic graph datasets (i.e., FLT and SCH), and four high dimensional datasets for axes ordering (i.e., Cars, AAUP, Coal Disaster and Census Income):

\begin{myitemize}
\item \emph{FashionMNIST} is a black and white clothing dataset containing images in ten categories. Each image is 28 $\times$ 28 pixels in size.
\item \emph{CIFAR-10} is a universal object dataset with ten categories, where the image size is 32 $\times$ 32 pixels.
\item \emph{ImageNet} is a large high-definition image dataset. We use dog images with 120 breeds. While the original images are different in size, we reduce their sizes to 64 $\times$ 64 pixels.
\item \emph{CORA} is a scholarly dataset with machine learning papers. It covers 2708 papers in seven categories. 
\item \emph{DBLP} is a dataset from the DBLP public bibliography. We use a coauthor network extracted by Meng et al.~\cite{meng2019co} from the publications in 172 computer science conferences.
\item \emph{FLT} is the ``flashtap'' data that was used for MultiPiles \cite{vanbeusekom2022simul}. It represents a functional brain connectivity network in a Parkinson’s disease study.
\item \emph{SCH} is a dynamic graph dataset, where the graphs encode social interaction between children and teachers at a primary school.

\item \emph{MNIST} is a large collection of handwritten digits.
\item \emph{Cars} is a dataset of car models, comprising seven attributes (e.g., years, makes, and types) for each of the total 398 car models.
\item \emph{AAUP} is the faculty salary data collected from the March-April 1994 issue of Academe~\cite{peng2004clutter}.
\item \emph{Coal Disaster} records the coal-mining disasters between March 1851 and March 1962~\cite{peng2004clutter}.
\item \emph{Census Income} is part of the U.S. Census Bureau database with 42 demographic and employment-related attributes such as age, education, occupation, and marital status~\cite{peng2004clutter}.

\end{myitemize}

\textbf{Baselines.}
We compare three variations of VON against four generic approaches (i.e., AM~\cite{kool2019attention}, SA~\cite{Granville1994Simulated}, NN~\cite{NN}), one layout algorithm (i.e., SM~\cite{Gansner2005Graph}), four specialized algorithms (i.e., C-LO-$\delta_I$, U-LO-$\delta_I$, C-BC and U-BC) for matrix reordering~\cite{behrisch2016matrix}, Random Swapping for axes reordering, and four specialized methods (i.e., LAS/FLAS~\cite{barthel2023improved}, Isomatch~\cite{fried2015isomatch} and Kernelized Sorting~\cite{quadrianto2008kernelized}) for images reordering as baselines:

\begin{myitemize}
\item \emph{Attention model (AM)}~\cite{kool2019attention} is an attention-based model designed for solving routing problems.

\item \emph{Simulated annealing (SA)}~\cite{Granville1994Simulated} is a randomized search algorithm that may accept worse solutions during the search process in order to escape local optima and find the global optimum.

\item \emph{Nearest neighbor (NN)}~\cite{NN} is a distance-based classification algorithm.

\item \emph{Stress majorization (SM)}~\cite{Gansner2005Graph} is a graph layout algorithm that preserves the target distance between data points in lower dimension. For ordering purpose, we use this algorithm to embed the data points in 1D.

\item \emph{C-LO-$\delta_I$, U-LO-$\delta_I$, C-BC, and U-BC}~\cite{vanbeusekom2022simul} are four specifically designed algorithms for matrix reordering based on the Moran's I.

\item \emph{Random Swapping} is a randomized algorithm that optimizes the order from an initial configuration by randomly swapping two data points. It can be used to solve axes reordering~\cite{peng2004clutter}.

\item \emph{IsoMatch}~\cite{fried2015isomatch} is a method to arrange collections of objects by minimizing an energy value, which is designed to preserve a given distance measure.

\item \emph{Kernelized Sorting}~\cite{quadrianto2008kernelized} is an approach that performs matching by requiring a similarity measure only within each class. It achieves this by maximizing the dependency between matched pairs of observations using the Hilbert-Schmidt Independence Criterion.

\item \emph{LAS/FLAS}~\cite{barthel2023improved} designed based on Linear Assignment Sorting (LAS). LAS combines ideas of self-organizing map and the swap sorting method to optimally swap all vectors simultaneously. FLAS handles larger quantities of images by replacing the global assignment with multiple local swaps.

\end{myitemize}

For the \textbf{the three generic solvers}, the following paragraphs provide further details:

\emph{AM} uses attention mechanisms and reinforcement learning for training. AM formulates a problem instance \( s \) as a graph with \( n \) nodes, where each node \( i \in \{1, \dots, n\} \) is represented by features \( x_i \). For TSP, \( x_i \) is the coordinate of the node \( i \), and the graph is fully connected (including self-connections). Generally, the model can be considered a Graph Attention Network~\cite{kool2019attention}, incorporating graph structure through a masking procedure. A solution (order) \( \pi = (\pi_1, \dots, \pi_n) \) is defined as a permutation of the nodes, where \( \pi_t \in \{1, \dots, n\} \) and \( \pi_t \neq \pi_{t'} \) for \( t \neq t' \). The attention-based encoder-decoder model defines a stochastic policy \( p(\pi|s) \) for selecting a solution \( \pi \) given a problem instance \( s \), factorized and parameterized by \( \theta \). The encoder generates embeddings for all input nodes, while the decoder produces the sequence
$\pi$ one node at a time, using the embeddings, a mask, and context. For ordering, after a partial order is fixed, the task is to find a sequence from the last node, through unvisited nodes, back to the first. The decoder context includes embeddings of the first and last node, and a mask indicates visited nodes.

\emph{SA} starts with an initial solution and generates a neighboring solution at each step, calculating its energy difference. If a new solution is better (i.e., has lower energy), the solution will be accepted; otherwise, the solution will only be accepted with a certain probability determined by the current ``temperature'' and energy difference, following~\cite{dennis2006Simulated}: $P = \exp\left(-\frac{\Delta E}{T}\right)$, where $P$ represents the acceptance probability, $\Delta E$ is the cost difference between the new solution and the current solution, $T$ is the current temperature. The probability of accepting worse solutions allows the algorithm to escape local optima, while the decreasing temperature is designed for convergence. In the ordering problem, SA explores the solution space by randomly swapping elements and optimizes the ordering objective function. Behrisch et al.~\cite{behrisch2016matrix} reported that Brusco et al. \cite{brusco2008heuristic} could find the optimal solution for a $35 \times 35$ connection in 20 to 40 minutes with the right settings. 

\emph{NN} categorizes new instances based on the majority class of their nearest neighbors. The nearest neighbor heuristic is a path-based approach that starts with a randomly selected single node and gradually adds the closest node to form a path, ultimately connecting the start and end nodes to form a tour. The algorithm always begins with the first node in the input to ensure deterministic results. In ordering tasks, NN builds a relatively ordered arrangement by minimizing the distance between adjacent elements.

For \textbf{1D layout}, \emph{SM} is an optimization algorithm used for multidimensional scaling (MDS), aiming to optimize the layout of points by minimizing a stress function. In ordering, it projects points onto a one-dimensional space so that the one-dimensional distances after projection are as consistent as possible with the original high-dimensional distances.

For \textbf{matrix reordering}, we use a union approach~\cite{behrisch2016matrix} combines all graphs in \( G \) into a single graph \( H \) on the same vertex set \( V \), where each edge weight reflects how often it appears across the graphs in \( G \). This is equivalent to summing all 0-1 adjacency matrices. Any ordering algorithm can then be applied to \( H \), with the resulting order used for all graphs in \( G \). For a vertex \( v \), \( N(H, v) = \sum_{G \in G} N(G, v) \) is a vector of length \( n \), where each entry represents the edge's weight, \( 0 \) to \( k \). We further elaborate on the baselines for matrix reordering:

\emph{U-LO-$\delta_I$} works on weighted graphs, requiring a distance measure \( \delta \), with MultiPiles using \( \delta(u, v) = L2(N(H, u), N(H, v)) \), the Euclidean distance between neighborhoods. While squaring does not affect distance comparisons, it may change the final ordering due to how distances are summed.

\emph{U-BC}, usually used in unweighted settings, extends to weighted graphs by multiplying the weights of intersecting edges and summing them across all pairs.

\emph{C-LO-$\delta_I$} focuses on pairwise distances, ensuring \( \delta(u, v) \) is low when \( u \) and \( v \) have similar neighborhoods across many graphs. Unlike the union approach, which aggregates neighborhoods first, the collection-aware method calculates \( \delta(u, v) = \sum_{G \in G} L2(N(G, u), N(G, v)) \). Alternative distance measures can replace \( L2 \), and squaring distances alters the clustering step.

\emph{C-BC} revert to counting crossings, but separately for each graph in \( G \), and then sum the results. In a collection-aware method, the target rank of a vertex is determined for each graph individually, and then aggregated. It computes the median rank of each vertex's neighbors in each graph, then computes the overall median of these ranks. If a vertex has no neighbors in a graph, it's excluded. Ordering is done based on these medians of medians.

For \textbf{axes reordering}, \emph{Random swapping} is a simple optimization technique used to explore solution spaces by randomly selecting two elements from a given solution and swapping their positions. This method begins with an initial solution, which may be generated randomly or based on a heuristic. The swap operation is applied iteratively, where two elements are randomly chosen and exchanged, leading to a new solution. After each swap, the solution is evaluated to determine if it improves the objective function, such as minimizing distance or maximizing a score. The new solution may be accepted based on improvement or according to a probabilistic rule, depending on the specific algorithm. While random swapping is easy to implement and encourages exploration of the solution space, it may not always guarantee efficiency, especially in large spaces, as purely random swaps do not always lead to significant improvements.

The \textbf{four specialized algorithms for image ordering} are explained in detail as follows:
\emph{IsoMatch} aligns images into a specified spatial arrangement based on a given distance matrix. The inputs include a set of images \( I \) and a symmetric pairwise distance matrix \( D \), where \( d_{ij} \) represents the distance between images \( I_i \) and \( I_j \). IsoMatch first uses nonlinear dimensionality reduction to project the images into a 2D space, producing an initial layout. This layout is then transformed to roughly fit the desired output pattern. A bipartite graph is constructed, and a bipartite matching algorithm minimizes the movement of images to their final positions, resulting in a distance-preserving output arrangement.

\emph{KS} aims to maximize the dependence between two sets of variables \( X \) and \( Y \) by permuting \( Y \) to align with \( X \), using the Hilbert-Schmidt Independence Criterion (HSIC) as the dependence measure. HSIC has several advantages: it is robust to small changes, computationally efficient since it only requires kernel matrices, and allows flexibility in kernel selection, enabling the incorporation of prior knowledge. The sorting problem is formulated as an optimization, which maximizes the trace of the product of kernel matrices for \( X \) and the permuted \( Y \):
$
\pi^* = \arg \max_{\pi \in \Pi_m} \text{tr} \left( K \pi^T L \pi \right)
$
This optimization is a special case of the quadratic assignment problem and is NP-hard. However, it is proven that, for simple cases such as scalar random variables with linear kernels, the optimal permutation of \( Y \) is equivalent to sorting \( Y \) in the same order as \( X \). Specifically, for \( X = Y = \mathbb{R} \), with kernels \( k(x, x') = xx' \) and \( l(y, y') = yy' \), sorting \( Y \) in ascending order or its reverse gives the optimal solution. Kernelized sorting generalizes the idea of sorting, using kernel functions to capture relationships between \( X \) and \( Y \) indirectly through their feature spaces rather than their direct values.

\emph{LAS/FLAS} are two linear assignment sorting algorithms. LAS combines the ideas of Self-Organizing Maps (SOM) and Swap Sorting (SSM) to optimally swap all vectors simultaneously for image sorting. Initially, input vectors are randomly placed on a map, which is then spatially low-pass filtered to create a smoothed version representing neighborhoods. The input vectors are subsequently assigned to their best matching positions on the map. While LAS provides high sorting quality, it becomes computationally expensive for large image sets. To address this, Fast Linear Assignment Sorting (FLAS) modifies LAS by replacing global assignments with multiple local swaps, significantly improving speed while maintaining sorting quality for large datasets.

\textbf{Parameter settings.} In all experiments, we fix the parameters of VON and the baselines, aiming to examine the method's usability without extensive tuning. For the specific parameter settings of VON and various baselines, please refer to our code. For the baselines, we report their settings as follows:

For the simulated annealing, the temperature setting significantly affects the final results. We follow the guidance of Dennis et al. \cite{dennis2006Simulated}, which suggests a cooling rate between 0.8 and 0.99, with no specific requirements for restarts, temperature length, or other parameters. Based on this, we set the cooling rate to 0.9, the temperature length to 100n, the number of restarts to 5, and the stopping temperature to 0.1. For the matrix reordering, where SA delivers the worst performance, we include an additional experiment to examine the impact of parameter tuning, as discussed in Appendix C.IV.2. 

For the image ordering methods, both IsoMatch and KS require a target grid to which the images are mapped. Because image ordering can be considered as mapping the images to a 1D grid, we use a target grid of $1\times m$ for these two methods.
For the comparison with LAS/FLAS, the methods require the number of images to be a perfect square. Thus, for 50 or 150 input images, only the first 49 and 144 images are ordered, respectively. \vonm uses the same subset for a fair comparison. The evaluation focuses on \vonm's objective functions rather than direct performance comparisons with the baselines, so the slight differences in image numbers do not affect \vonm's assessment. Default parameter settings from each method's code are used for all experiments, and \vonm's parameters remain unchanged for consistency. For LAS/FLAS, the default parameters are LAS (radius factor=0.85) and FLAS (nc=49, radius factor=0.7). In the experimental setup, we maintain the default parameter settings for these methods as provided in their code.

\textbf{Quality Metrics.}
We study two quality metrics (i.e., TSP and stress) for 1D layout, four metrics (i.e., Moran's I, LA, PR, and BW) for matrix reordering, two metrics (i.e., Symmetry and Correlation) for axes reordering, and three metrics (i.e., DPQ, Energy Value, and KS objective function) for images reordering:

\begin{myitemize}
\item \emph{TSP} distance is the total path length in the traveling salesman problem (TSP). In terms of visualization, minimizing TSP distance will place similar items closer and enhance perception. For example, TSP is used in ordering rows and columns in a matrix~\cite{henry2006matrixexplorer, behrisch2016matrix}.

\item \emph{Stress} is the loss in stress majorization, a widely used algorithm for graph layout \cite{Gansner2005Graph}. Stress majorization preserves the target distance in low dimension by minimizing the stress $\sum w_{ij}(|x_i - x_j|-d_{ij})^2$. While stress majorization is commonly used for 2D layout, it is adopted to arrange river branches in 1D~\cite{zhang2021surfriver}. Stress majorization brings similar items closer as well.

\item \emph{Moran's I} is a statistical measure used to determine spatial autocorrelation in a dataset \cite{moran1950Notes}. It measures the similarity between the values of a variable and neighboring variables. This metric was used in matrix reordering~\cite{vanbeusekom2022simul}.

\item \emph{Linear arrangement (LA)} is the sum of the distances between the vertices of the edges of a graph (i.e., $LA(\phi, G)=\sum_{(u,v) \in E} \lambda((u,v), \phi, G)$) \cite{behrisch2016matrix}.

\item \emph{Profile (PR)} is the sum, for each column $i$ of the matrix, which can be intuitively understood as the “raggedness”: the distance from the diagonal (with coordinates $(i,i)$) to the farthest-away non-zero cell for that column (with coordinates $(i, j)$) \cite{behrisch2016matrix}.

\item \emph{Bandwidth (BW)} is the maximum distance between two vertices given an order $\phi$ \cite{behrisch2016matrix}.

\item \emph{Symmetry} is the balanced and proportionate similarity found in two halves of an object, mirroring each other across a central axis or point. It can be used in axes reordering for star plots~\cite{miller2019evaluating}.

\item \emph{Correlation} is a statistical measure that evaluates to what extent two or more variables fluctuate together. It can be used in axes reordering~\cite{miller2019evaluating}.

\item \emph{Energy value} is a general objective function to evaluate how the solutions preserve pairwise distance in IsoMatch~\cite{fried2015isomatch}.

\item \emph{KS objective function} measures the similarity between two sets of images in high-dimensional spaces under a certain arrangement using the matrix trace, which guides the optimization process to find the optimal arrangement. It calculates the kernel matrix of the arranged images and grid data, performs a de-centering operation, and finally uses a locally optimal smoothing formula as the cost~\cite{quadrianto2008kernelized}.

\item \emph{Distance Preservation Quality (DPQ)} is the ratio of the p-norms of the distance preservation gains of the actual arrangement to a perfect arrangement. Larger DPQ indicates better quality~\cite{barthel2023improved}.

\end{myitemize}

\textbf{Embedding methods.} We study five embedding approaches: 
\begin{myitemize}

\item \emph{Principal component analysis (PCA)} is a linear dimensionality reduction technique that transforms high-dimensional data into a lower-dimensional representation while preserving the differences along axes exhibiting the largest variations.

\item \emph{Truncated singular value decomposition (T-SVD)} is a dimensionality reduction method that works well with sparse data by selecting a subset of singular values and their components. 

\item \emph{Autoencoder (AE)} is a neural network-based technique for unsupervised dimensionality reduction and feature learning. 

\item \emph{Isomap} is a nonlinear dimensionality reduction technique that focuses on preserving the pairwise geodesic distances (shortest path distances) between data points in a lower-dimensional space. 

\item \emph{Locally linear embedding (LLE)} is another nonlinear dimensionality reduction method that seeks to preserve local relationships between data points.
\end{myitemize}

\textbf{Sampling sizes.} We decide the sampling sizes by referencing similar scenarios. For example, in for visualizing image collections, Nguyen et al.~\cite{nguyen2008interactive} and Han et al.~\cite{han2016tree} evaluated the visualization with 25, 50, 75, 100, and 200 images, with 100 identified as an effective choice for balancing usability and clarity. Similarly, image search engines like Google Images typically display around 50 images per page on a 27-inch 4K screen at default zoom settings. In visualizing graph clusters, graphs with 50 and 100 nodes were used to demonstrate correlations between nodes in matrix visualizations~\cite{al2024improved}. For axis reordering tasks, examples commonly feature numbers of axes ranging from 5 to 14~\cite{johansson2015evaluation, miller2019evaluating}. While a famous early study~\cite{miller1956magical} suggested that the cognitive load of objects that humans could perceive simultaneously was about seven, with two floating up and down. A larger number of data items may be used as an overview before detailed investigation. Based on these references, three sampling sizes: 50, 100, and 150 are considered for each sampling strategy.

\section*{C. Additional Results}
\label{sec:add-results}
\subsection*{C.I Ablation Study}

\subsubsection*{C.I.1 Impact of VON variants and sampling strategies}

\textbf{Further investigation of training data sizes and scales.}
We further examine the performance using training samples of varying point set sizes, as shown in \cref{tab:FM-ablation}.II.
We still find that the model performs best when the size and scale of training point sets match those of testing sets. This further confirms our hypothesis that data of different scales exhibit different structures. For optimal performance, we suggest training the model on samples of equivalent scales. However, in general, we find that $\vonm$ may better transfer knowledge across sample sizes and scales, as shown in \cref{tab:FM-ablation}.I.

\begin{figure}[!thbp]
  \centering
  \includegraphics[width=1\linewidth]{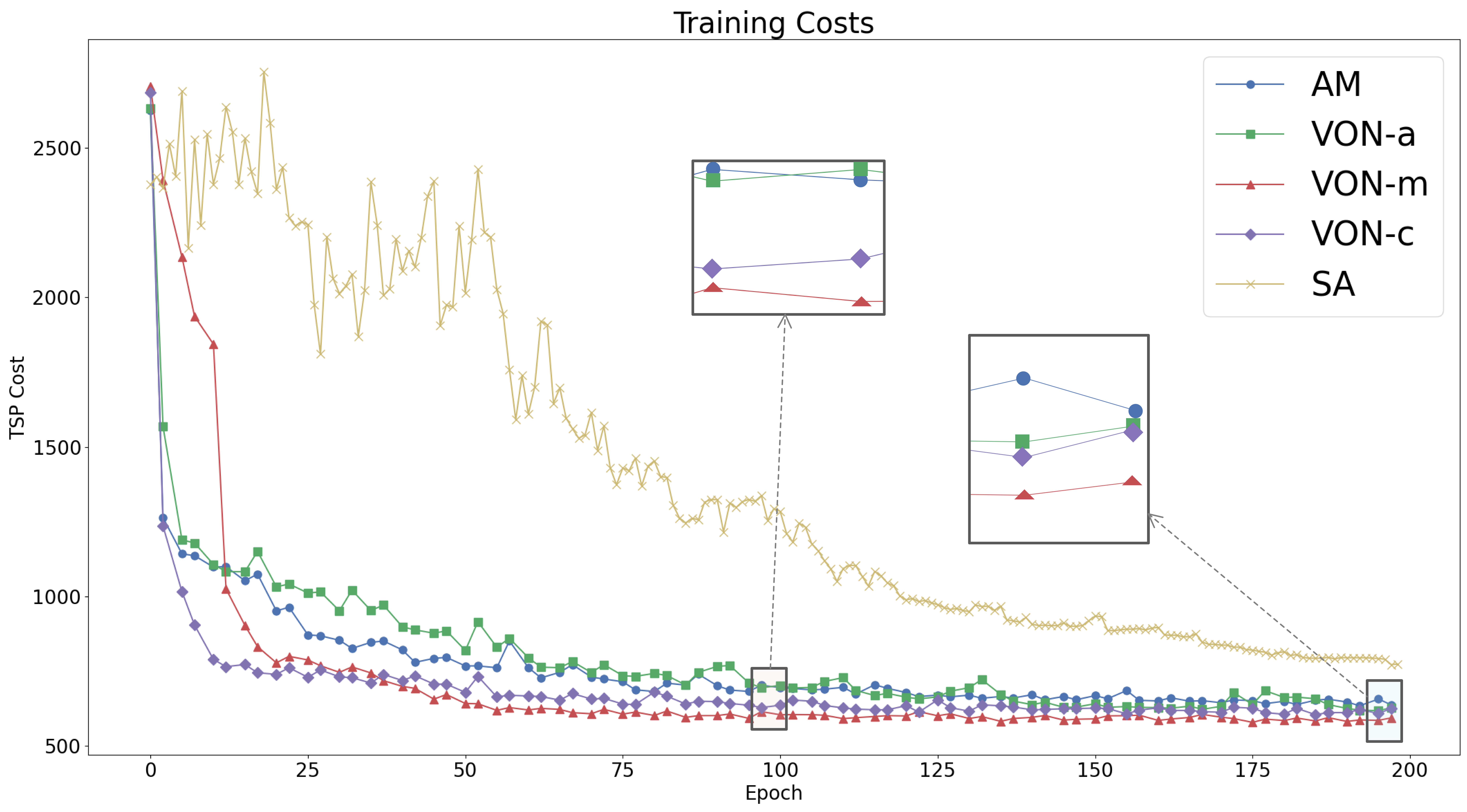}
  \caption{The training loss over 200 epochs for ordering images in Fashion-MNIST dataset using TSP metric. All models are trained using \fmdatafmix.}
  \label{fig:FM-loss}
\end{figure}

\textbf{Impact of mixed sampling strategies.}
Prior experiments reveal the impact of training data using individual sampling strategies. 
Here, we further study three mixed strategies (namely, $mix$, $gl$, and $lg$), to examine the impact of how the global and local samples are ordered in the training data. As shown in \cref{tab:FM-ablation}.III, the three models of \vonm trained with different data share similar performance, demonstrating a superior ability to learn across scales. In contrast, the performance of AM-based models heavily relies on the order of training samples. The models seem to be shaped by the first half of training samples. For example, while delivering reasonable performance in local data, $\amflg$ results in large distances in global data, such as $\fmdatafg$ (2150.1, $\times 3.10$ as $\vonmflg$) and $\fmdataag$ (2373.9, $\times 2.20$ as $\vonmflg$). Mixing the training samples across scales seems to achieve a balance: it may not lead to the best results, but it often delivers decent results and avoids the worst. Therefore, \emph{we will examine the performance using the $mix$ of the four sampling strategies in the remaining experiments.}

\textbf{Performance over training epochs.}
We study the learning efficiency by investigating how the loss changes over training epochs for each model. In \cref{fig:FM-loss}, while \vonc and AM have a rapid downward trend in the early stages of training, $\vonm$ shows the lowest loss after 100 epochs. The loss of SA fluctuates in the first 50 epochs and gradually decreases after that. Over the entire 200 epochs, the loss of SA is much higher than the learning-based techniques. The learning-based techniques seem to discover a reasonable ordering strategy by 100 epochs. 

\subsubsection*{C.I.2 Impact of individual modules}
In this section, we study the individual impact of core modules using the best variant (i.e., \vonm) and the most balanced sampling strategy (i.e., $mix$). Specifically, we study the \emph{encoder} to evaluate the benefit of transforming the points to latent space for ordering, and study the \emph{reposition} module as it is the core contribution to our decoder design. We compare the original model \vonm with three variants (namely, \abvonmue without the encoder, \abvonmur without repositioning, and \abvonmub without both) using 1D layout and matrix reordering.

\textbf{Ablation study using 1D layout.}
We use the Fashion-MNIST dataset and TSP distance for this study. \cref{tab:1D-ablation} shows \abvonmub (without both encoder and repositioning) performs significantly worse in all tests compared to other models. The variants \abvonmue and \abvonmur are consistently outperformed by the original \vonm, highlighting the critical roles of these modules. However, their individual impact seems to be similar, with \abvonmue excelling in \fmdatafl, \fmdataag, and \fmdataal, while lagging in \fmdatafg, \fmdataafg, and \fmdataafl. Notably, \abvonmur (without repositioning) closely matches \vonm in \fmdatafg (675.9 vs 609.5) but performs much worse in \fmdatafl (456.4 vs 199.6). This underscores the loss of transferability across distributions without the repositioning module, especially given the varied local sample distributions.

\begin{table}[!thbp]
  \caption{Ablation study on the performance in 1D layout using TSP distance and Fashion-MNIST dataset. \abvonmue, \abvonmur \ and \abvonmub \ are the variants of $\vonm$ without the encoder, repositioning, and both, respectively. All models are trained with data samples of ``\fmdatafmix''.}
  
  \scriptsize%
  \centering%
  \setlength{\tabcolsep}{1.6mm}{
  \begin{tabular}{%
            c|%
  	  	*{6}{r}%
  	}
  	\toprule
  	& \fmdatafg & \fmdatafl & \fmdataag & \fmdataal & \fmdataafg & \fmdataafl   \\
  	\midrule
        \vonm & \textbf{609.5} & \textbf{199.6} & \textbf{385.3} & \textbf{379.2} & \textbf{540.4} & \textbf{590.9} \\
        \abvonmue & 1381.3 &  295.9 & 623.3 & 743.2 & 1046.9 & 1304.0 \\
        \abvonmur & 676.9 & 456.4 & 849.7 & 912.6 & 985.1 & 1054.3 \\
        \abvonmub & 2925.1 & 682.7 & 2797.1 & 1857.8 & 4456.7 & 4515.9 \\              
  	\midrule               
  	\bottomrule
  \end{tabular}}
  \label{tab:1D-ablation}
\end{table}

\textbf{Ablation study using matrix reordering.}
As shown in \cref{tab:matrix-ablation}, the original \vonm models outperform the three variants for all three metrics, affirming the significance of both modules. However, the reposition module is less important in this task, as matrix reordering considers all data items and becomes less demanding in transferability. \abvonmur, with the encoder but lacking the repositioning, surpasses the other two variants in every metric. Among the other variants, \abvonmue slightly outperforms \abvonmub, though the difference is less pronounced. This suggests that while the repositioning module may partially offset the absence of the encoder, it cannot fully replace it.

\textbf{Findings.}
In general, the ablation study shows that the original module \vonm always performs the best in every case, and the encoder has a significant impact in every setting. The study also confirms the impact of the repositioning module, when the transferability is required to order data points across distribution (i.e., the \fmdatafl, \fmdataag, \fmdataal, \fmdataafg, and \fmdataafl settings for 1D layout). However, when the data follows the same distribution (i.e., matrix reordering and the \fmdatafg setting in 1D layout), the repositioning module only has a slight marginal effect.

\begin{table}[!htbp]
  \caption{%
  Ablation study on the performance in matrix reordering using SCH dataset and three metrics (i.e., LA, PR and BW).
 \abvonmue, \abvonmur \ and \abvonmub \ are the variants of $\vonm$ without encoder, repositioning, and both, respectively.}%
  
  \scriptsize%
  \centering%
  \setlength{\tabcolsep}{1.6mm}{
  \begin{tabular}{%
            c|%
  	  	*{3}{r}%
  	}
  	\toprule
  	& LA & PR & BW   \\
        \midrule
        \vonm & \textbf{51.77k} &  \textbf{11.60k} & \textbf{206.88}\\
        \abvonmue &  67.66k & 12.94k & 221.94\\
        \abvonmur & 56.72k & 12.35k & 214.41\\
        \abvonmub & 68.18k & 13.29k & 224.65\\
  	\midrule               
  	\bottomrule
  \end{tabular}}
  \label{tab:matrix-ablation}
\end{table}

\begin{figure*}[!htb]
  \centering
  \includegraphics[width=0.95\linewidth]{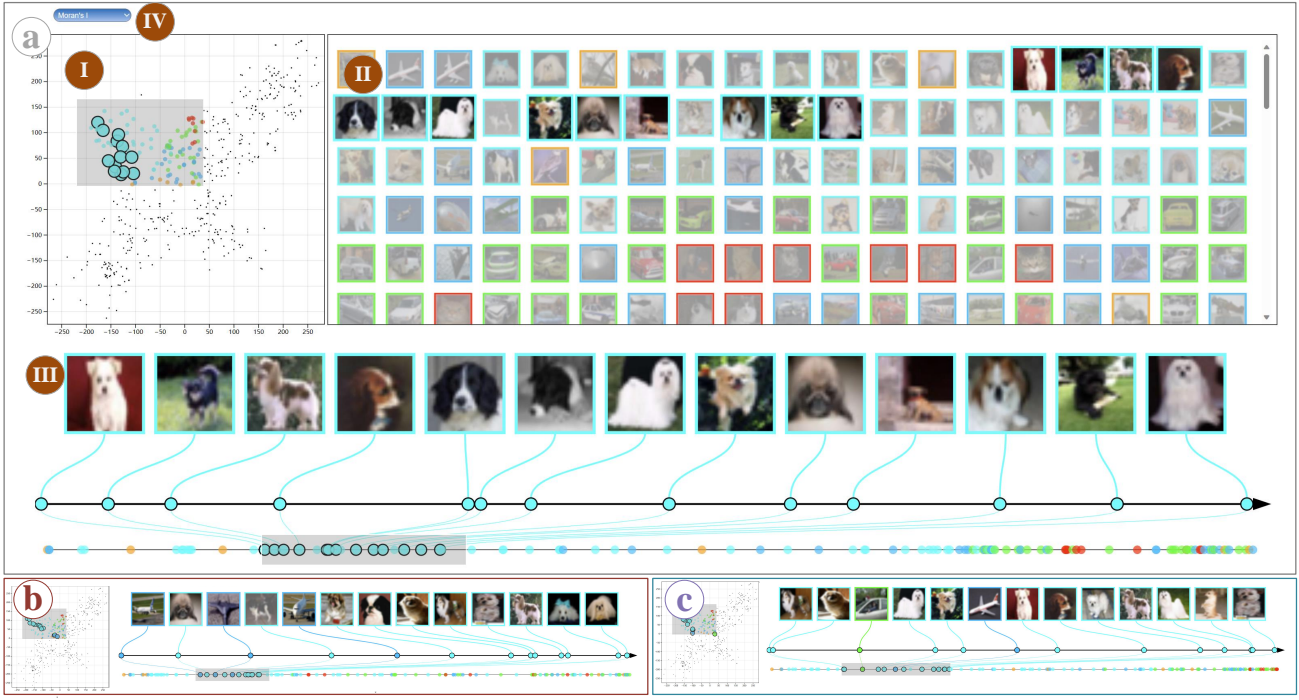}
  \caption{Exploring CIFAR-10 dataset using VON. The demo interface consists of: (I) a scatter plot for selecting images, (II) the collection of selected images ordered by VON, (III) a detailed view for exploring selected images along an axis, and (IV) a drop-down list for selecting quality metrics. The color of a dot indicates the category of the corresponding image. (a), (b), and (c) show the ordering results using Moran's I, TSP, and stress majorization, respectively.}
  \label{fig:VON-metric-evaluation}
\end{figure*}

\subsection*{C.II Dynamic Scenario}
\label{sec:dynamic-appendix}

\begin{figure*}[!htb]
  \centering
  \includegraphics[width=\linewidth]{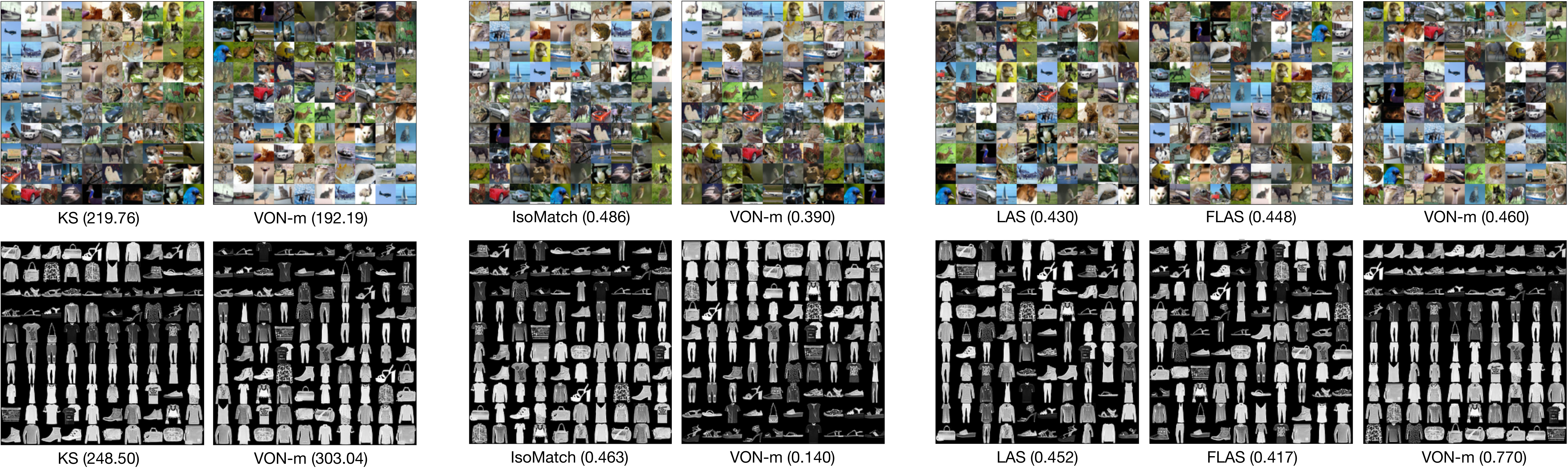}
  \caption{Comparison of image ordering results using the respective quality metrics of specialized baselines. The first row shows the results of CIFAR-10 dataset and the second row shows the results of Fashion-MNIST. 
  Each label shows the corresponding ordering approach and evaluated metric value. Note that the metric depends on the baseline being compared, and the images are ordered in 1D and displayed in the squared grid from left to right and top to bottom.
  }
  \label{fig:FM_CIFAR_baselineplus}
\end{figure*}

\textbf{Demo.}
We develop a demo system to explore the CIFAR-10 image collection. \cref{fig:VON-metric-evaluation} (a) shows the exploration results using Moran's I as the quality metric. We select a region containing four categories of images, indicated by the light blue, blue, green, and red dots, respectively, as shown in (a.I). Using Moran's I, the ordered images in (a.II) and the axis in (a.III) are correlated to their categories, with most dog images (light blue) on the left and the other three categories on the right. Users may further select images along the axis in (a.III) to investigate details. In \cref{fig:VON-metric-evaluation} (b) and (c), we switch the quality metric to TSP and stress, respectively. The order of images is still similar. Although several planes appear between the dog pictures, the overall appearance of the images is still similar. We should note that Moran's I seem to better separate the images of different categories. This may be due to the computation of Moran's I considers all pairwise relationships among images, instead of relationships between neighbors as in TSP. Please refer to the supplemental video for a better viewing experience of this usage case.

\begin{figure*}[!htb]
\begin{center}
$\begin{array}{c@{\hspace{0.2in}}c}
\includegraphics[width=.47\linewidth]{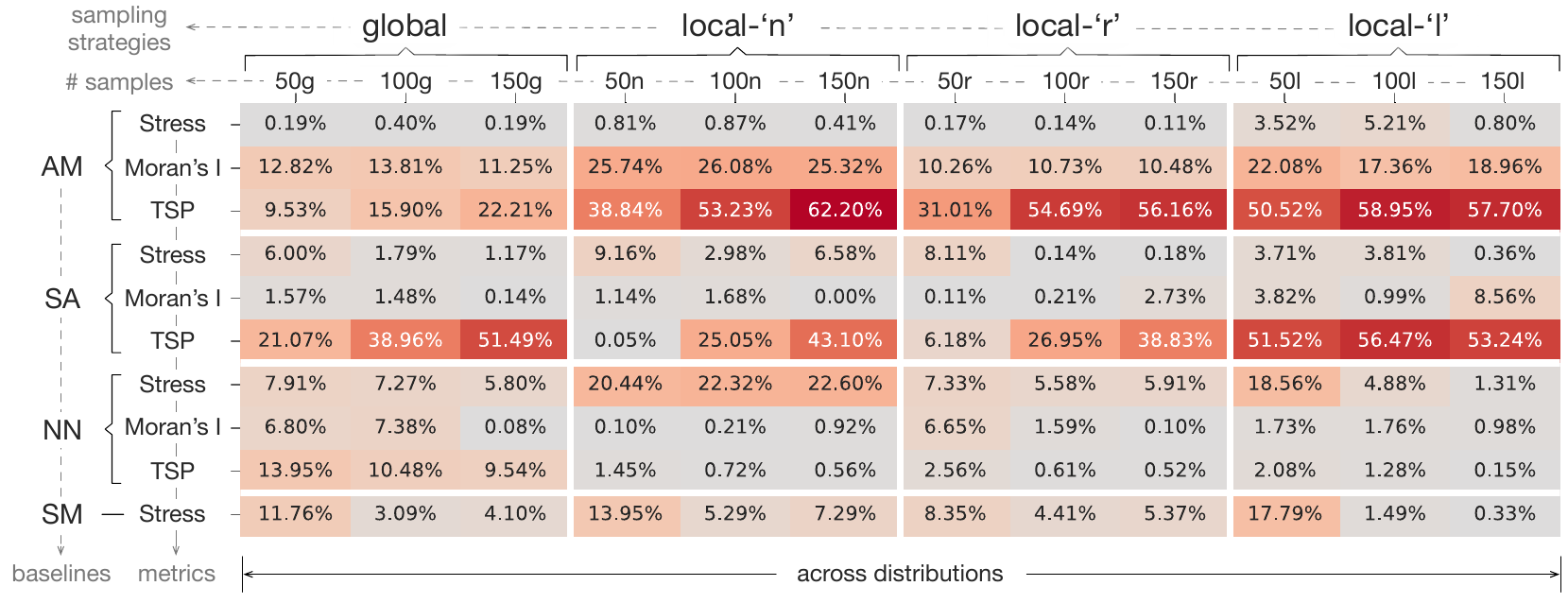}&
\includegraphics[width=.47\linewidth]{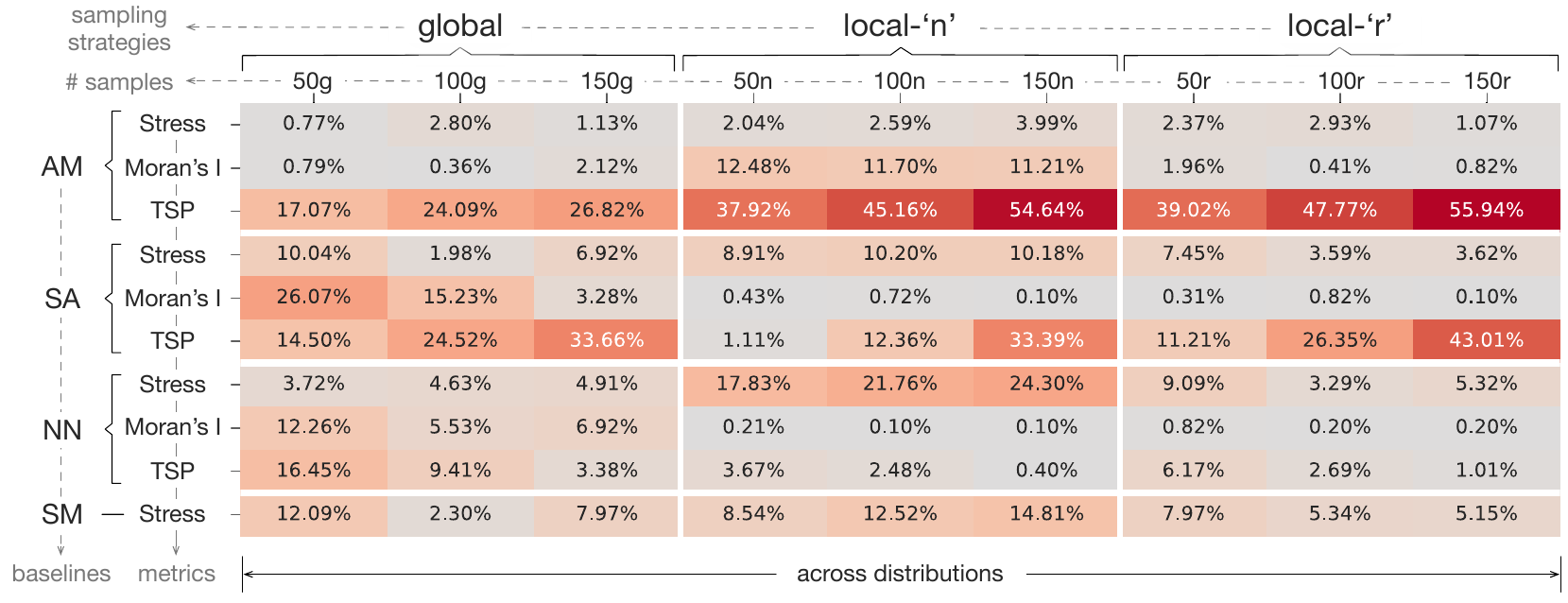}\\
\mbox{(a) MNIST} & \mbox{(b) ImageNet} \\
\includegraphics[width=.47\linewidth]{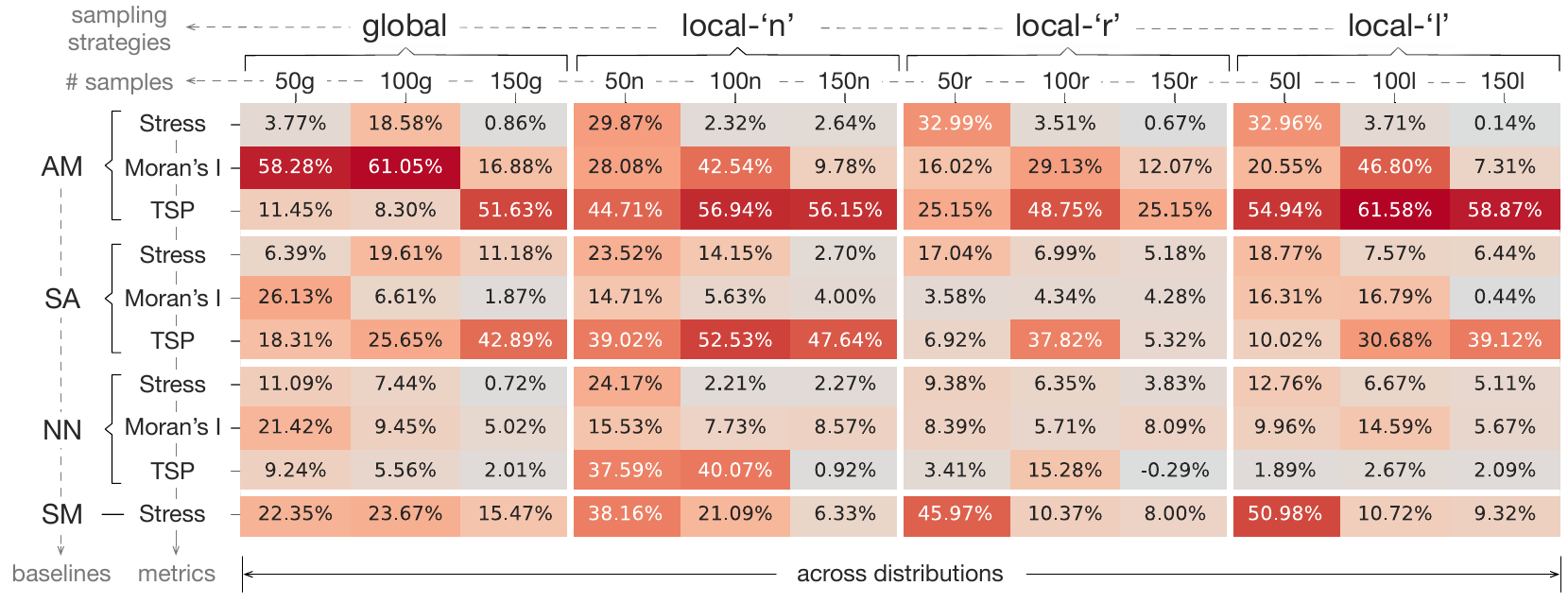}&
\includegraphics[width=.47\linewidth]{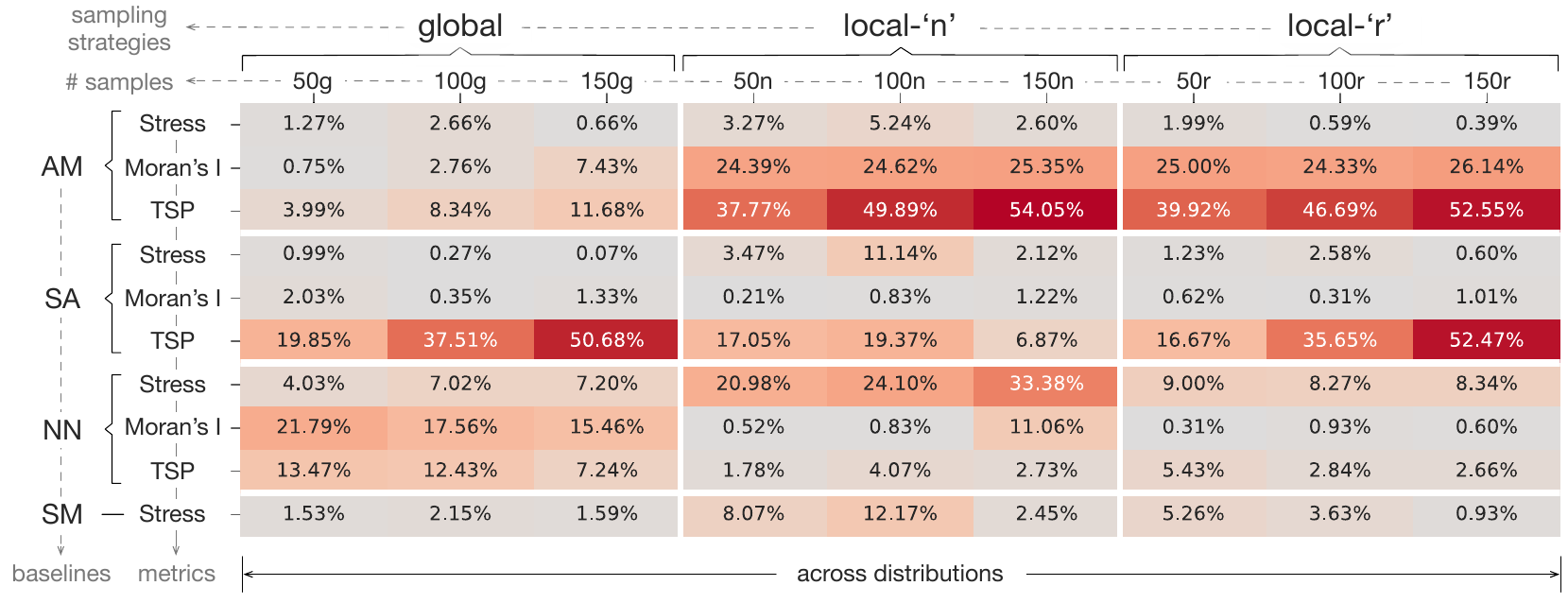}\\
\mbox{(c) Fashion-MNIST} & \mbox{(d) CORA} \\
\includegraphics[width=.47\linewidth]{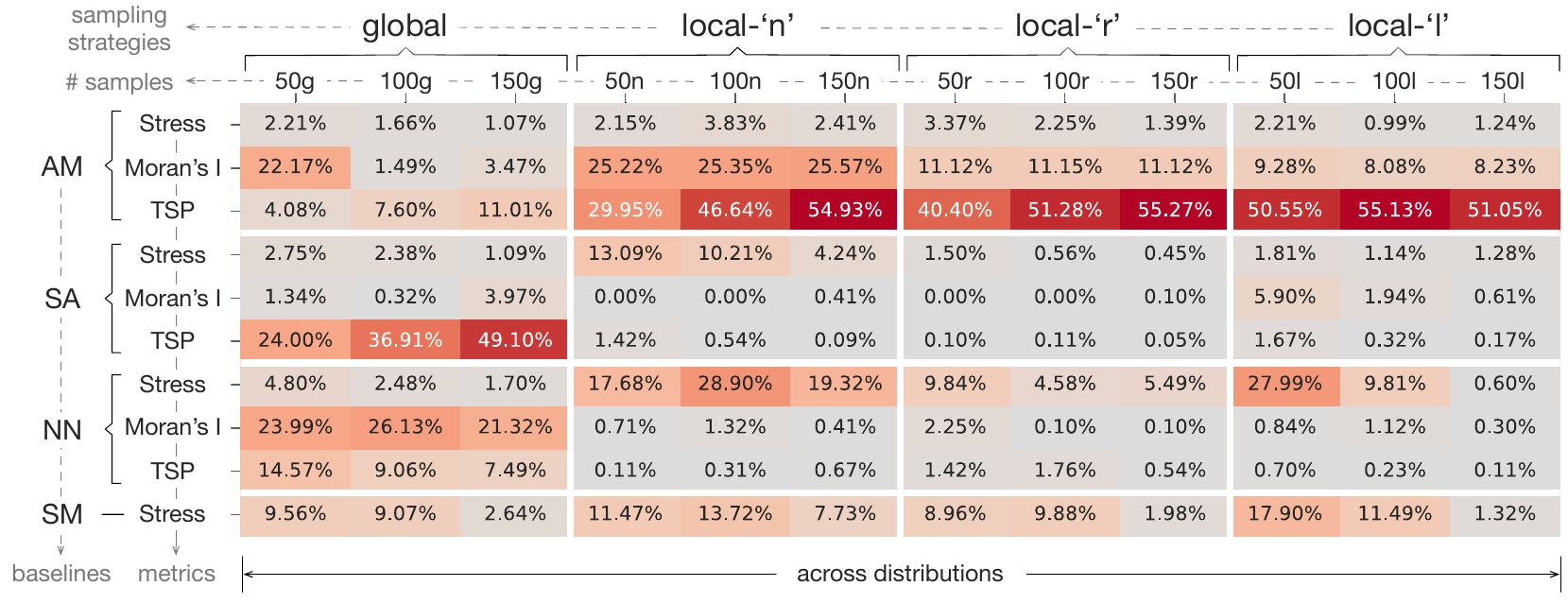}&
\includegraphics[width=.47\linewidth]{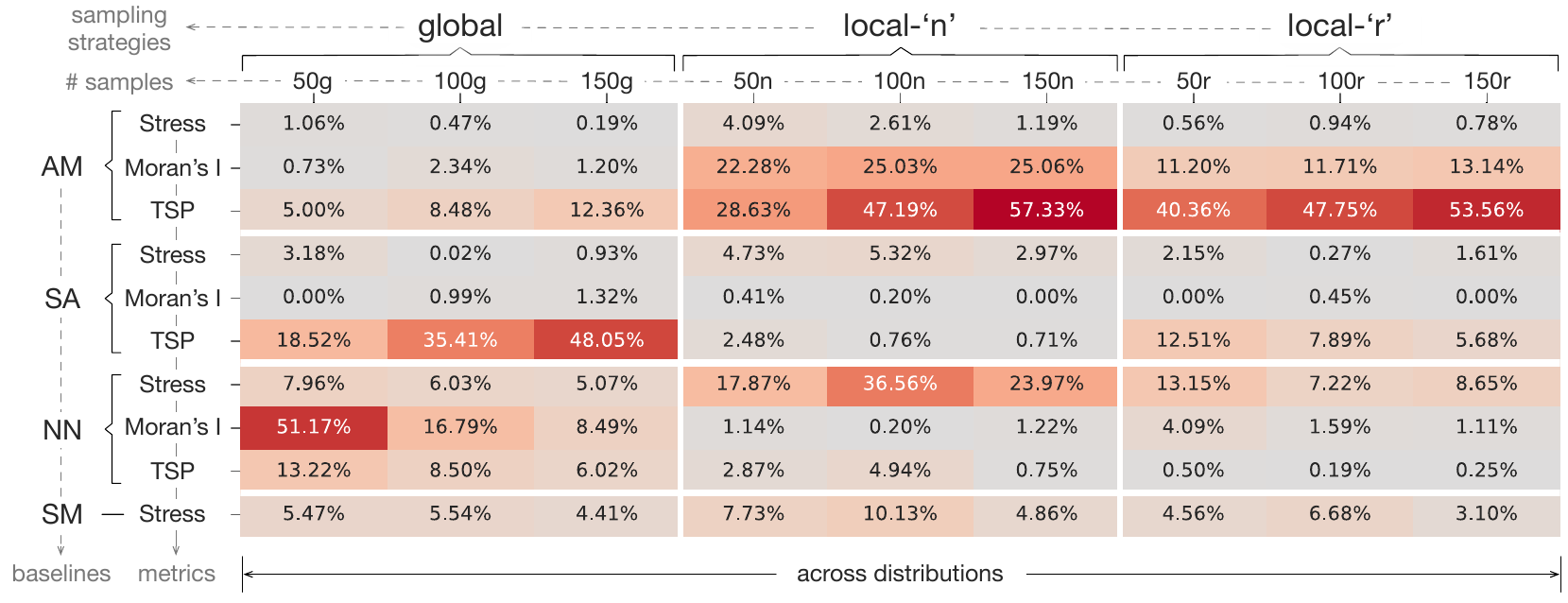}\\
\mbox{(e) CIFAR-10} & \mbox{(f) DBLP} \\
\end{array}$
\end{center}
\caption{Performance improvement of VON over AM, SA, NN, and SM for three metrics, using (a) MNIST, (b) ImageNet, (c) Fashion-MNIST, (d) CORA, (e) CIFAR-10, and (f) DBLP datasets.}
\label{fig:dynamic-individual}
\end{figure*}

\begin{figure*}[!htb]
\begin{center}
$\begin{array}{c@{\hspace{0.2in}}c}
\includegraphics[width=.47\linewidth]{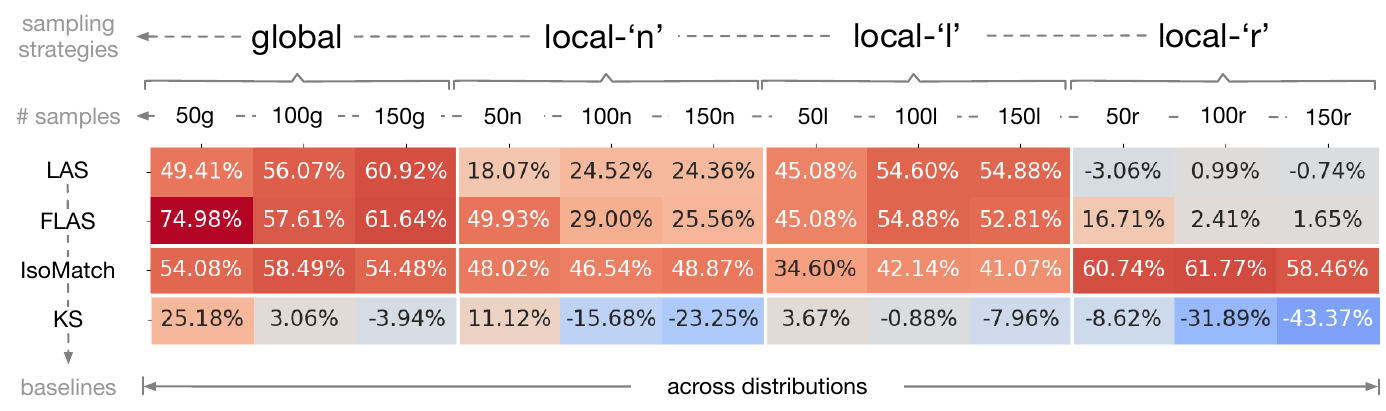}&
\includegraphics[width=.47\linewidth]{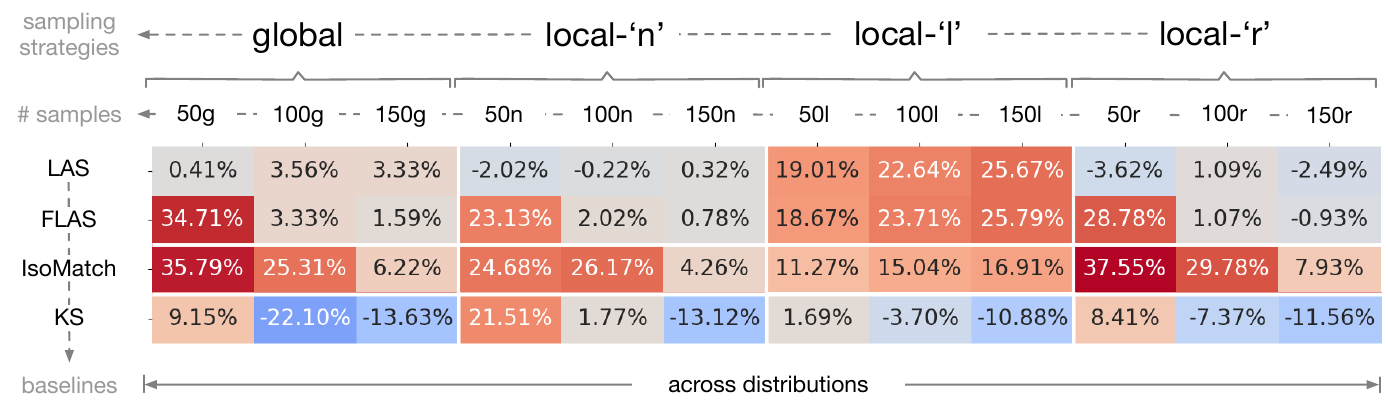}\\
\mbox{(a) MNIST} & \mbox{(b) CIFAR-10} \\
\includegraphics[width=.47\linewidth]{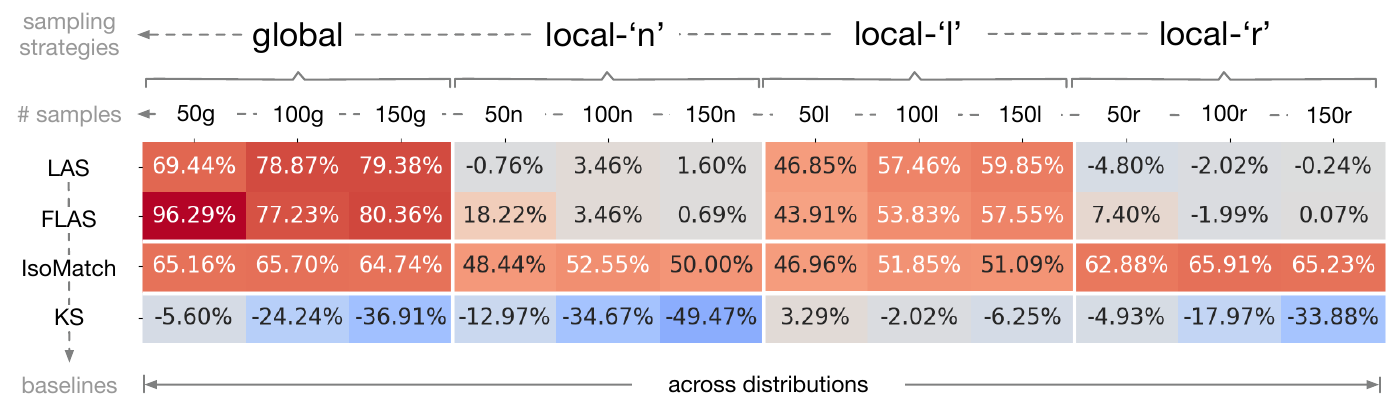}&
\includegraphics[width=.47\linewidth]{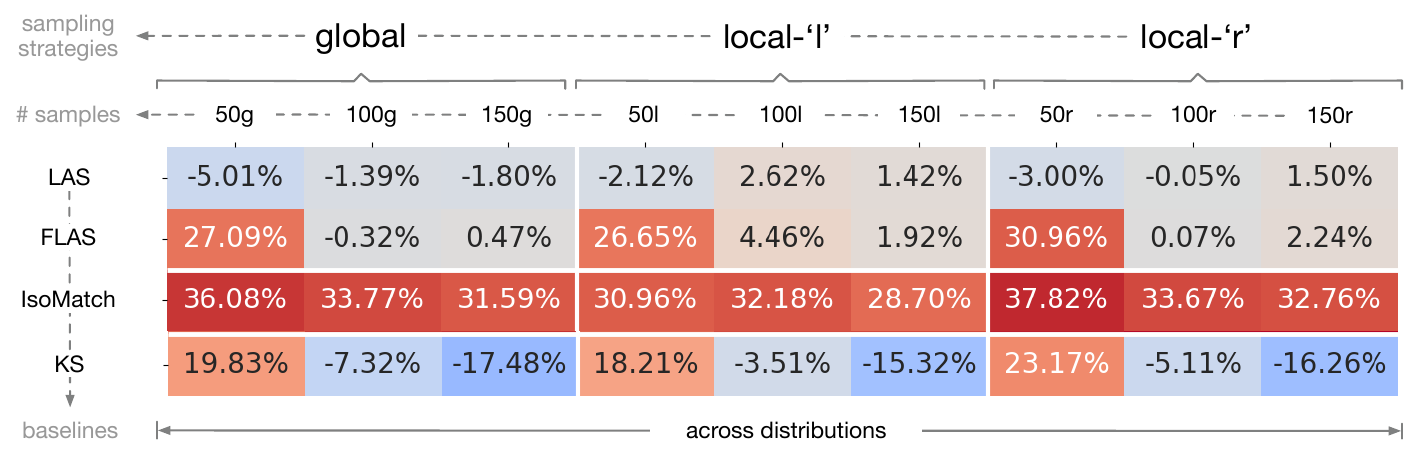}\\
\mbox{(c) Fashion-MNIST} & \mbox{(d) ImageNet} \\
\end{array}$
\end{center}
\caption{Performance improvement of VON over LAS, FLAS, IsoMatch, and KS for their respective metrics, using (a) MNIST, (b) CIFAR-10, (c) Fashion-MNIST, (d) ImageNet datasets.}
\label{fig:dynamic-individual_spc}
\end{figure*}

\textbf{Performance in dynamic scenario for individual datasets.}
\cref{fig:dynamic-individual} shows the complete results for individual datasets for dynamic scenarios. Please refer to \cref{sec:dynamic-quantitative} for the analysis of the average results over these datasets. Compared to the average improvement in \cref{fig:dynamic-all}, the improvement patterns for the individual datasets are similar in general. For example, similar to the average, we find significant improvement of \vonm over AM and SA in TSP for each dataset, especially using local sampling strategies. We also find that \vonm consistently outperforms the baselines, with a single negative improvement (-0.29\%). 

However, differences also exist. For example, in CIFAR-10 and DBLP, \vonm shows greater performance improvement over NN in Moran's I for the global samples, but less improvement for the local samples, when comparing to the average. Especially for the 50g samples, the improvement over NN reaches the maximum (51.17\%) in all cases. In the Fashion-MNIST and MNIST datasets, \vonm shows more improvement over the baselines, especially for AM and SA. However, unlike the average, the performance improvement often becomes most significant using the medium number of samples (100). Interestingly, we find that the improvement over NN in TSP is much larger than average in 50n (37.5\% vs. 7.91\%) and 100n (40.07\% vs. 8.77\%). This is unexpected, because NN usually excels in TSP and even achieves better performance than \vonm with the only negative improvement (-0.29\%) in TSP with 150r samples of the same dataset. In the ImageNet and CORA datasets, different from the average, \vonm shows a greater advantage over NN in stress, especially for the local-`n' samples.

For image reordering, we compare \vonm against IsoMatch \cite{fried2015isomatch} using energy value, Kernelized Sorting (KS)~\cite{quadrianto2008kernelized} using its objective, and LAS/FLAS~\cite{barthel2023improved} using DPQ. 
\cref{fig:dynamic-individual_spc} presents the detailed performance improvements across different datasets. It shows that \vonm generally outperforms FLAS and IsoMatch on all image datasets, especially when the testing size matches the size (50) used in training. However, we also find gaps between \vonm and LAS or KS, particularly for KS when the size of testing sets deviates from the training data. This indicates that the ordering strategy learned by \vonm becomes less effective when applied to larger collections of data.
We also examine the results qualitatively, by showing the ordering results of two randomly selected collections of 100 images in \cref{fig:FM_CIFAR_baselineplus}.
In all the cases, although the quantitative difference may be marginal to distinguish perceptually, \vonm produces comparable results to the respective baselines in general. This confirms to some degree that \vonm can be used in image ordering tasks.

To conclude, although the overall patterns are similar, the improvement percentage in a specific case may significantly fluctuate, leading to a large improvement value. We should note that although \vonm may not outperform the specialized methods in every case, it is generally usable, especially when a specialized algorithm is not available. This renders \vonm a reliable choice to avoid the worst cases.

\textbf{Timing performance.} We use the Fashion-MNIST dataset to compare the timing performance of different approaches for ordering 50 images, as shown in \cref{tab:losstime_dynamic}. The timing performance of the two learning-based approaches (\vonm and AM) are similar as they share similar sizes of networks. For these approaches, the training time varies across metrics. The reason is that the evaluation time across metrics is different, and these approaches need to evaluate the metrics during training. Their ordering time is stable and interactive, because once the metric is learned by the network, these approaches do not rely on the evaluation for ordering. In contrast, SA needs to evaluate the metrics during the ordering stage, leading to significantly different ordering time, which can be prohibitive. NN uses a fixed heuristic for ordering, achieving similar ordering performance as \vonm and AM.

\begin{table}[!htbp]
  \caption{%
  The timing results for dynamic scenario using Fashion-MNIST datasets. For the learning-based approaches, both training and ordering time are reported.
  }
  \scriptsize%
  \centering%
  \setlength{\tabcolsep}{1mm}{
  \begin{tabular}{%
            c|c|%
  	  	*{5}{c}%
  	}
  	\toprule
  	& & \vonm & AM & SA & NN & SM\\
  	\midrule
        TSP & Train & 2s/e & 2s/e & - & - & - \\
         & Order & 0.82s & 0.81s & 2.23s & 0.86s & - \\
        \midrule
        Stress & Train & 240s/e & 240s/e & - & - & - \\
         & Order & 0.82s & 0.81s & 932.59s & 0.86s & 20.93s\\
         \midrule
         Moran's I & Train & 11s/e & 11s/e & - & - & - \\
          & Order & 0.82s & 0.81s & 619.52s & 0.86s & - \\
  	\midrule               
  	\bottomrule
  \end{tabular}}
  \label{tab:losstime_dynamic}
\end{table}

\begin{figure*}[!htb]
\begin{center}
$\begin{array}{c@{\hspace{.2in}}c}
\includegraphics[width=.475\linewidth]{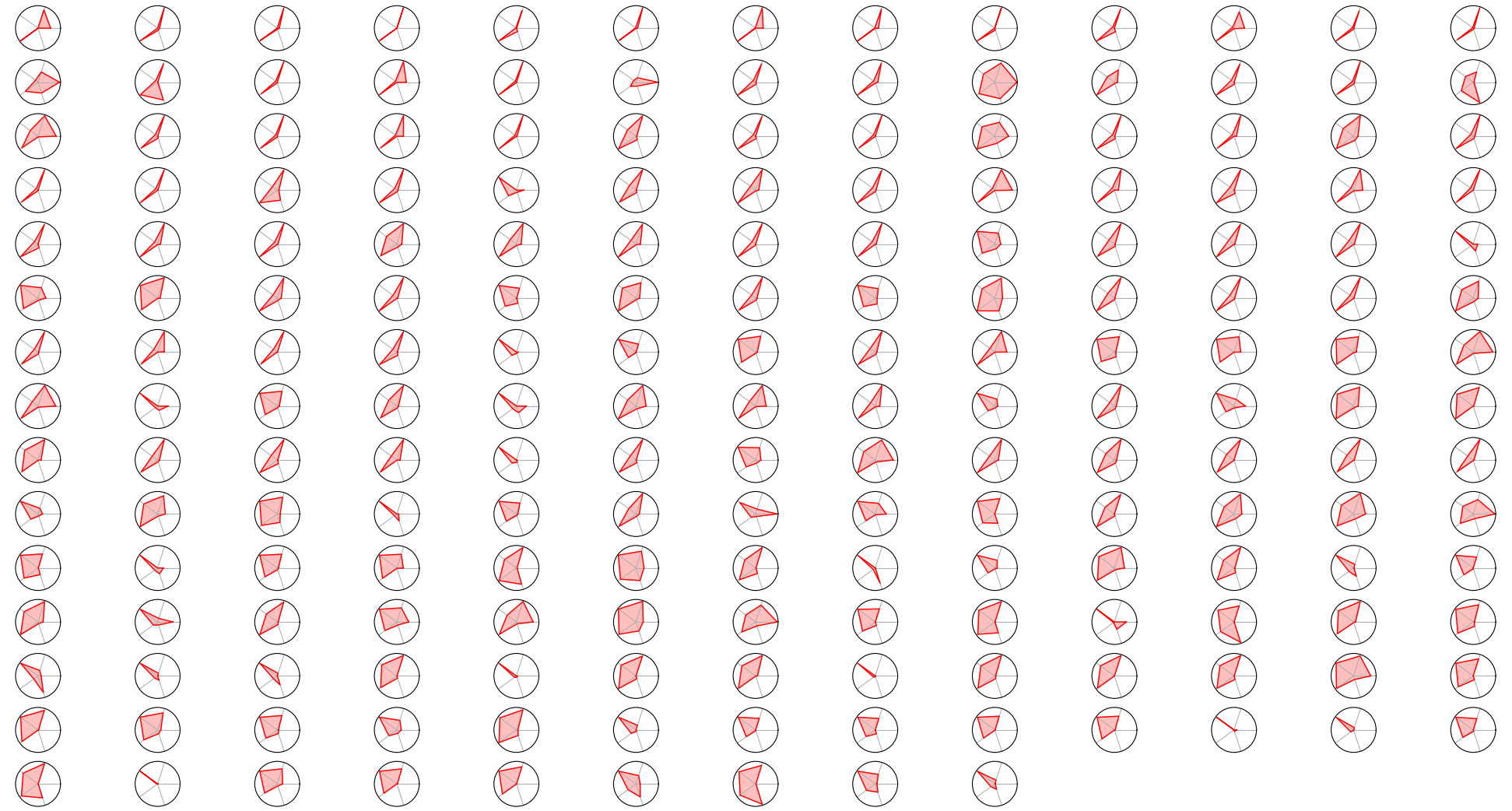}&
\includegraphics[width=.475\linewidth]{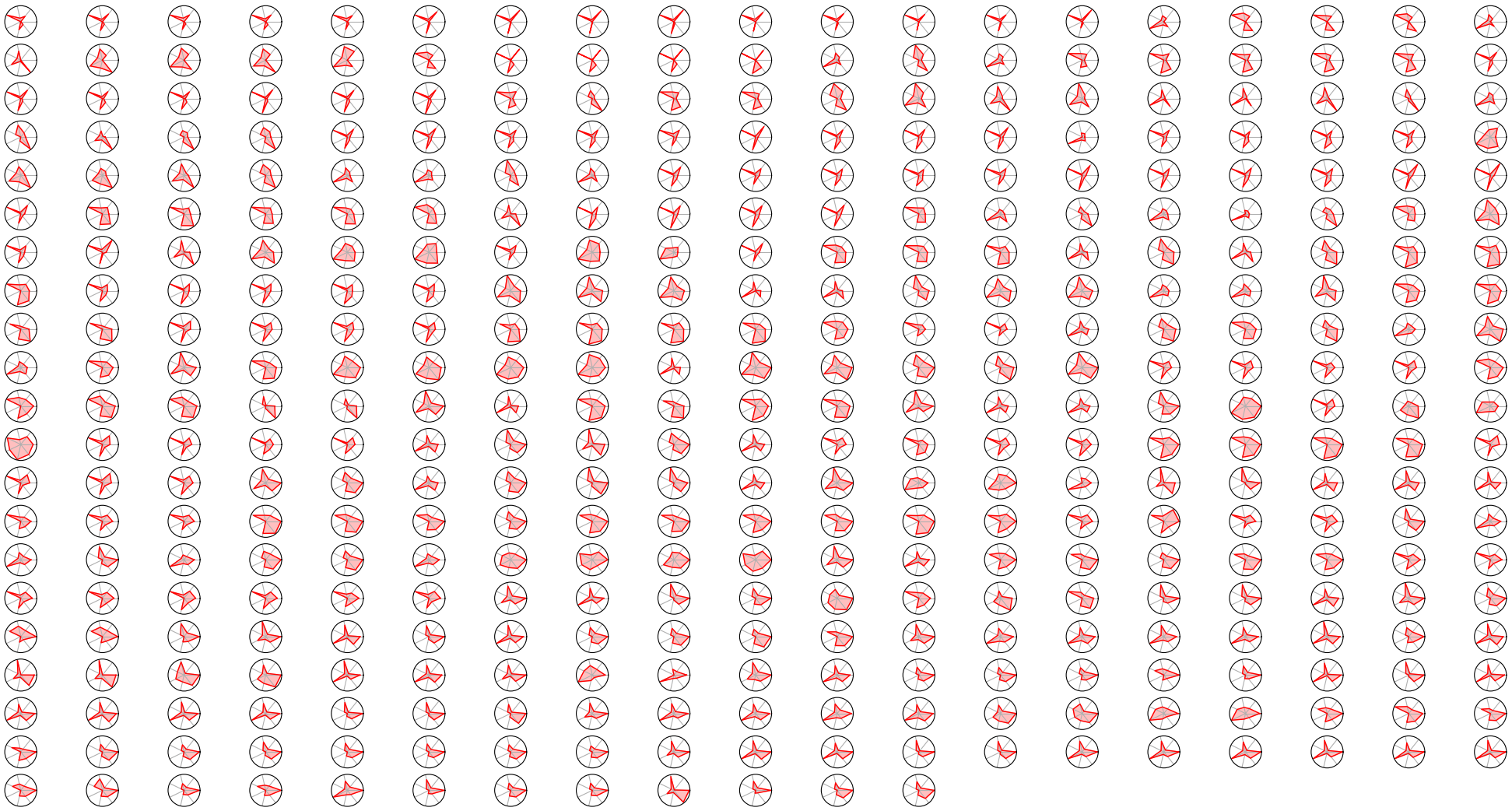}\\
\mbox{\vonm (0.0743)} & \mbox{\vonm (0.2166)}\\[10pt]
\includegraphics[width=.475\linewidth]{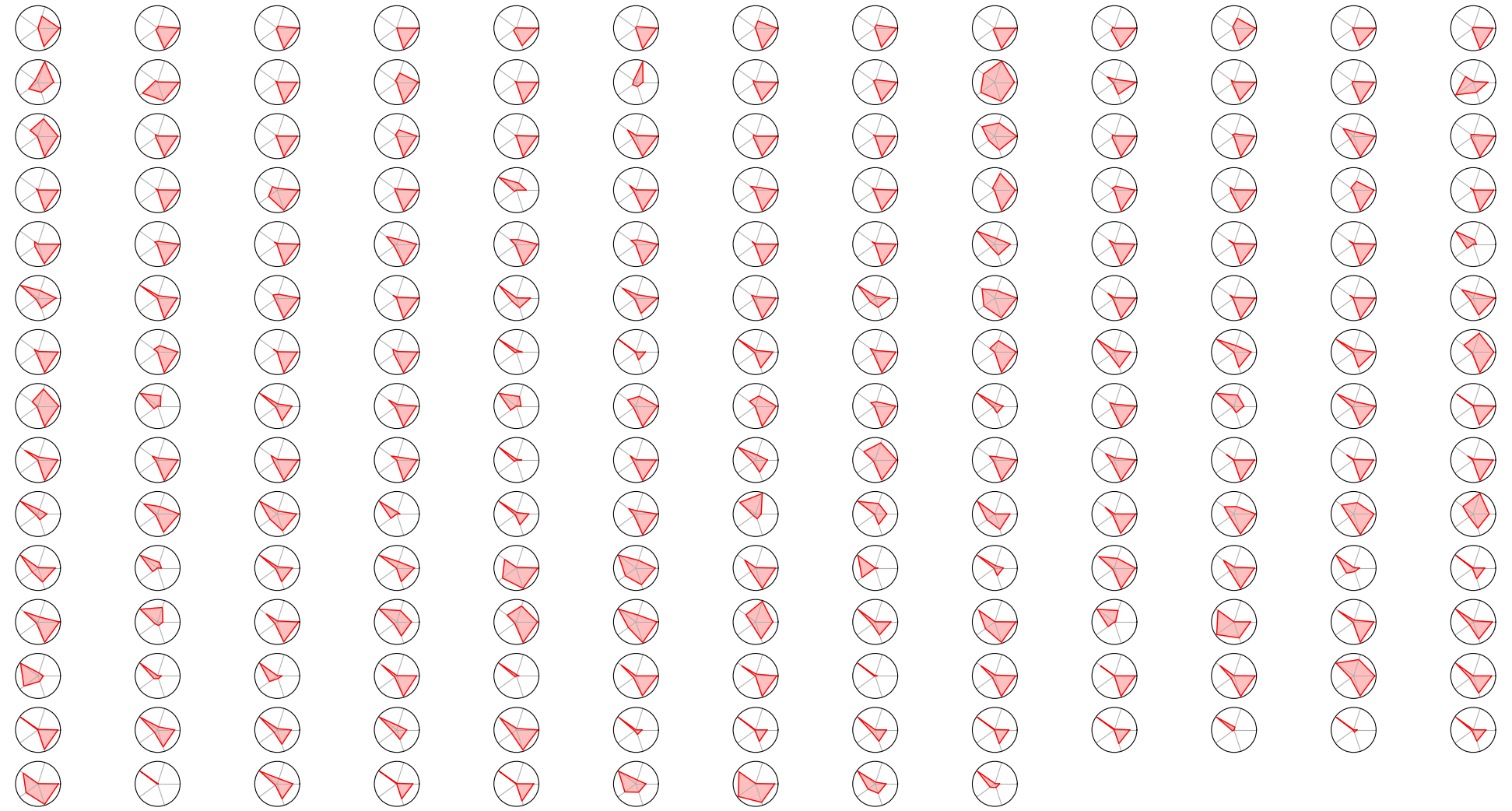}&
\includegraphics[width=.475\linewidth]{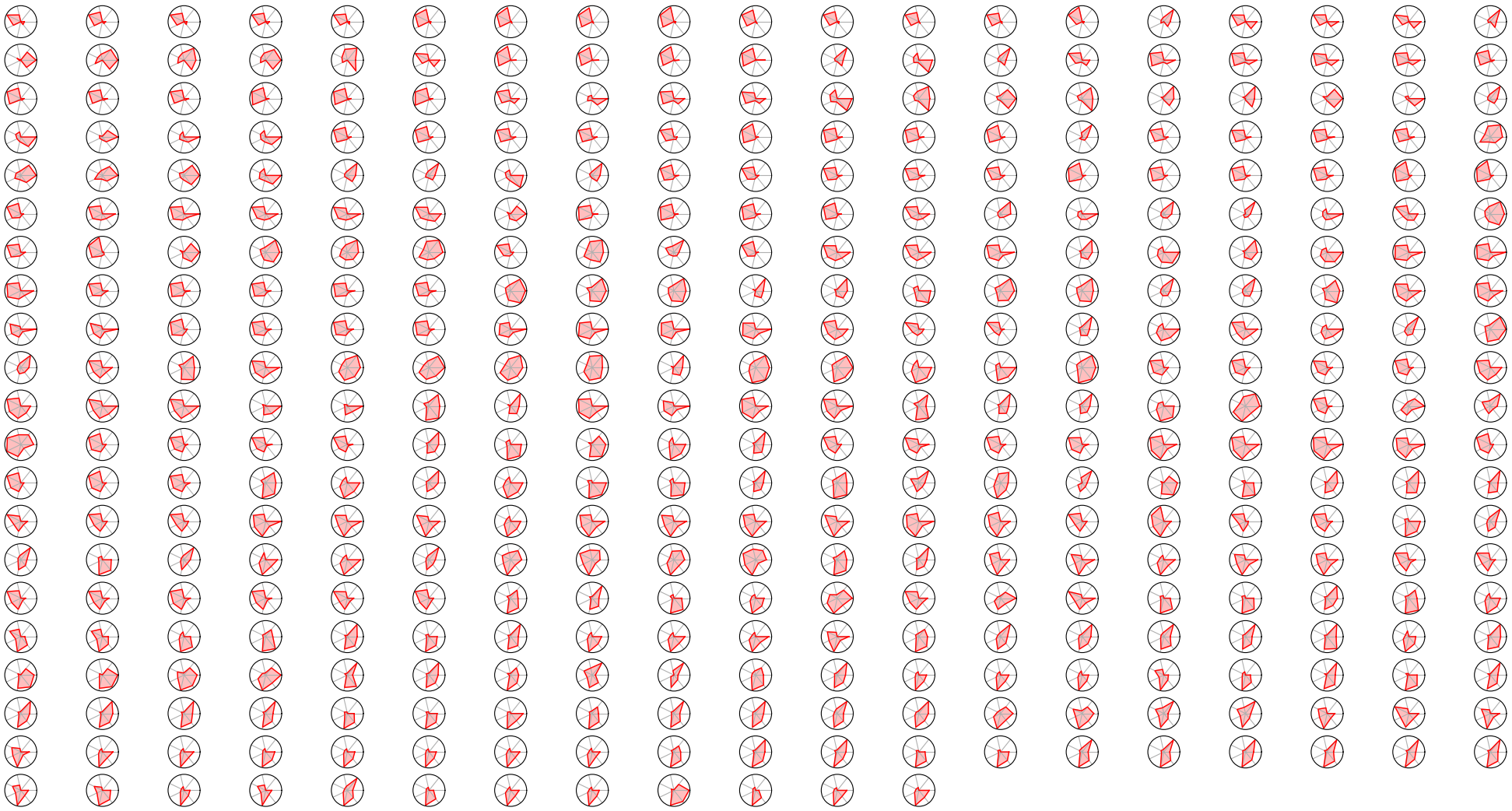}\\
\mbox{AM (0.0743)} & \mbox{AM (0.2285)}\\[10pt]
\includegraphics[width=.475\linewidth]{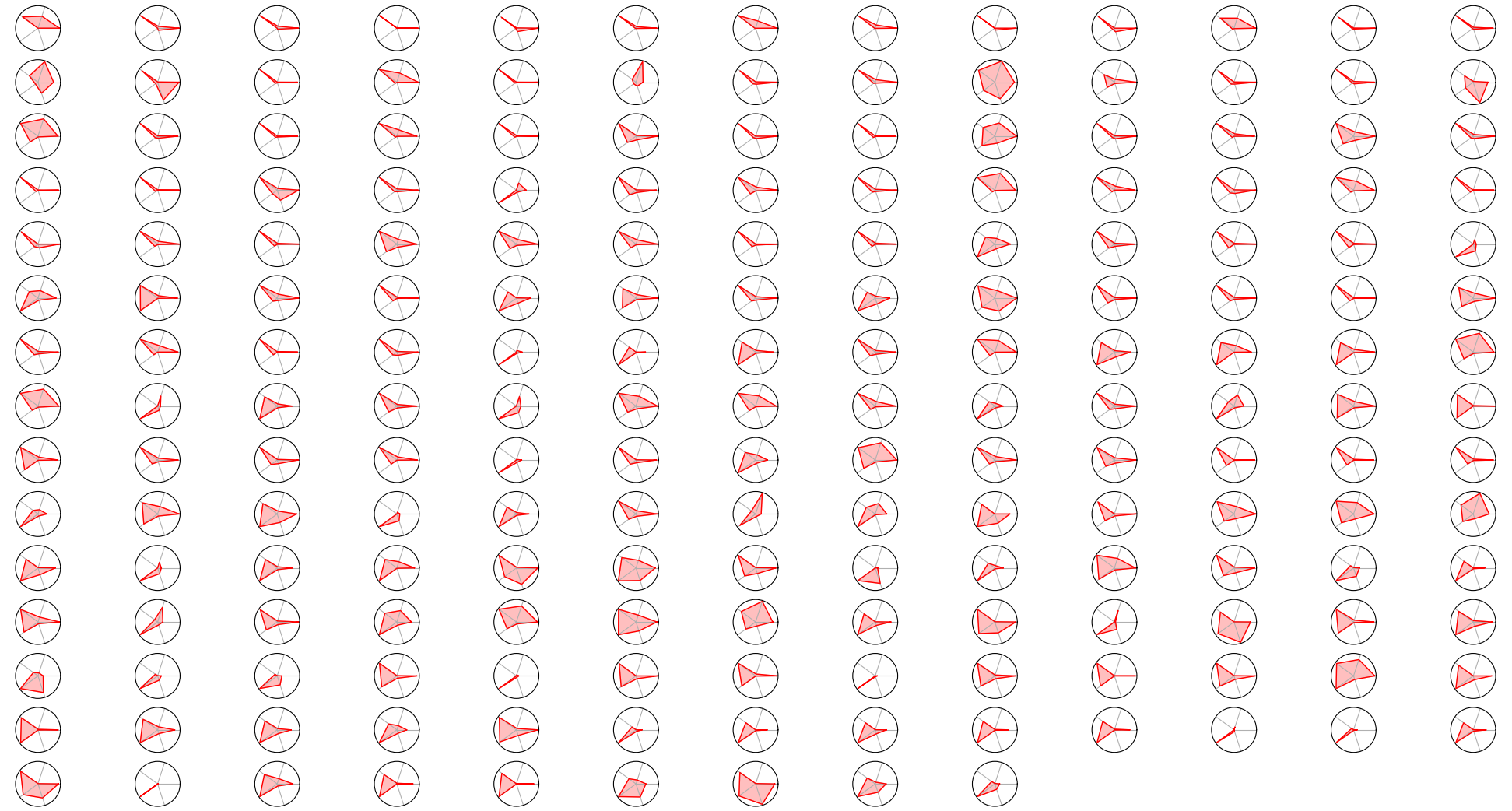}&
\includegraphics[width=.475\linewidth]{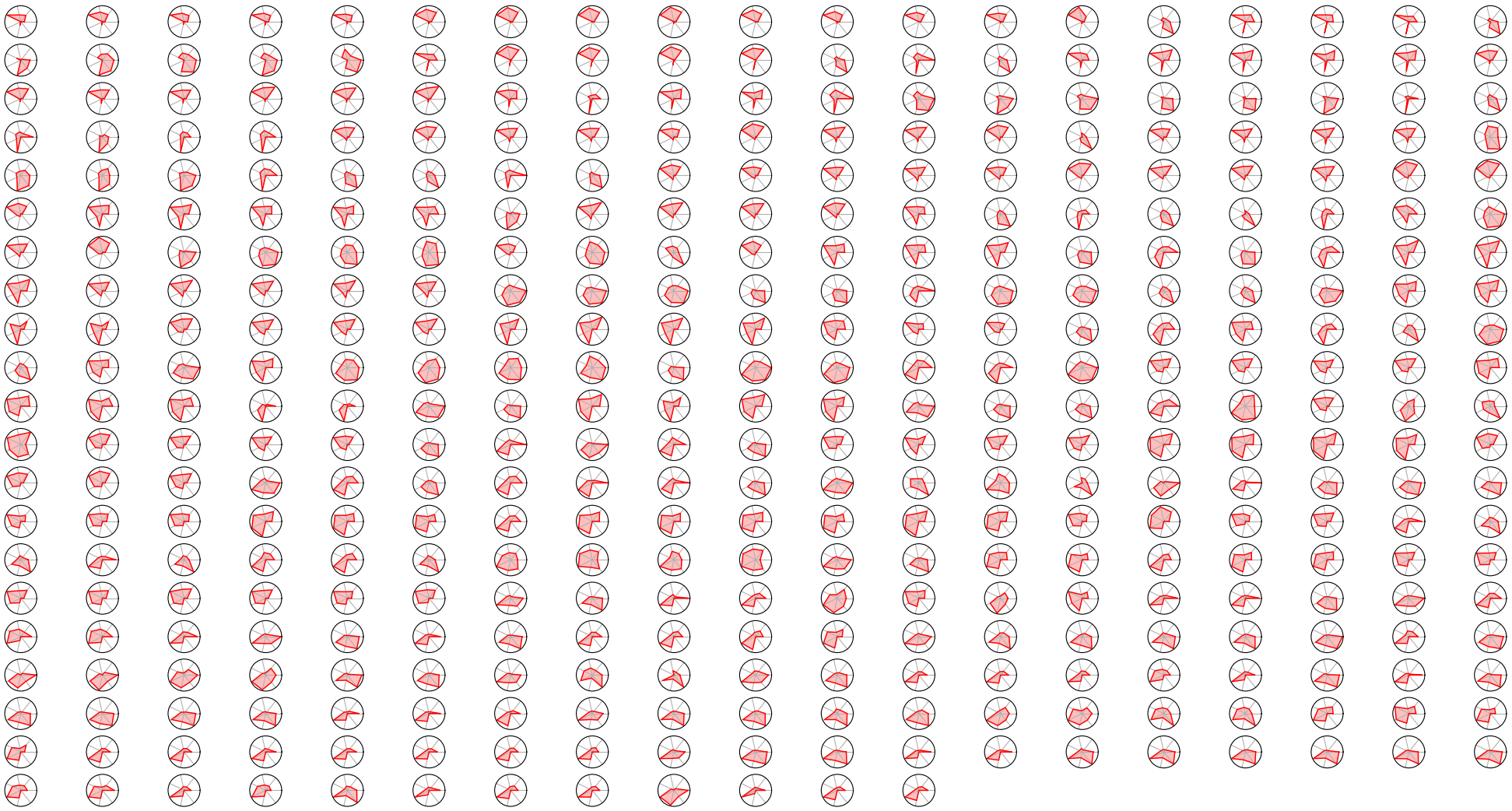}\\
\mbox{RS (0.0743)} & \mbox{RS (0.2166)}\\[6pt]
\mbox{(a) Coal Disaster} & \mbox{(b) Cars}
\end{array}$
\end{center}
\caption{Comparison of star plot axes reordering results of \vonm, AM, and random swapping (RS) using (a) Coal Disaster and (b) Cars datasets.}
\label{fig:star-pc-all}
\end{figure*}

\subsection*{C.III Star plot}
\textbf{Visualization results for the star plot.}
\cref{fig:star-pc-all} displays the axes reordering results reordering results for the star plot using the Cars and Coal Disaster datasets. The symmetry metric is used following \cref{sec:axes-reorder}. A smaller symmetry value indicates a better order of axes. In Coal Disaster with only five axes, all approaches identify the optimal solution. However, in the visualization results, we find that the orders produced by different approaches are visually distinct. In Cars with seven axes, a higher complexity can be observed in the metric, as \vonm and RS find the numerically optimal solution while AM does not. However, in the visualization results, we find that \vonm seems to learn a different strategy from AM and RS. \vonm tends to place attributes with high and low values alternately, resulting in `star'-like shapes. In contrast, AM and RS place attributes with similar values contiguously, leading to `fan'-like shapes. This example demonstrates a gap between the metric and human perception: while \vonm and RS share similar metric values, the visualization results of AM and RS seem to be visually closer. As our approach focuses on optimizing for metrics, we will not further explore this aspect. However, this example highlights the necessity of developing metrics that are better aligned with human perceptions. \emph{Our VON can facilitate this kind of research by providing a reliable and readily available tool for optimizing the customized metrics, so that perception researchers can focus on the design of metrics and visually assess whether the optimized results are consistent with human perception.}

\begin{table*}[!thbp]
  \caption{%
    Matrix reordering performance of \vonm using SCH dataset and different embedding approaches. Each approach embeds the data points into 2, 8, 16, and 32 dimensions to produce the input for \vonm. The ordering results are evaluated using LA, PR, and BW metrics~\cite{behrisch2020guiro}.
  }
  
  \scriptsize%
  \centering%
  \setlength{\tabcolsep}{1.6mm}{
  \begin{tabular}{%
            c|%
  	  	*{3}{c}|*{3}{c}|*{3}{c}|*{3}{c}%
  	}
  	\toprule
        &  & 2-dim &  &  & 8-dim & & & 16-dim & & & 32-dim & \\
        \midrule
  	& LA & PR & BW & LA & PR & BW & LA & PR & BW & LA & PR & BW \\
  	\midrule
        T-SVD & \textbf{40.71k} & 11.21k & 167.59 & 52.19k & 9.16k & 182.29 & \textbf{43.97k} & \textbf{9.50k} & \textbf{188.94} & 58.53k & \textbf{11.88k} & 208.94 \\
        PCA & 41.47k & \textbf{8.88k} & \textbf{162.18} & \textbf{51.52k} & \textbf{9.12k} & \textbf{179.35} & 44.23k & 10.02k & 201.65 & \textbf{54.78k} & 12.07k & \textbf{188.12} \\
        LLE & 68.60k & 13.74k & 204.12 & 70.71k & 12.72k & 196.35 & 80.78k & 14.38k & 214.65 & 102.4k & 15.92k & 222.65 \\
        AE & 59.52k & 12.04k & 207.24 & 67.28k & 12.25k & 215.35 & 63.59k & 13.48k & 195.18 & 70.56k & 13.41k & 188.76 \\
        Isomap & 58.25k &  12.30k & 203.59 & 60.47k & 11.95k & 196.94 & 64.29k & 11.66k & 210.76 & 67.76k & 12.60k & 208.47 \\
  	\midrule               
  	\bottomrule
  \end{tabular}}
  \label{tab:sch-emb-ablation}
\end{table*}

\textbf{Timing performance.} 
We use the Census Income dataset, which has the most number of axes, to examine the timing performance. The computation time for one epoch is around half a second using each approach: it takes 0.61s for \vonm, 0.50s for AM, 0.36s for SA, and 0.36s for RS. Note that learning-based approaches (\vonm and AM) are slower, because they require additional representation time beyond metric evaluation. This may damage the performance in static scenario, but it will be worth the time in dynamic scenario, as the network will collect useful information for ordering efficiently.

\begin{table}[!thbp]
  \caption{%
  Perfomance of \vonm on matrix reordering using SCH dataset and different embedding approaches to generate initial coordinates. The results are evaluated using LA, PR, and BW metrics.
  }
  
  \scriptsize%
  \centering%
  \setlength{\tabcolsep}{1.6mm}{
  \begin{tabular}{%
            c|ccc%
  	  	*{3}{r}%
  	}
  	\toprule
  	& LA & PR & BW   \\
        \midrule
        PCA-8D to t-SNE & 40.01k & 8.28k & \textbf{174.59} \\
        T-SVD-8D to t-SNE & \textbf{39.08k} & \textbf{8.23k} & 178.41 \\
        t-SNE & 51.77k & 11.60k & 206.88 \\              
  	\midrule               
  	\bottomrule
  \end{tabular}}
  \label{tab:sch-ablation}
\end{table}

\subsection*{C.IV Matrix reordering}

\subsubsection*{C.IV.1 Impact of embedding approaches}
\label{sec:embed}
The embedding approach is used to generate the initial point coordinates for graphs without node coordinates. In this experiment, we study its impact by embedding the adjacency matrix of the SCH dataset for ordering.

\textbf{t-SNE and 2-dimensional embedding.}
We only use t-SNE for 2D embedding, as t-SNE recommends an additional dimension reduction approach for high dimensional data (e.g., PCA for dense data and T-SVD for sparse data) before applying t-SNE. We compare t-SNE with ``PCA-8D to t-SNE'' (using PCA to reduce data in 8-dimension and then applying t-SNE) and ``T-SVD-8D to t-SNE'' (using T-SVD before t-SNE). The results are shown in \cref{tab:sch-ablation}. PCA and T-SVD may help to reduce noise in the high dimensional data, which leads to better ordering results than the original t-SNE in all three metrics.

\textbf{Impact of embedding approaches and numbers of dimensions.}
We further study other embedding approaches using various numbers of dimensions. Surprisingly, we find that the two linear dimension reduction approaches (i.e., PCA and T-SVD) perform exceptionally well. These two approaches share similar performance to each other and outperform other competitors in almost every setting. For example, for LA with 16D embedding, T-SVD (43.97k) and PCA (44.23k) perform similarly, while all the other approaches have LA values larger than 64k. In terms of the number of dimensions, we do not find a clear clue to favor a smaller or a larger number of dimensions. In general, we will follow the suggestion from the developer of t-SNE to use PCA for dense data and T-SVD for sparse data, and we will suggest using 2D embedding space for efficiency.

\subsubsection*{C.IV.2 Additional results}

\begin{table}[!htbp]
  \caption{%
  The timing results for matrix reordering using FTL datasets. For the learning-based approaches, both training and ordering time is reported. For the other approaches, only the ordering time is included.
  }
  \scriptsize%
  \centering%
  \setlength{\tabcolsep}{1mm}{
  \begin{tabular}{%
            c|%
  	  	*{5}{c}%
  	}
  	\toprule
  	 & VON-a & VON-m & VON-c & AM & SA \\
  	\midrule
        Train & 17s/e & 17s/e & 17s/e & 17s/e & -\\
        Order & 0.81s & 0.81s & 1.14s & 0.81s & 40s\\
  	\bottomrule
        \toprule
  	 & C-LO-$\delta_I$ & U-LO-$\delta_I$ & U-BC & C-BC & NN \\
  	\midrule
        Order & 0.21s & 0.19s & 0.16s & 0.17s & 0.70s\\
  	\midrule               
  	\bottomrule
  \end{tabular}}
  \label{tab:losstime}
\end{table}

\textbf{Timing performance.}
The timing results for matrix reordering are shown in \cref{tab:losstime}. \vona and \vonm cost similar training and ordering time as the other learning-based approach AM. \vonc has similar training time, but requires more time for ordering. These learning-based approaches are all much faster than SA, which requires a large amount of random moves to optimize the order. The specialized approaches are several times faster in matrix ordering, while VON is still interactive with less than one second in response time. Additionally, we should note that VON can be used for various metrics, which reduces the effort to develop and implement specialized algorithms.

\begin{figure*}[!htb]
\includegraphics[width=1\linewidth]{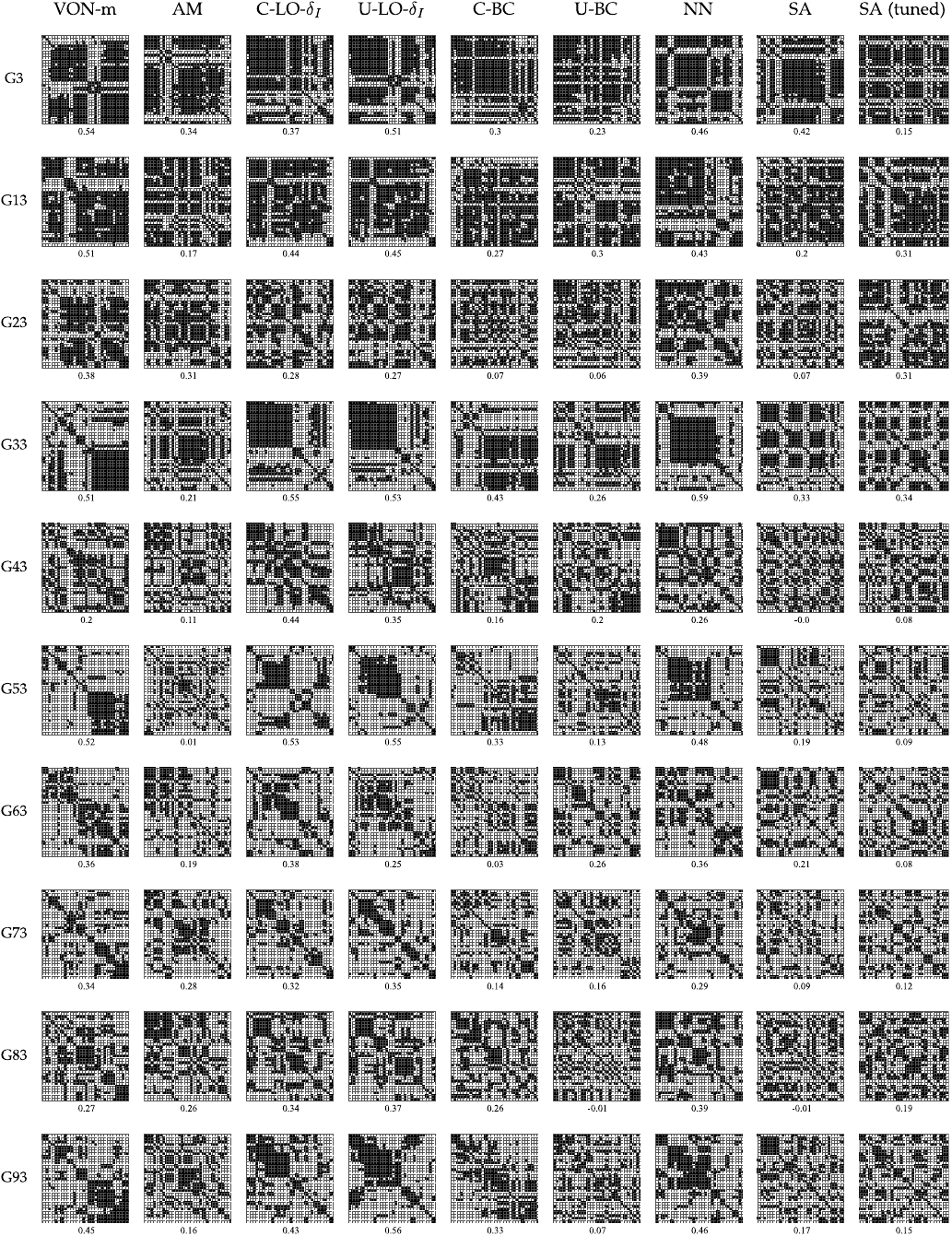}
\caption{Comparison of matrix reordering results of \vonm, AM, C-LO-$\delta_I$, U-LO-$\delta_I$, C-BC, U-BC, NN, SA, and SA with parameter tuning using the FLT dataset. Each row shows the results corresponding to one time step of the dynamic graph, evenly sampled from all steps starting from time step 3.}
\label{fig:matrix-all}
\end{figure*}

\textbf{Visualization results.}
\cref{fig:matrix-all} shows the visualization results of matrix reordering using FLT dataset. The results are evenly sampled with an interval of 10, starting from~3. Please refer to the supplemental material for the complete results. We find that \vonm, C-LO-$\delta_I$, U-LO-$\delta_I$, and NN deliver better performance than the other approaches, as we can observe several matrices with large block of black cells, especially at $G3$, $G33$, and $G53$. For the other four approaches, these kinds of blocks are rare, indicating the lack of ability to capture graph communities. This is inline with the quantitative results in \cref{sec:matrix-reordering}. We do find the C-LO-$\delta_I$ and U-LO-$\delta_I$ achieve slightly larger Moran's I values, but the improvement seems to be marginal both quantitatively and qualitatively. 

\textbf{The impact of parameter tuning for simulated annealing (SA).} We conduct additional experiments to verify the matrix ordering results using SA, as our default parameter setting leads to inferior performance for this task. 
SA involves three key parameters: initial temperature, restart, and cooling rate. The suggested cooling rate is between 0.8 and 0.99~\cite{dennis2006Simulated}, and we use the median 0.9 as the cooling rate in our default setting. The restart is essentially the number of repeated experiments. In our default setting, we use 5 to strike a balance between the performance gain and the additional cost of time. 

Our default setting of SA leads to an average Moran's I of 0.09 for the 96 graphs using FLT dataset, which is similar to the UB-C performance reported by Vanbeusekom et al.~\cite{vanbeusekom2022simul}. Following Brusco et al.\cite{brusco2008heuristic}'s settings, we use a cooling rate of 0.95 and restart of 20 for comparison. This setting leads to a decreased rate of cooling and an increased number of repeated randomized procedure. With this setting, the mean Moran's I increases from 0.09 to 0.258, but the timing cost to sort one collection also rises from 40 seconds to 261.3 seconds. Considering the timing performance, we do not experiment the more exhaustive settings. For a qualitative assessment, please refer to \cref{fig:matrix-all}.
\clearpage

\end{document}